\pdfoutput=1
\documentclass[11pt]{article}
\usepackage[OT1]{fontenc}
\usepackage{smile}

\usepackage[margin=1in]{geometry}

\newcommand{\bbP}{\mathbb{P}}
\newcommand{\bbR}{\mathbb{R}}
\newcommand{\bbE}{\mathbb{E}}
\newcommand{\wh}{\widehat}
\newcommand{\wt}{\widetilde}
\newcommand{\ol}{\overline}
\title{\huge Decoding Rewards in Competitive Games:\\ Inverse Game Theory with Entropy Regularization}
\author{
Junyi Liao$^1$,\ \ 
Zihan Zhu$^2$,\ \ 
Ethan X. Fang$^3$,\ \ 
Zhuoran Yang$^4$,\ \ 
Vahid Tarokh$^1$\\
~\\
$^{1}$\small\it Department of Electrical and Computer Engineering, Duke University \\
$^{2}$\small\it Department of Statistics and Data Science, University of Pennsylvania \\
$^{3}$\small\it Department of Biostatistics and Bioinformatics, Duke University \\
$^{4}$\small\it Department of Statistics and Data Science, Yale University
\\
\small\sf \{junyi.liao, ethan.fang, vahid.tarokh\}@duke.edu, zhzhu1@wharton.upenn.edu, zhuoran.yang@yale.edu
}
\date{}
\begin{document}
\maketitle
\begin{abstract}
Estimating the unknown reward functions driving agents' behaviors is of central interest in inverse reinforcement learning and game theory. To tackle this problem, we develop a unified framework for reward function recovery in two-player zero-sum matrix games and Markov games with entropy regularization, where we aim to reconstruct the underlying reward functions given observed players' strategies and actions. This task is challenging due to the inherent ambiguity of inverse problems, the non-uniqueness of feasible rewards, and limited observational data coverage. To address these challenges, we establish the reward function's identifiability using the quantal response equilibrium (QRE) under linear assumptions. Building upon this theoretical foundation, we propose a novel algorithm to learn reward functions from observed actions. Our algorithm works in both static and dynamic settings and is adaptable to incorporate different methods, such as Maximum Likelihood Estimation (MLE). We provide strong theoretical guarantees for the reliability and sample efficiency of our algorithm. Further, we conduct extensive numerical studies to demonstrate the practical effectiveness of the proposed framework, offering new insights into decision-making in competitive environments.
\end{abstract}
\section{Introduction}\label{sec:Intro}

Understanding the underlying reward functions that drive agents' behavior is of central importance in inverse reinforcement learning (IRL) (\citealp{Ng2000invRL}; \citealp{arora2020surveyinversereinforcementlearning}). While classical reinforcement learning (RL) literature (\citealp{sze2010RL}; \citealp{SuttonRL}) focuses on solving optimal policies given a known reward function, IRL inverts this process, aiming to infer the reward function from the observed behaviors of different agents. In competitive settings, such as two-player zero-sum games, this problem becomes even more complicated, as the agents’ strategies depend not only on their own rewards but also on their opponents’ strategies (\citealp{pmlr-v80-wang18d}; \citealp{savas2019}; \citealp{wei2021lastiterateconvergencedecentralizedoptimistic}). These challenges further motivate the study of inverse game theory (\citealp{LinIRLZeroSumGames}; \citealp{yu2019multiagentadversarialinversereinforcement}).

From a practical perspective, inferring the reward functions in competitive games has wide-ranging applications in economics, cyber security, robotics, and autonomous systems (\citealp{Ng2000invRL}; \citealp{Ziebart2008MaximumEI}). For particular examples, recovering reward functions behind players' actions in adversarial settings helps optimize resource allocation in cyber security (\citealp{miehling2018}), model strategic interactions in economic markets (\citealp{CHOW20152}), or design better AI systems for competitive tasks (\citealp{huang2019robot}). 

Meanwhile, recovering reward functions in competitive games raises several ubiquitous challenges: (i) Inverse problems are inherently ill-posed (\citealp{ahuja2001invop}; \citealp{yu2019multiagentadversarialinversereinforcement}), as multiple reward functions can potentially lead to the same optimal strategy and equilibrium solutions. Thus, a well-designed algorithm should not merely recover a single reward function but instead identify the entire set of feasible reward functions (\citealp{metelli21a}; \citealp{lindner2022active}; \citealp{metelli23a}). (ii) In the offline setting (\citealp{jarboui2021offlineinversereinforcementlearning}), insufficient dataset coverage  raises some significant challenges. Specifically, observed strategies may fail to comprehensively cover the state-action space, making it challenging to ensure robust reward function recovery. 
These challenges are further exacerbated in Markov games (\citealp{LITTMAN1994157}), where agents’ strategies evolve dynamically over time, adding complexity to both reward identifiability and estimation.

\subsection{Motivating Applications}
\paragraph{Economics and Marketing.} In competitive markets, firms often adjust prices, advertising strategies, or product features in response to rivals’ actions. This strategic interaction can be modeled as a two-player zero-sum or general-sum game, where each firm’s policy reflects a trade-off between profit maximization and risk/uncertainty. Recent studies (\citealp{bichler2025algorithmicpricingalgorithmiccollusion}) highlight how such systems—especially in repeated pricing environments—can implicitly learn to maintain supra‑competitive prices even without explicit coordination. Our inverse game theory framework provides a valuable perspective here. By observing how pricing algorithms respond to competitors over time, we  infer the underlying reward structures—such as profit priorities, market positioning, or risk aversion—that drive these strategies. Once we estimate the reward function, we can simulate alternative market scenarios, audit pricing behavior, or design fairer algorithmic policies grounded in empirical behavior rather than assumptions.

\paragraph{Cybersecurity.} Another important application of inverse game theory is in cybersecurity, where defenders must anticipate and counter strategic attacks on critical systems. For example, \cite{miehling2018} model this interaction as a sequential decision process in which the defender must infer an attacker’s hidden strategy via noisy security alerts, and respond  to block exploits and minimize damage. In such settings, our framework for inverse game theory infers the defender’s underlying objectives directly from observed attacker–defender interactions, without assuming they are known in advance. This allows security systems to adapt their defensive strategies based on actual observed behavior, improving resilience against evolving and unpredictable threats. 

\paragraph{Operations Research.} In logistics and transportation networks, multiple service providers—such as carriers or delivery companies—often compete over shared infrastructure, making strategic routing decisions that influence overall network flows. \citealp{Castelli2004} study such a case where two players control overlapping portions of a freight transportation network, each pursuing different objectives—one minimizing transportation costs, the other maximizing profit from traffic through their controlled routes. Using a bilevel optimization framework, they analyze equilibrium conditions and apply the model to international freight traffic through Europe. Our approach complements this by enabling the inverse problem: given observed routing patterns under such competitive dynamics, we  recover the underlying reward functions that explain each player's strategic behavior. By solving the inverse problem, planners and analysts can identify the underlying incentives driving competitive behavior in transportation networks, which enables better policy design and evaluation.

\subsection{Major Contributions}
We propose a unified framework for inverse game problems that addresses the identifiability and estimation of reward functions in competitive games in both static and dynamic settings. In particular, our major contribution is four-fold:

\begin{itemize}
\item Identifiability of reward functions: We study the identifiability problem using the quantal response equilibrium (QRE) under a linear assumption. Specifically, we derive the conditions when the reward function's parameters are identifiable,  and characterize the feasible set when parameters are not uniquely identifiable.

\item Algorithm for reward estimation: Building upon the identifiability results, we propose a new algorithm that efficiently estimates reward functions by constructing confidence sets to capture all feasible reward parameters. 

\item Extension to Markov games: We extend our framework to entropy-regularized Markov games, combining reward recovery with transition kernel estimation to handle dynamic settings. This approach is designed to be sample-efficient and adaptable, incorporating methods like Maximum Likelihood Estimation (MLE). 

\item Theoretical and empirical validation: We provide strong theoretical guarantees to establish the reliability and efficiency of our algorithm. Additionally, we conduct extensive numerical experiments to demonstrate the effectiveness of our framework in accurately recovering reward functions across various competitive scenarios.
\end{itemize}

\subsection{Related Work}
\paragraph{Zero-Sum Markov Games.} The zero-sum Markov game (\citealp{Shapley1953StochasticG}; \citealp{pmlr-v125-xie20a}; \citealp{cen2023fastpolicyextragradientmethods}; \citealp{Kalogiannis2023}) models the competitive interactions between two players in dynamic environments. The solution typically focuses on finding equilibrium strategies (\citealp{nash1950noncooper}; \citealp{MCKELVEY1956}; \citealp{pmlr-v125-xie20a}) where neither player can unilaterally improve their outcome. With a primary focus on learning in a sample-efficient manner, learning algorithms are proposed, including policy-based methods (\citealp{cen2021fastglobalconvergencenatural}; \citealp{wei2021lastiterateconvergencedecentralizedoptimistic}; \citealp{pmlr-v151-zhao22b}; 
\citealp{cen2023fastpolicyextragradientmethods}) and value-based methods (\citealp{pmlr-v125-xie20a}; \citealp{chen2022sample}; \citealp{Kalogiannis2023}). A related but distinct line of work studies regret-gap objectives for multi-agent imitation learning in Markov games \citep{tang2024multi}, where the goal is policy imitation rather than reward recovery.

\paragraph{Inverse Optimization and Inverse Reinforcement Learning (IRL).} Inverse optimization (\citealp{ahuja2001invop}; \citealp{chan2022inverseoptimizationtheoryapplications}; \citealp{ahmadi2023inverselearningsolvingpartially}) reverses the traditional optimization process by taking observed decisions as input to infer an objective function (\citealp{ahuja2001invop}; \citealp{nourollahi2018}) and constraints (\citealp{chan2019inverseoptimizationrecoveryconstraint}; \citealp{GHOBADI2021829}) that make these decisions approximately or exactly optimal. In practice, inverse optimization offers a powerful framework for modeling decision-making in complex systems across fields like marketing (\citealp{CHOW20152}; \citealp{Vatandoust2023}), operations research (\citealp{luce2005}; \citealp{Agarwal2010}; \citealp{yu2021learningriskpreferencesinvestment}), and machine learning (\citealp{Konstantakopoulos2017robustinvop}; \citealp{dong2018onlineinvop}; \citealp{delong2019deepinv}).

Compared with traditional RL, inverse reinforcement learning (\citealp{Ng2000invRL}; \citealp{Ziebart2008MaximumEI}; \citealp{pmlr-v51-herman16}; \citealp{wulfmeier2016maximumentropydeepinverse}; \citealp{arora2020surveyinversereinforcementlearning}) focuses on inferring the reward function based on the observed behavior or strategy of agents and experts, which is crucial for understanding various decision-making processes, from single-agent processes (\citealp{boularias11a}; \citealp{pmlr-v51-herman16}; \citealp{fu2018learningrobustrewardsadversarial}) to competitive or cooperative games (\citealp{vorobeychik2007,waugh2013computational,ling2018gameplayingendtoendlearning,pmlr-v80-wang18d,Wu2024invstackelberg}). A popular approach within the field of IRL is the Maximum Entropy IRL (\citealp{Ziebart2008MaximumEI};  \citealp{Ziebart2010ModelingPA}; \citealp{wulfmeier2016maximumentropydeepinverse}; \citealp{Snoswell_2020}), which is based on the principle of maximum entropy and is provably efficient in handling the uncertainty of agent behaviors (\citealp{waugh2013computational,Snoswell_2020,gleave2022primermaximumcausalentropy}) and high-dimensional observations (\citealp{wulfmeier2016maximumentropydeepinverse}; \citealp{Snoswell_2020}; \citealp{SONG2022328}). Among these works, \citet{waugh2013computational} is particularly relevant to our setting, as it studies an inverse equilibrium problem in multi-agent games under linear utility features and uses maximum entropy to select predictive behavior from the set of rationalizable strategies. However, their formulation is based on regret-based rationality constraints and the inverse correlated equilibrium (ICE) polytope, with the primary goal of behavior prediction. In contrast, our work focuses on \textit{entropy-regularized zero-sum matrix games} and \textit{Markov games}, where the observed strategies are modeled as quantal response equilibria (QRE), leading to a different identification framework for payoff and reward recovery.

\paragraph{Entropy Regularization in RL and Games.} Entropy regularization has become a widely used technique in reinforcement learning (\citealp{sze2010RL}; \citealp{Ziebart2010ModelingPA}) and game theory (\citealp{savas2019}; \citealp{guan2021learningnashequilibriazerosum}; \citealp{cen2023fastpolicyextragradientmethods}). It is provably effective in addressing challenges such as the exploration–exploitation tradeoff (\citealp{haarnoja2018softactorcriticoffpolicymaximum}; \citealp{wang2019explorationversusexploitationreinforcement}; \citealp{pmlr-v97-ahmed19a}; \citealp{neu2017unifiedviewentropyregularizedmarkov}), algorithm robustness (\citealp{zhao2020maxentr}; \citealp{guo2021entropyregularizationmeanfield}), and convergence acceleration (\citealp{cen2021fastglobalconvergencenatural}; \citealp{cen2023fastpolicyextragradientmethods}; \citealp{zhan2023policymirrordescentregularized}).
Importantly, entropy regularization has also been shown to improve identifiability in inverse reinforcement learning (IRL) problems. Recent works in single-agent IRL, such as \citet{cao2021identifiability} and \citet{rolland2022identifiability}, leverage entropy-regularized policies to transform ill-posed IRL problems into identifiable ones under mild assumptions. Our work builds on this insight by extending it to competitive multi-agent settings, where identifiability becomes even more subtle due to strategic interactions. 


\vspace{4pt}

{\noindent\bf Paper Organization.} The rest of the paper is organized as follows. In Section \ref{sec:2}, we develop the framework of inverse game theory for entropy-regularized zero-sum matrix games, providing identification conditions and confidence set constructions. In Section \ref{sec:3}, we extend this framework to sequential decision-making problems, focusing on entropy-regularized zero-sum Markov games. We present both theoretical guarantees and algorithmic implementations in this dynamic setting. In Section \ref{sec:4}, we conduct numerical experiments to validate our theoretical findings and demonstrate the effectiveness of our approach in recovering reward functions and predicting agent behavior. We conclude with a discussion of related work and future research directions in Section \ref{sec:5}.



\vspace{4pt}

{\noindent\bf Notations.}  Throughout this paper, we denote the set of $\{1,2,\cdots,n\}$ as $[n]$ for any positive integer~$n$. For two positive sequences $\{a_n\}_{n\in\mathbb{N}}$ and  $\{b_n\}_{n\in\mathbb{N}}$, we write $a_n = \mathcal{O}(b_n)$ or $a_n\lesssim b_n$ if there exists a positive constant $C$ such that $a_n\leq C\cdot b_n$. For any integer $d$, we denote the $d$-dimensional Euclidean space by $\mathbb{R}^d$, with inner product $\langle x,y\rangle=x^\top y$ and induced norm $\lVert x \rVert = \sqrt{\langle x,x\rangle}$. For any matrix $A = (a_{ij})$, the Frobenius norm and the operator norm (or spectral norm) of $A$ are $\lVert A\rVert_\mathrm{F} = (\sum_{i,j}a_{ij}^2)^{1/2}$ and $\lVert A\rVert_{\text{op}} = \sigma_1(A)$, where $\sigma_1(A)$ stands for the largest singular value of $A$. For any square matrix $A = (a_{ij})$, denote its trace by $\text{tr}(A) = \sum_i a_{ii}$. For a discrete space $\mathcal{A}$, denote by $\Delta(\mathcal{A})$ the set of all probability distributions on $\mathcal{A}$. For any two distributions $\mu,\wt\mu\in\Delta(\mathcal{A})$, we write $\mathrm{TV}(\mu,\wt\mu)=\frac{1}{2}\Vert\mu-\wt\mu\Vert_1$ for the total variation distance between them.
\section{Entropy-Regularized Zero-Sum Matrix Games}\label{sec:2}
We first formulate the inverse game problem for entropy-regularized two-player zero-sum matrix games. Next, we consider the identifiability problem of payoff matrices under the linear parametric assumption, and derive a sufficient and necessary condition for strong identifiability. Finally, we propose new methods to recover the identifiable sets and payoff matrices.

\subsection{Problem Formulation}\label{sec:2.1}
A two-player zero-sum matrix game is specified by a triple $(\mathcal{A}, \mathcal{B}, Q)$,
where $\mathcal{A} = \{1,2,\cdots,m\}$ and $\mathcal{B} = \{1,2,\cdots,n\}$
are two finite sets of feasible actions that players $i \in \{1, 2\}$ can take, respectively, and $Q:\mathcal{A}\times\mathcal{B}\to\bbR$ is the payoff function. Since both sets $\mathcal{A}$ and $\mathcal{B}$ are finite, we consider $Q=(Q(a,b))_{a\in\mathcal{A},b\in\mathcal{B}}$ as a matrix. Then, we formulate the zero-sum game  as the following min-max optimization problem that
\[\max_{\mu}\min_{\nu}\mu^\top Q\nu,\]
where $\mu\in\Delta(\mathcal{A})$ and $\nu\in\Delta(\mathcal{B})$ are policies for each player, and $Q = (Q(a,b))_{a\in\mathcal{A},b\in\mathcal{B}}\in\mathbb{R}^{m\times n}$ denotes the payoff matrix. Note that the solution of this optimization problem achieves the \textit{Nash equilibrium} (\citealp{nash1950noncooper}), where both agents play the best response against each other.

\paragraph{Entropy-Regularized Two-Player Zero-Sum Matrix Game.} 
We focus on the entropy-regularized matrix game. Formally, this amounts to solving the following matrix game with entropy regularization (\citealp{mert2016RKreg}) that
\begin{equation}
    \max_{\mu}\min_{\nu}\mu^\top Q\nu+\eta^{-1}\mathcal{H}(\mu)-\eta^{-1}\mathcal{H}(\nu),\label{entrmgobj}
\end{equation}
where $\eta>0$ is the regularization parameter, and $$\mathcal{H}(\pi)=-\sum_{i}\pi_i\log(\pi_i)$$ denotes the Shannon entropy (\citealp{shannon1948}) of a discrete probability distribution $\pi$. With the regularization terms, the objective function (\ref{entrmgobj}) is convex with respect to~$\nu$ and concave with respect to $\mu$. By the von-Neumann minimax theorem (\citealp{Neumann1928ZurTD}), there exists a solution $(\mu^*,\nu^*)$ to this min-max problem, denoted as the quantal
response equilibrium (QRE) (\citealp{MCKELVEY1956}), which satisfies the following fixed point equations that
\begin{equation}
    \begin{cases}
        \displaystyle\mu^*(a) = \frac{e^{\eta Q(a,\cdot)\nu^*}}{\sum_{a\in\mathcal{A}}e^{\eta Q(a,\cdot)\nu^*}},&\text{for all }a\in\mathcal{A},\vspace{0.2cm}\\
        \displaystyle\nu^*(b) = \frac{e^{-\eta  Q(\cdot,b)^\top\mu^*}}{\sum_{b\in\mathcal{B}}e^{-\eta  Q(\cdot,b)^\top\mu^*}},&\text{for all }b\in\mathcal{B}.
    \end{cases}
    \label{qreconstraints}
\end{equation}

Our goal is to study the inverse game theory for this entropy-regularized zero-sum matrix game. To elaborate, we observe sample strategy pairs $(a^k,b^k)\stackrel{\text{i.i.d.}}{\sim}(\mu^*,\nu^*)$ following the QRE, and we aim to recover all the feasible payoff functions $Q(\cdot,\cdot)$.

To derive inverse game theory, it is crucial to first study the identifiability of the payoff matrix. In the next subsection, we study the strong identifiability problem under the linear structure assumption, where we  derive a sufficient and necessary condition under which there exists a unique payoff matrix that satisfies the QRE constraint. We  further generalize the analysis to the partially identifiable case in Section~\ref{sec:piden}, where there is potentially a set of payoff matrices that satisfy the constraint.

\subsection{Strong Identifiability}\label{sec:siden}
We denote by $(\mu^*,\nu^*)$ the true QRE for two players, and assume that we have an estimator  $(\widehat{\mu},\widehat{\nu})$. For analytical tractability, we impose the following linear parametric assumption.
\begin{assumption}[Linear payoff functions]\label{a3.1}
There exists a vector-valued function $\phi:\mathcal{A}\times\mathcal{B}\rightarrow \mathbb{R}^d$ and a vector $\theta^*\in\mathbb{R}^d$ such that $\Vert\theta^*\Vert^2\leq M$ for some $M>0$, and $Q(a,b) = \langle \phi(a,b),\theta^*\rangle$ for all $(a,b)\in\mathcal{A}\times\mathcal{B}$.
\end{assumption}

To estimate the payoff matrix $Q$ from the observed data under this assumption, we essentially need to estimate the parameter $\theta^*$. Under the linear assumption, the non-linear equation system \eqref{qreconstraints} is equivalent to the linear system that
    \begin{equation*}
    \begin{cases}
        \displaystyle\left(Q(a,\cdot)-Q(1,\cdot)\right)\nu^* =\eta^{-1}\log\frac{\mu^*(a)}{\mu^*(1)},\vspace{0.2cm}\\
        \displaystyle\left(Q(\cdot,b)-Q(\cdot,1)\right)^\top\mu^*= -\eta^{-1}\log\frac{\nu^*(b)}{\nu^*(1)},
    \end{cases}
\end{equation*}
where $a\in\mathcal{A}\backslash\{1\}$ and $b\in\mathcal{B}\backslash\{1\}$. Note that there are $m+n-2$ linear equations in total. Under Assumption \ref{a3.1}, these linear equations are equivalent to 
\begin{equation*}
    \begin{cases}
        \displaystyle\langle\left(\phi(a,\cdot)-\phi(1,\cdot)\right)\nu^*,\theta\rangle = \eta^{-1}\log\frac{\mu^*(a)}{\mu^*(1)}, \vspace{0.2cm}\\
        \displaystyle\langle\left(\phi(\cdot,b)-\phi(\cdot,1)\right)^\top\mu^*,\theta\rangle= -\eta^{-1}\log\frac{\nu^*(b)}{\nu^*(1)},
    \end{cases}
\end{equation*}
where  $\left(\phi(a,\cdot)-\phi(1,\cdot)\right)\nu^*$ and $\left(\phi(\cdot,b)-\phi(\cdot,1)\right)^\top\mu^*$ are $d$-dimensional vectors. To simplify the notations, we let
\begin{equation*}
\begin{aligned}
A(\nu)&=(\left(\phi(a,\cdot)-\phi(1,\cdot)\right)\nu)_{a=2}^m\in\mathbb{R}^{(m-1)\times d},\quad B(\mu)=(\left(\phi(\cdot,b)-\phi(\cdot,1)\right)^\top\mu)_{b=2}^n\in\mathbb{R}^{(n-1)\times d},
\end{aligned}
\end{equation*}
and 
\begin{equation*}
 \begin{aligned}
c(\mu)&= (\log(\mu(a)/\mu(1))/\eta)_{a=2}^m\in\mathbb{R}^{m-1},\quad d(\nu)= (-\log(\nu(b)/\nu(1))/\eta)_{b=2}^n\in\mathbb{R}^{n-1}
\end{aligned}   
\end{equation*}
We then write the linear system as
\begin{equation}\label{e3.1}
    \begin{bmatrix}
    A(\nu^*)\\
    B(\mu^*)
\end{bmatrix}\theta = \begin{bmatrix}
    c(\mu^*)\\
    d(\nu^*)
\end{bmatrix}.
\end{equation}
This linear system has $m+n-2$ constraints, where $\theta\in\bbR^d$. Thus, if $d\leq m+n-2$ and the linear constraints are of full rank, there is at most one solution of the above linear equations. We show that this is a sufficient and necessary condition for strong identifiability for the parameter $\theta$ in the next proposition.

\begin{proposition}[Sufficient and necessary condition for strong identifiability]\label{p3.2} Under Assumption~\ref{a3.1}, there is a unique $\theta\in\mathbb{R}^d$ such that $Q(a,b) = \langle\phi(a,b),\theta\rangle$ (i.e. $\theta = \theta^*$) for all $(a,b)\in\mathcal{A}\times\mathcal{B}$ if and only if the QRE satisfies the rank condition that
	\begin{equation}\label{eq3.2}
		\text{rank}\left(\begin{bmatrix}
			A(\nu^*) \\ B(\mu^*)
		\end{bmatrix}\right) = d.
	\end{equation}
\end{proposition}
\proof{Proof.}
See Appendix \ref{proofprop1} for the complete proof.
\endproof
The rank condition in \eqref{eq3.2} implies that the observed QRE provides sufficiently diverse information about the game structure. Intuitively, if the players' strategies span a sufficiently rich space of responses, the underlying payoff function can be uniquely determined. In the remainder of this subsection, we assume that the rank condition in Proposition \ref{p3.2} holds, so that the game is strongly identifiable. Then, to estimate $\theta^*$ in the offline setting, we propose a two-step method that
\begin{enumerate}
    \item Estimate the QRE $(\mu^*,\nu^*)$ from the observed data with some consistent method and obtain estimator $(\widehat{\mu},\widehat{\nu})$.
    \item Following \eqref{e3.1}, we estimate $\theta^*$ using the least-square estimator  $\widehat{\theta}$. Specifically, we let
\begin{equation}\label{e3.2}
    \widehat{\theta}= \mathop{\arg\min}_{\theta\in\mathbb{R}^d}\left\lVert \begin{bmatrix}
    A(\widehat{\nu})\\
    B(\widehat{\mu})
\end{bmatrix}\theta - \begin{bmatrix}
    c(\widehat{\mu})\\
    d(\widehat{\nu})
\end{bmatrix}\right\rVert^2.
\end{equation}
\end{enumerate}
Note that in the first step, we purposely do not specify the method for estimating $\mu^*$ and $\nu^*$, and we present a general bound in the next theorem. For the second step, intuitively, if the sample size is  large and the QRE estimation errors, denoted as $\text{TV}(\widehat{\mu},\mu^*)$ and $\text{TV}(\widehat{\nu},\nu^*)$, 
are close to zero, we have that the coefficient matrix in \eqref{e3.2} is of full column rank with probability goes to 1. In this case, we have a closed-form solution for problem \eqref{e3.2} that
\begin{equation}\label{e3.3}
\begin{aligned}
    \widehat{\theta} &= \left(\begin{bmatrix}
    A(\widehat{\nu})\\
    B(\widehat{\mu})
\end{bmatrix}^\top\begin{bmatrix}
    A(\widehat{\nu})\\
    B(\widehat{\mu})
\end{bmatrix}\right)^{-1}\begin{bmatrix}
    A(\widehat{\nu})\\
    B(\widehat{\mu})
\end{bmatrix}^\top\begin{bmatrix}
    c(\widehat{\mu})\\
    d(\widehat{\nu})
\end{bmatrix}\\
&=\left(A(\widehat{\nu})^\top A(\widehat{\nu})+B(\widehat{\mu})^\top B(\widehat{\mu})\right)^{-1}\left[
    A(\widehat{\nu})^\top c(\widehat{\mu})+B(\widehat{\mu})^\top d(\widehat{\nu})
\right].
\end{aligned}
\end{equation}

Next, we derive the estimation error of the two-step method. Namely, given some bounds for $\text{TV}(\widehat{\mu},\mu^*)$ and $\text{TV}(\widehat{\nu},\nu^*)$, we bound the Euclidean distance between $\widehat{\theta}$ and $\theta^*$.

\begin{theorem}[Parameter estimation error]\label{mxe}
Let $\epsilon_1$ and $\epsilon_2$ be two small positive numbers satisfying $\epsilon_1<\min_{a\in[m]}\mu^*(a)$ and $\epsilon_2<\min_{b\in[n]}\nu^*(b)$.
Under Assumption \ref{a3.1} and the rank condition \eqref{eq3.2}, if $(\widehat \mu, \widehat\nu)$ satisfies $\text{TV}(\widehat{\mu},\mu^*)\leq \epsilon_1/2$ and $\text{TV}(\widehat{\nu},\nu^*)\leq \epsilon_2/2$, then $\widehat\theta$ constructed by  \eqref{e3.2} satisfies
\begin{equation*}
\begin{aligned}
\lVert\widehat{\theta}-\theta^*\rVert^2&\lesssim \epsilon_1^2\cdot\left(1+m\cdot(\epsilon_2^2+1)\right)+\epsilon_2^2\cdot\left(1+n\cdot(\epsilon_1^2+1)\right).
\end{aligned}
\end{equation*}
\end{theorem}

\proof{Proof.}
    See Appendix~\ref{pmxe} for the complete proof.
\endproof

Theorem \ref{mxe} provides an upper bound on the parameter estimation error $\Vert\wh\theta-\theta^*\Vert^2$, which scales quadratically with the errors in constructing the equilibrium distributions $\wh\mu$ and $\wh\nu$. Since $\theta^*$ is uniquely determined given the true QRE $(\mu^*,\nu^*)$, any deviation in $\wh\theta$ from $\theta^*$ arises due to the randomness in data collection and the estimation error. Moreover, the presence of $m$ and $n$ in the bound characterizes how the estimation error is shaped by the complexity of the action spaces: when the number of actions on either side increases, the corresponding terms involving $\epsilon_1$ and $\epsilon_2$ are amplified, which reflects the additional complexity introduced by larger action sets.
 
Next, we present the finite sample error bound of estimating the payoff matrix $Q$. In the two-step method, given a dataset of agent actions following the true QRE, we first construct an estimator for the QRE. Thus, the sample complexity depends on the convergence rate of the QRE estimator. Here, an intuitive choice for the QRE estimator is the frequency estimator:
\begin{equation}
\widehat{\mu}(a) = \frac{1}{N}\sum_{k=1}^N\mathbf{1}_{\{a^k=a\}},\quad \widehat{\nu}(b)=\frac{1}{N}\sum_{k=1}^N\mathbf{1}_{\{b^k=b\}}.\label{freqQRE}
\end{equation}
In the next theorem, we leverage the estimation error of the QRE frequency estimator and the error bound in Theorem~\ref{mxe} to obtain the estimation error of the payoff matrix.
\begin{theorem}[Finite sample error bound]\label{mxs}
    Given $N$ samples $\{(a^k,b^k)\}_{k\in[N]}$ following the true QRE $(\mu^*,\nu^*)$, we let $\widehat{\mu},\widehat{\nu}$ be the frequency estimator \eqref{freqQRE}. Fix any $\delta\in(0,1)$. Then with probability at least $1-\delta$,  the estimation error bound of the payoff matrix satisfies
    \begin{equation*}
    \lVert\widehat{Q}-Q\rVert_F^2\lesssim\mathcal{O}\left(\frac{m^2+n^2+(m+n)\log(1/\delta)}{N}\right).
    \end{equation*}
\end{theorem}
\proof{Proof.}
    See Appendix \ref{pmxs} for the complete proof.
\endproof

Theorem \ref{mxs} provides a probabilistic guarantee for the accuracy of the reconstructed payoff matrix $\wh{Q}$ in a finite-sample setting. The bound explicitly depends on the sample size $N$, the action space dimensions $m,n$, and the confidence parameter $\delta$. The estimation error decreases at a rate of $\mathcal{O}(1/N)$, which is consistent with the standard empirical process result of the frequency estimator~\eqref{freqQRE} (\citealp{Vaart_1998}). As the sample size $N$ increases, the errors of $\wh\mu$ and $\wh\nu$ decrease, leading to a more accurate reconstruction of the reward $Q^*$.

On the other hand, the bound grows with the action space size in terms of $m^2+n^2$, indicating that larger action spaces require more samples to achieve the same estimation accuracy. This suggests that in problems with large action spaces, the frequency estimator \eqref{freqQRE} may become inefficient, and that adopting alternative estimators, such as maximum likelihood methods for parametric models, could lead to better performance. We will later demonstrate the result about maximum likelihood methods in Section~\ref{sec:3.4}.

    

\subsection{Partial Identifiability}\label{sec:piden}
 The rank condition \eqref{eq3.2} highlights the importance of exploration in equilibrium play. If the observed strategies fail to sufficiently explore the action space, certain directions in the parameter space may remain underdetermined, leading to a partial identifiability rather than strong identifiability. In practical scenarios, a higher dimension $d$ for the parameter $\theta$ generally enhances the model’s representation power, allowing the kernel map $\phi$ to capture more complex payoff structures. However, increasing $d$ also raises the risk of violating the rank condition \eqref{eq3.2}, as the available data may not provide enough independent constraints to fully determine all parameters. Thus, studying the case of partial identifiability is crucial, as it reflects realistic scenarios where limited exploration or insufficient data prevents full recovery of the underlying payoff structure.

In case the rank condition \eqref{eq3.2} does not hold, the identifiable set of parameters $\theta\in\bbR^d$ that satisfy the QRE constraints (\ref{e3.1}) is infinite and unbounded, which is impractical for estimation or interpretation. Recall that in Assumption \eqref{a3.1}, we define the \textit{feasible set} $\Theta\subset\mathbb{R}^d$ as
\begin{equation*}
\Theta= \left\{\theta\in\bbR^d:\begin{bmatrix}
    A(\nu^*)\\
    B(\mu^*)
\end{bmatrix}\theta = \begin{bmatrix}
    c(\mu^*)\\
    d(\nu^*)
\end{bmatrix},\Vert\theta\Vert^2\leq M\right\}.
\end{equation*}
Here we impose the constraint $\Vert\theta\Vert^2\leq M$ to maintain stable estimation and avoid overfitting.

When the true $\theta^*$ is partially identified, instead of providing a single estimator, we construct a confidence set that contains the feasible set with high probability. Specifically, given $N$ strategy pair data following the true QRE, we first estimate the QRE using the observed data by frequency estimators $\widehat{\mu}$ and $\widehat{\nu}$ in \eqref{freqQRE}. Next, we select a threshold $\kappa>0$, and construct the confidence set by
\begin{equation}
\wh{\Theta}=\left\{\theta:\left\lVert \begin{bmatrix}
    A(\widehat{\nu})\\
    B(\widehat{\mu})
\end{bmatrix}\theta -\begin{bmatrix}
    c(\widehat{\mu})\\
    d(\widehat{\nu})
\end{bmatrix}\right\rVert^2\leq\kappa,\\
\Vert\theta\Vert^2\leq M\right\}.\label{thetaconfset}
\end{equation}
This set quantifies the uncertainty in estimating $\theta^*$ due to limited samples. Finally, to recover the feasible payoff functions, we  compute $\wh{Q}(a,b)=\phi(a,b)^\top\wh{\theta}$ for all $\wh\theta\in\wh\Theta$ following the linear assumption. We summarize the procedure in Algorithm \hyperref[algpff]{1}.

\begin{algorithm}[H]\label{algpff}
\caption{Learning payoff from actions}
\begin{algorithmic}[1]
\Require{Dataset $\mathcal{D} = \{(a^k,b^k)\}_{k\in[N]}$, kernel $\phi(\cdot,\cdot)$, entropy regularization parameter $\eta$, threshold parameter $\kappa$, ridge regularization term $\lambda$.}
\For {$(a,b)\in\mathcal{A}\times\mathcal{B}$}
\State Compute the empirical QRE by \eqref{freqQRE};
\State Construct the confidence set $\wh\Theta$ by \eqref{thetaconfset};
\State Compute the set of feasible payoff matrices
\begin{equation*}
\mathcal{Q} = \left\{\wh{Q}(a,b)=\langle\phi(a,b),\wh\theta\rangle: \ \wh{\theta}\in\wh{\Theta}\right\}.
\end{equation*}
\EndFor
\end{algorithmic}
\end{algorithm}

We demonstrate the effectiveness of Algorithm \hyperref[algpff]{1} by showing it indeed constructs a valid confidence set. In particular, we show that the confidence set $\wh{\Theta}_N$ converges to the feasible set $\Theta$ when the sample size $N$ increases. To facilitate our discussion, we employ the Hausdorff distance $d_H(\cdot,\cdot)$ defined below to quantify the difference between two sets.
\begin{definition}[Hausdorff distance]
    Let $(\mathcal{M},d)$ be a metric space. For each pair of non-empty subsets $X\subset\mathcal{M}$ and $Y\subset \mathcal{M}$, the \textit{Hausdorff distance} between $X$ and $Y$ is 
    \[d_H(X,Y)=\max\left\{\sup_{x\in X}d(x,Y),\sup_{y\in Y}d(X,y)\right\}.\]
\end{definition}

The key to approximating the identifiable set $\Theta$ is selecting an appropriate threshold $\kappa$ that ensures the confidence set $\wh\Theta_N$ is ``similar" to $\Theta$. To be concrete, we endow the space $\bbR^d$ with the Euclidean metric $d$, and compute the Hausdorff distance $d_H$ with respect to this metric. We choose $\kappa$ properly to ensure the convergence that 
\[\lim_{N\rightarrow\infty} d_H(\Theta,\widehat{\Theta}_N) = 0.\]
In particular, when $\theta^*$ is not uniquely determined due to rank deficiency in \eqref{eq3.2}, the confidence set should fully capture the feasible set, and does not collapse to a single point. The next lemma formalizes this intuition.

\begin{lemma}[Construction error]\label{mxc}
Let $\epsilon_1,\epsilon_2>0$ satisfy $\epsilon_1<\min_{a\in[m]}\mu^*(a)$ and $\epsilon_2<\min_{b\in[n]}\nu^*(b)$.
Let the normalized feature matrices $\Phi_1$ and $\Phi_2$ be
\begin{equation*}
\begin{aligned}
\Phi_1&= (\left(\phi(i,\cdot)-\phi(1,\cdot)\right))_{i=2}^m\in\mathbb{R}^{d\times (m-1)n},\quad \Phi_2= (\left(\phi(\cdot,j)-\phi(\cdot,1)\right))_{j=2}^n\in\mathbb{R}^{d\times (n-1)m},
\end{aligned}
\end{equation*} 
where we construct normalized features of the two players' actions by  comparing with their baseline actions, which we fix as action 1 for each player.
Recall that the parameter $M$ is specified in 
Assumption \ref{a3.1}. Under this assumption, we define 
\begin{equation}
\begin{aligned}
\kappa&= 2\left(M\lVert \Phi_1\rVert_{\mathrm{op}}^2+\frac{n}{\eta^2\left(\min_{j\in[n]}\nu_j-\epsilon_2\right)^2}\right)\epsilon_2^2+2\left(M\lVert\Phi_2\rVert_{\mathrm{op}}^2+\frac{m}{\eta^2\left(\min_{i\in[m]}\mu_i-\epsilon_1\right)^2}\right)\epsilon_1^2.
\end{aligned}
\end{equation}Then, the confidence set $\widehat \Theta $ constructed in \eqref{thetaconfset} with parameters $\kappa $ and $M$ satisfies 
$\Theta\subseteq\widehat{\Theta}$. That is, $\widehat \Theta $ contains all feasible parameters.  Moreover, 
when $\epsilon_1,\epsilon_2$ are sufficiently small, we have 
    \begin{equation}
        d_H(\Theta,\widehat{\Theta})\lesssim\sqrt{\kappa}.\label{hausdorffbound}
    \end{equation}
\end{lemma}
\proof{Proof.}
    See Appendix \ref{pmxc} for the complete proof.
\endproof
Lemma \ref{mxc} ensures that the constructed confidence set $\wh\Theta$  fully captures the feasible set $\Theta$ that every valid parameter satisfying the QRE constraints is included. The bound \eqref{hausdorffbound} on the Hausdorff distance quantifies how closely $\wh\Theta$ approximates $\Theta$. As the sample size increases, $\epsilon_1$ and $\epsilon_2$ decrease, and the threshold $\kappa$ decreases. Also,  $\wh\Theta$ shrinks towards $\Theta$ while still fully capturing it. Furthermore, the dependence of $\kappa$ on the TV errors $\epsilon_1$ and $\epsilon_2$ accommodates the uncertainty in QRE estimation, and prevents false exclusion of valid solutions.

Next, we  leverage Lemma \ref{mxc} to establish the consistency result for our confidence set when using the frequency-based QRE estimator \eqref{freqQRE}.
\begin{theorem}[Convergence of confidence set]\label{tmxc}
Under Assumption \ref{a3.1}, suppose that we observe $N$ samples $\{(a^k,b^k)\}_{k\in[N]}$ following the true QRE $(\mu^*,\nu^*)$, and set $(\widehat{\mu},\widehat{\nu})$ as the frequency estimator. Set the confidence set  $\wh{\Theta}_N$ as
\begin{equation*}
\wh{\Theta}_N=\left\{\theta:\left\lVert \begin{bmatrix}
A(\widehat{\nu})\\ B(\widehat{\mu})
\end{bmatrix}\theta -\begin{bmatrix}
    c(\widehat{\mu})\\
    d(\widehat{\nu})
\end{bmatrix}\right\rVert^2\leq \kappa_N,\Vert\theta\Vert\leq M\right\},
    \end{equation*}
where $\kappa_N = \mathcal{O}(N^{-1})$. Then $d_H(\Theta,\wh{\Theta}_N)\stackrel{\mathbb{P}}{\longrightarrow} 0$. In addition, with probability at least $1-\delta$, we have
\begin{equation}
d_H(\Theta,\widehat{\Theta}_N)\lesssim\sqrt{\frac{m^2+n^2+(m+n)\log(1/\delta)}{N}}.\label{confsampcomplexity}
\end{equation}
\end{theorem}
\proof{Proof.}
    The result follows immediately from Lemma \ref{mxc}. 
\endproof
\begin{remark}[Convergence of payoff function]
The convergence of the confidence set $\wh{\Theta}_N$ to $\Theta$ further implies the convergence of the associated payoff function $Q_\theta$. 
In particular, for any $\theta\in\wh{\Theta}_N$ there exists a $\theta^*\in\Theta$ such that 
\[
\vert Q_\theta - Q_{\theta^*}\vert \;\longrightarrow\; 0 \quad \text{as } N\to\infty,
\]
where the norm can be taken as, for example, the $\ell_\infty$ or Frobenius norm on the payoff matrix. Moreover, the finite-sample bound in \eqref{confsampcomplexity} implies a bound on the payoff estimation error that
\[
\vert Q_\theta - Q_{\theta^*}\vert \;\lesssim\; 
\sqrt{\frac{m^2+n^2+(m+n)\log(1/\delta)}{N}}
\]
with probability at least $1-\delta$, since the feature map $\phi$ is bounded. Thus, our inference procedure yields not only consistency of the parameter set but also recovers  the payoff function itself.
\end{remark}

Theorem \ref{tmxc} establishes the asymptotic consistency of our confidence set $\wh\Theta_N$ in the finite-sample setting, showing that it converges to the true feasible set $\Theta$ as the number of observed samples increases. The finite-sample bound \eqref{confsampcomplexity} demonstrates that the estimation error decreases at the rate of $\mathcal{O}(N^{-1/2})$, which matches the standard concentration rate for empirical frequency estimators. The dependence on $m$ and $n$ highlights that larger action spaces require more samples for the same level of accuracy. 

\paragraph{Parameter Selection.} 
When the rank condition \eqref{eq3.2} does not hold, the feasible set $\Theta$ contains infinitely many solutions. To avoid instability, one natural approach is to select the minimum-norm solution. We defer the detailed discussion of this parameter selection procedure and its convergence properties to Appendix~\ref{sec:2.4}.

\section{Entropy-Regularized Zero-Sum Markov Games}\label{sec:3}
We extend the method  developed in Section \ref{sec:2} to handle the more complicated inverse game for entropy-regularized two-player zero-sum Markov games. To elaborate, assume that both players take actions following the optimal QRE $(\mu^*,\nu^*)$, and we only observe a collection of states and actions $\{(s_h^t,a_h^t,b_h^t)\}_{h\in[H],t\in[T]}$, where $H$ is the horizon, and $T$ is the number of episodes. We remark that the rewards $r_h^t$ and the transition kernels $\mathbb{P}_h$ are unknown.

\subsection{Background and Problem Formulation}\label{sec:3.1}
{Markov games (\citealp{zachrisson1964markov}) naturally arise in many operations research problems where agents repeatedly interact and the system evolves stochastically. In transportation systems (\citealp{Agarwal2010}, \citealp{Castelli2004}), drivers (or firms) repeatedly make routing choices while congestion builds dynamically. Each stage’s state (traffic density, travel time) evolves with their actions, forming a Markov congestion game. Observing traffic flows and route choices allows us to infer the implicit reward functions (e.g., costs, delays, toll sensitivity) of agents. Similarly, in competitive supply chains (\citealp{cachon2006game}), retailers interact over multiple periods by setting prices and inventory decisions. The state encodes inventory levels and demand, while actions influence both profits and future market dynamics. Modeling this as a Markov game allows us to recover the underlying reward structures from observed sales episodes and ensure that the reconstructed quantal response equilibrium strategies align with empirical market behavior.}

We briefly review the setting of a two-player zero-sum Markov game (\citealp{LITTMAN1994157}), which is a framework that extends Markov decision processes (MDPs) to multi-agent settings. In this setting,  two players with opposing objectives interact in a shared environment. A two-player zero-sum simultaneous-move episodic Markov game is defined by a six-tuple $(\mathcal{S}, \mathcal{A}, \mathcal{B}, r, \mathbb{P}, H)$,
where 
\begin{itemize}\setlength\itemsep{0em}
\item $\mathcal{S}$ is the state space, with $\vert\mathcal{S}\vert=S$;
\item $\mathcal{A}$ and $\mathcal{B}$
are two finite sets of actions that players $i\in\{1,2\}$ can take respectively; 
\item $H$ is the length of time horizon, with $[H]=\{1,2,\cdots,H\}$;
\item $r = \{r_h\}_{h\in[H]}$ is the collection of reward functions; 
\item  $\mathbb{P} = \{\mathbb{P}_h\}_{h\in[H]}$ is the collection of transition kernels.
\end{itemize} 
At each time step $h \in [H]$,  players $1$ and $2$ simultaneously take actions $a_h \in \mathcal{A}$ and $b_h \in \mathcal{B}$ respectively upon
observing state $s_h\in \mathcal{S}$, and then player $1$ receives reward $r_h(s_h,a_h, b_h)\in[0,1]$, while player $2$ receives $-r_h(s_h,a_h,b_h)$. Namely, the gain of one player equals the loss of the other. The system then transitions to a new state $s^\prime\sim\mathbb{P}_h(\cdot|s, a, b)$ following the transition kernel $\mathbb{P}_h$. 

\paragraph{Entropy-regularized two-player zero-sum Markov game.} We focus on the two-player zero-sum Markov game with entropy regularization. With some slight abuse of notation, we denote by $(\mu,\nu)$  the strategies of two players, where $\mu = \{\mu_h\}_{h = 1}^H$ and $\nu = \{\nu_h\}_{h = 1}^H$. At step $h$, the entropy-regularized V-function (\citealp{haarnoja2018softactorcriticoffpolicymaximum}; \citealp{cen2021fastglobalconvergencenatural}) is 
\begin{equation*}
\begin{aligned}
&V_h^{\mu,\nu}(s) = \mathbb{E}\Biggl[\sum_{h^\prime = h}^H\gamma^{h^\prime-h}[r_{h^\prime}(s_{h^\prime},a_{h^\prime},b_{h^\prime})-\eta^{-1}\log{\mu_{h^\prime}}(a_{h^\prime}|s_{h^\prime})+\eta^{-1}\log{\nu_{h^\prime}}(b_{h^\prime}|s_{h^\prime})]\Bigg|s_h = s\Biggr],
\end{aligned}
\end{equation*}
where $\gamma\in(0,1]$ is the discount factor, and $\eta>0$ is the parameter of regularization. This value function measures the expected cumulative reward under policies $\mu$ and $\nu$, with additional terms that promote stochasticity and lead to the QRE. Following this definition, we define the entropy-regularized Q-function as
\begin{equation}\label{bel}
\begin{aligned}
&Q_h^{\mu,\nu}(s,a,b) = r_h(s,a,b)+\gamma\mathbb{E}_{s^\prime\sim\mathbb{P}_h(\cdot|s,a,b)}\left[V_{h+1}^{\mu,\nu}(s^\prime)\right].
\end{aligned}
\end{equation}
For notational simplicity, we denote by $Q_h^{\mu,\nu}(s)\in\mathbb{R}^{ m\times n}$ the collection of Q-functions at state~$s$, which is represented as a matrix $[Q_h^{\mu,\nu}(s,a,b)]_{(a,b)\in\mathcal{A}\times\mathcal{B}}$. Adopting this notation, we have
\begin{equation}\label{vrep}
\begin{aligned}
&V_h^{\mu,\nu}(s)= \mu_h(s)^{\top}Q_h^{\mu,\nu}(s)\nu_h(s) +\eta^{-1}\mathcal{H}(\mu_h(s))-\eta^{-1}\mathcal{H}(\nu_h(s)).
\end{aligned}
\end{equation}
Note that equations \eqref{bel} and \eqref{vrep} are known as Bellman equations for Markov games. In a zero-sum game, one player seeks to maximize the value function while the other player aims to minimize it. The minimax game value at state $s$ and the beginning stage is  
\[V^*_1(s) = \mathop{\max}_\mu\mathop{\min}_\nu V_1^{\mu,\nu}(s) = \mathop{\min}_\nu\mathop{\max}_\mu V_1^{\mu,\nu}(s),\]
and the corresponding minimax Q-function $Q^*(s,a,b)$ is 
\[Q^*_1(s,a,b) = r_1(s,a,b)+\gamma\mathbb{E}_{s^\prime\sim\mathbb{P}_1(\cdot|s,a,b)}\left[V_{2}^*(s^\prime)\right].\]
We then define the quantal response equilibrium for the entropy-regularized Markov game, and its identified reward set.

\begin{definition}[Quantal response equilibrium]
For each time step $h$, there is a unique pair of optimal policies $(\mu^*,\nu^*)$ of the entropy-regularized Markov game, i.e. the quantal response equilibrium (QRE), characterized by the  minimax problem that
\begin{equation*}
    V_h^{\mu^*,\nu^*}(s) = \mathop{\max}_{\mu_h}\mathop{\min}_{\nu_h} V_h^{\mu,\nu}(s) = \mathop{\min}_{\nu_h}\mathop{\max}_{\mu_h} V_h^{\mu,\nu}(s),
\end{equation*}
which is equivalent to
\begin{equation}\label{eq 3.3}
\begin{aligned}
    &V_h^{\mu^*,\nu^*}(s)= \max_{\mu_h}\min_{\nu_h}\bigl[\mu_h(s)^{\top}Q_h^{\mu,\nu}(s)\nu_h(s) +\eta^{-1}\mathcal{H}(\mu_h(\cdot|s))-\eta^{-1}\mathcal{H}(\nu_h(\cdot|s))\bigr],
\end{aligned}
\end{equation}
where $\mu_h(\cdot)\in\mathbb{R}^{m}$ is the policy adopted by player 1, $\nu_h(\cdot)\in\mathbb{R}^{n}$ is the policy adopted by player 2, and $\mathcal{H}(\pi)=  -\sum_{i}\pi_i\log(\pi_i)$ is the Shannon entropy of  distribution $\pi$.
Also, the unique solution of this max-min problem (the QRE) satisfies the following fixed point equations that
\begin{equation}\label{1.1}
    \begin{cases}
        \displaystyle\mu_h^*(a|s) = \frac{e^{\eta\langle Q_h^*(s,a,\cdot),\nu_h^*(\cdot|s)\rangle_{\mathcal{B}}}}{\sum_{a\in\mathcal{A}}e^{\eta\langle Q_h^*(s,a,\cdot),\nu_h^*(\cdot|s)\rangle_{\mathcal{B}}}},\vspace{0.2cm}\\
        \displaystyle\nu_h^*(b|s) = \frac{e^{-\eta\langle Q_h^*(s,\cdot,b),\mu^*_h(\cdot|s)\rangle_{\mathcal{A}}}}{\sum_{b\in\mathcal{B}}e^{-\eta\langle Q_h^*(s,\cdot,b),\mu^*_h(\cdot|s)\rangle_{\mathcal{A}}}},
    \end{cases}
\end{equation}
where $a\in\mathcal{A}\backslash\{1\}$ and $b\in\mathcal{B}\backslash\{1\}$.
\end{definition}

\begin{definition}[identified reward set] Given state and action space $\mathcal{S}\times\mathcal{A}\times\mathcal{B}$ and the quantal response equilibrium $(\mu^*,\nu^*)$, a reward function $r:\mathcal{S}\times\mathcal{A}\times\mathcal{B}\rightarrow\mathbb{R}^H$ is \textit{identified} if $(\mu_h^*,\nu_h^*)$ is the solution to the minimax problem \eqref{eq 3.3} induced by the reward function $r_h$ for all $h\in[H]$. Then, the \textit{identified reward set} is
\begin{equation*}
\begin{aligned}
\mathcal{R}^{\mu^*,\nu^*}=\bigl\{r:\,&\mathcal{S}\times\mathcal{A}\times\mathcal{B}\rightarrow\mathbb{R}^H,\,r\
 \text{is identified given}\ (\mu^*,\nu^*)\bigr\}.
\end{aligned}
\end{equation*}
\end{definition}

In the remainder of this section, we study the inverse game theory for this entropy-regularized zero-sum Markov game. In particular, we observe $\{(s_h^t,a_h^t,b_h^t)\}_{h\in[H],t\in[T]}$ following the QRE $(a_h^t,b_h^t)\stackrel{\text{i.i.d.}}{\sim}(\mu_h^*(\cdot|s_h^t),\nu_h^*(\cdot|s_h^t))$,  where we have $T$ independent sample paths. Our goal is to recover all  feasible reward functions $r$. 

\subsection{Learning Reward Functions from Actions}\label{sec:3.2}
We propose a novel algorithm to find all  feasible reward functions that lead to the QRE. Similar to our setting in Section \ref{sec:2}, we assume that both the reward function and transition kernel have linear structures (\citealp{Bradtke2004LinearLA}; \citealp{pmlr-v125-jin20a}).

\begin{assumption}[Linear MDP]\label{alin}
    Let $\phi:\mathcal{S}\times\mathcal{A}\times\mathcal{B}\to\bbR^d$ be a feature map. For the underlying MDP, we assume that for every reward function $r_h:\mathcal{S}\times\mathcal{A}\times\mathcal{B} \rightarrow [0, 1]$ and every transition kernel $\mathbb{P}_h:\mathcal{S}\times\mathcal{A}\times\mathcal{B}\to\Delta
    (\mathcal{S})$,
there exist $\omega_h,\pi_h(\cdot) \in \mathbb{R}^d$
such that 
\begin{equation*}
\begin{aligned}
r_h(s,a,b) &=\phi(s, a,b)^\top \omega_h,\quad\text{and}\quad \mathbb{P}_h(\cdot|s, a,b) &= \phi(s, a,b)^\top \pi_h(\cdot)
\end{aligned}
\end{equation*}
for all $(s,a,b)\in\mathcal{S}\times\mathcal{A}\times\mathcal{B}$.  In addition, the Q-function is linear with respect to $\phi$. Namely, for any QRE $(\mu,\nu)$ and $h\in[H]$, there exists a vector $\theta_h\in\mathbb{R}^d$ such that
\[Q_h(s,a,b) = \phi(s,a,b)^\top\theta_h.\] In addition, we assume $\lVert\phi(\cdot,\cdot,\cdot)\rVert\leq 1,\lVert \theta_h\rVert\leq R$ for some constant $R>0$, and $\lVert \pi_h(\mathcal{S})\rVert\leq \sqrt{d}$.
\end{assumption}

In Assumption \ref{alin}, since the reward functions $r_h$'s are normalized to the unit interval $[0,1]$, and the time horizon $[H]$ is finite, every Q-function $Q_h$ is bounded by some constant, and the constant $R\geq H(1+\log m+\log n)$ exists. Since each $\omega_h$ can be recovered by the corresponding $\theta_h$, we only make an assumption on $\theta_h$  for ease of presentation.

Primarily, we aim to find plausible parameters $\omega_h$ for all $h\in[H]$ under Assumption \ref{alin}. Analogous to matrix games discussed in the previous section, {we first examine the identifiability of the Q-functions that whether the equilibrium condition \eqref{1.1} admits a unique parameter $\theta_h$ for the QRE. Building on this constraint, we establish a strong identifiability result in the next proposition.} 
\begin{proposition}[Strong Q-identifiability]\label{p3.3}
Under Assumption \ref{alin}, for each $h\in[H]$, let $\theta_h^*$ be a solution to the following system of $|\mathcal{S}|$ linear equations that
\begin{equation}\label{linmg}
    \begin{bmatrix}
    A_h(s,\nu_h^*)\\
    B_h(s,\mu_h^*)
\end{bmatrix}\theta_h = \begin{bmatrix}
    c_h(s,\mu_h^*)\\
    d_h(s,\nu_h^*)
\end{bmatrix}\quad\text{for all $s\in\mathcal{S}$}.
\end{equation} 
Here we let $A_h$, $B_h$, $c_h$, and $d_h$ be 
\begin{equation}
\begin{aligned}
A_h(s,\nu_h)&= \bigl(\left(\phi(s,a,\cdot)-\phi(s,1,\cdot)\right) \nu_h(s)\bigr)_{a\in\mathcal{A}\backslash\{1\}}\in\bbR^{(m-1)\times d},\\
B_h(s,\mu_h)&=\bigl(\left(\phi(s,\cdot,b)-\phi(s,\cdot,1)\right)\cdot \mu_h(s)\bigr)_{b\in\mathcal{B}\backslash\{1\}}\in\bbR^{(n-1)\times d}, \\
c_h(s,\mu_h)& = \bigl(\eta^{-1} \cdot  \log \mu_h(a|s)-  \eta^{-1} \cdot  \log \mu_h(1|s)\bigr)_{a\in\mathcal{A}\backslash\{1\}}\in\mathbb{R}^{m-1}, \\
d_h(s,\nu_h) & =  \bigl (-\eta^{-1}\cdot\log\nu_h(b|s) + \eta^{-1} \cdot \log \nu_h(1|s)\bigr)_{b\in\mathcal{B}\backslash\{1\}}\in\mathbb{R}^{n-1}.
\end{aligned}\label{mknotations}
\end{equation}
Letting 
$Q_h^*(s,a,b) = \phi(s,a,b)^\top\theta_h^*$ for all $(s,a,b)\in\mathcal{S}\times\mathcal{A}\times\mathcal{B}$, we have that $Q^* = \{Q_h^*\}_{h\in[H]}$ is identified for $\mu^*$ and $\nu^*$. That is, $(\mu_h^*, \nu_h^*)$ is the solution to the entropy-regularized matrix game~\eqref{eq 3.3}, with $Q_h^*$ being the payoff matrix. 

Moreover, for any joint policy $(\mu, \nu)$, with some slight abuse of notation, we let matrix-valued functions $A_h$ and $B_h$ be
\begin{equation*}
\begin{aligned}
A_h(\nu_h) &= \begin{bmatrix}
    A_h(1,\nu_h)\\
    \vdots\\
    A_h(|\mathcal{S}|,\nu_h)
\end{bmatrix}\in\bbR^{S(m-1)\times d},\quad
B_h(\mu_h)= \begin{bmatrix}
    B_h(1,\mu_h)\\
    \vdots\\
    B_h(|\mathcal{S}|,\mu_h)
\end{bmatrix}\in\bbR^{S(n-1)\times d}.
\end{aligned}
\end{equation*}
Consider the following rank condition for each pair of $A_h(\nu^*_h)$ and $B_h(\mu_h^*)$ that  
    \begin{equation}\label{eq3.3}
        \text{rank}\left(\begin{bmatrix}
    A_{h}(\nu_h^*)\\ B_{h}(\mu_h^*)
\end{bmatrix}\right) = d,
\end{equation}
where $d$ is the dimension of kernel function $\phi$. We have that $Q^*$ is the unique function that is identified for $(\mu^*, \nu^*)$  {\bf if and only if} the above rank condition holds for all $h\in [H]$.   
\end{proposition}
\proof{Proof.}
    See Appendix \ref{pmgc} for the complete proof.
\endproof

By the Bellman equation \eqref{bel}, $r_h$ is an identified reward function if and only if there exists a feasible Q function $Q_h$ and V function $V_{h+1}$ such that
\begin{equation}\label{eq3.7}
\begin{aligned}
&r_h(s,a,b) = Q_h(s,a,b) -\gamma \cdot \mathbb{E}_{s^\prime\sim\mathbb{P}_h(\cdot|s,a,b)}\left[V_{h+1}(s^\prime)\right].
\end{aligned}
\end{equation}
Following this observation, together with the strong identifiability of Q-function, we have the strong identifiability of reward function in the next proposition.

\begin{proposition}[Strong identifiability of reward function]\label{p3.5}
    Under Assumption \ref{alin}, for each $h\in[H]$, there exists a unique reward function $r_h$ if and only if the QRE satisfies the rank condition
    \begin{equation}\label{rankcondmarkov}
        \mathrm{rank}\left(\begin{bmatrix}
    A_{h^\prime}(\nu_{h^\prime}^*) \\ B_{h^\prime}(\mu_{h^\prime}^*)
\end{bmatrix}\right) = d,\quad\text{for all $h\leq h^\prime\leq H$,}
\end{equation}
i.e. there exists a unique feasible $Q_{h^\prime}$, for all $h^\prime\in\{h,h+1,\cdots, H\}$. 
\end{proposition}
\proof{Proof.}
    Combining \eqref{bel} and \eqref{vrep}, the claim holds immediately by Proposition \ref{p3.3}.
\endproof

Propositions \ref{p3.3} and \ref{p3.5} provide the rank conditions for the strong identifiability of the reward function in Markov games. However, in practical settings,  rank condition \eqref{rankcondmarkov} may not hold due to limited exploration of the state-action space or structural dependencies in the equilibrium strategies, especially when the time horizon $H$ is large. When the QRE does not satisfy the full-rank condition across all time steps $h\in[H]$, the set of identified Q-functions is no longer a singleton, leading to partial identifiability of the reward function. This means that while we can still infer a set of plausible reward functions consistent with the observed equilibrium behavior, the true reward function may not be uniquely determined. Addressing this issue requires a more flexible formulation that accounts for multiple possible Q-functions while maintaining meaningful constraints on the parameter space. Similar to our discussion in Section \ref{sec:piden}, we define the feasible reward set below.

\begin{definition}[feasible reward sets]\label{rfeasible}
Let $\phi:\mathcal{S}\times\mathcal{A}\times\mathcal{B}\to\bbR^d$ be the kernel map to parameterize our Markov game. For each $h\in[H]$, a parameter $\theta_h$ is \textit{feasible}, if it satisfies the identifiability equation \eqref{linmg} and the norm constraint $\Vert\theta_h\Vert\leq R$. Let $\Theta_h$ be the set of all feasible parameters $\theta_h$ that
\begin{equation*}
\begin{aligned}
\Theta_h=\bigl\{\theta_h:\ \theta_h\ &\text{satisfies the identifiability condition \eqref{linmg} and}\ \Vert\theta_h\Vert\leq R\bigr\}.
\end{aligned}
\end{equation*}
This yields the set of feasible Q-functions that
\begin{equation}
\begin{aligned}
\mathcal{Q}_h&=\left\{Q_h(\cdot,\cdot,\cdot)=\theta_h^\top\phi(\cdot,\cdot,\cdot):\theta_h\in\Theta_h\right\},\quad \mathcal{Q}=\mathcal{Q}_1\times\mathcal{Q}_2\times\cdots\times\mathcal{Q}_H.
\end{aligned}
\end{equation}
In addition, the set of feasible reward functions corresponding to the QRE $(\mu^*,\nu^*)$ is
\begin{equation*}
\begin{aligned}
\mathcal{R}=\bigl\{&r:\mathcal{S}\times\mathcal{A}\times\mathcal{B}\to\bbR^H,\ r\ \text{is a reward function corresponding to some}\ Q\in\mathcal{Q}\bigr\}.
\end{aligned}
\end{equation*} 
\end{definition}
Equivalently, a reward function $r$ is feasible if the induced Q-functions satisfy the QRE constraints~\eqref{1.1}, and the associated parameters $\theta_h$'s satisfy the norm constraint $\Vert\theta_h\Vert \leq R$. This ensures that (i) the given QRE $(\mu^*, \nu^*)$ is a solution to the corresponding minimax problem under $r$, and (ii)~the reward lies within a parameter space that avoids overfitting by regularizing model complexity.

Our formulation provides a principled way to handle the partial identifiability in Markov games. Instead of forcing a single estimated reward function, we construct a structured set of feasible rewards, which offers a more robust approach to analyzing decision-making in complex multi-step strategic settings. Intuitively, the norm constraint $\Vert\theta_h\Vert\leq R$ plays a key role in ensuring that the estimated reward functions remain well-conditioned, and do not include arbitrarily large coefficients. Additionally, by linking the feasible reward set to the recursive Bellman equations \eqref{bel}-\eqref{vrep}, our definition ensures that every element of $\wh{\mathcal{R}}$ maintains the temporal consistency. In other words, the inferred rewards lead to equilibrium strategies that are valid over multiple decision-making steps.

\paragraph{Algorithm for robust reward learning.} We propose an algorithm to recover the feasible reward functions. For all $h\in[H]$, the algorithm  works that
\begin{itemize}
    \item Estimate the QRE $(\mu_h^*,\nu^*_h)$ from the observed data. We compute the frequency estimator that for all $(s,a,b)\in\mathcal{S}\times\mathcal{A}\times\mathcal{B}$
\begin{equation}
\begin{cases}
    \displaystyle\widehat{\mu}_h(a|s) = \frac{1}{N_h(s)\vee1}\sum_{(s_h,a_h)\in\mathcal{D}}\mathbf{1}_{(s_h,a_h) = (s,a)},\vspace{0.1cm}\\
    \displaystyle\widehat{\nu}_h(b|s)=\frac{1}{N_h(s)\vee1}\sum_{(s_h,b_h)\in\mathcal{D}}\mathbf{1}_{(s_h,b_h) = (s,b)},
\end{cases}\label{mgqrefrequency}
\end{equation}
where $N_h(s) =\sum_{(s_h,a_h,b_h)\in\mathcal{D}_h}\mathbf{1}_{\{s_h=s\}}$;
    \item Recover the feasible set $\Theta_h$ by solving the least square problem associated with the linear system \eqref{linmg} that
    \begin{equation}
\wh{\Theta}_h = \left\{\theta:\left\lVert \begin{bmatrix}
    A_h(\widehat{\nu}_h)\\
    B_h(\widehat{\mu}_h)
\end{bmatrix}\theta -\begin{bmatrix}
    c_h(\widehat{\mu}_h)\\
    d_h(\widehat{\nu}_h)
\end{bmatrix}\right\rVert^2\leq \kappa_h,\lVert\theta\rVert\leq R\right\};\label{mkconfset}
    \end{equation}
    \item Calculate the feasible Q and V functions ($Q_h$ and $V_h$) for all $\wh\theta_h\in\wh\Theta_h$ by
\begin{equation}
\begin{aligned}
&\wh{Q}_h(s,a,b)= \phi(s,a,b)^\top\widehat{\theta}_h,\quad\text{and}\\
&\wh{V}_{h}(s)= \widehat{\mu}_h(s)^{\top}\widehat{Q}_h(s)\widehat{\nu}_h(s)+\eta^{-1}\mathcal{H}(\widehat{\mu}_h(s))-\eta^{-1}\mathcal{H}(\widehat{\nu}_h(s)).
\end{aligned}\label{learnQV}
\end{equation}
\end{itemize}
Up to this point, our approach to recovering value functions of the Markov game follows the same outline as the matrix game in Section~\ref{sec:piden}. However, to reconstruct the reward functions, it is essential to incorporate an estimator for the transition dynamics. Specifically, since the rewards in Markov games depend on future states, directly recovering them requires knowledge of the transition kernel. Therefore, we extend our framework by first estimating the transition model $\bbP_h$ from observed data. Given the estimator for the transition model, we leverage a plug-in version of the Bellman equation to compute the feasible reward set $\wh{\mathcal{R}}_h$. This route ensures that the recovered rewards remain consistent with the estimated Q-functions and equilibrium conditions. In particular, our algorithm works as follows that
\begin{itemize}
    \item Estimate the transition kernel $\mathbb{P}_h$ from the observed data. Since the transition kernel has a linear structure, we employ ridge regression (\citealp{pmlr-v125-jin20a}) that
    \begin{equation}
\begin{aligned}
    \Lambda_h & = \sum_{t = 1}^T\phi(s_h^t,a_h^t,b_h^t)\phi(s_h^t,a_h^t,b_h^t)^\top+\lambda\mathbf{I}_d,\\
\widehat{\mathbb{P}}_h\widehat{V}_{h+1}(s,a,b) & = \phi(s,a,b)^\top\Lambda_h^{-1}\sum_{t = 1}^T\phi(s_h^t,a_h^t,b_h^t)\widehat{V}_{h+1}(s_{h+1}^t);
\end{aligned}\label{mkridge}
\end{equation}
The regularization coefficient $\lambda$ in \eqref{mkridge} mitigates overfitting of the transition model to finite samples and prevents instability in downstream computations, such as the estimation of rewards. From the bias-variance tradeoff perspective, a larger $\lambda$ introduces more bias but reduces variance, while a smaller $\lambda$ yields an unbiased yet potentially unstable estimate. In practice, $\lambda$ can be selected by cross-validation or set as a small constant, e.g., $10^{-3}$.

    \item Recover the feasible reward set $\mathcal{R}_h$ at step $h$ by  a plug-in version of Bellman equation \eqref{eq3.7} that
    \begin{equation}
        \wh{r}_h(s,a,b)=\wh{Q}_h(s,a,b)-\gamma\cdot\wh{\bbP}_h\wh{V}_{h+1}(s,a,b),\label{mkbelrecover}
    \end{equation}
    where $\wh{Q}_h\in\wh{\mathcal{Q}}_h$ and $\wh{V}_{h+1}\in\wh{\mathcal{V}}_{h+1}$.
\end{itemize}
We summarize the steps in Algorithm \hyperref[a1]{2}.

\begin{algorithm}
\caption{Learning reward from actions (frequency estimator case)}
\begin{algorithmic}[1]\label{a1}
\Require{Dataset $\mathcal{D} = \{(s_h^t,a_h^t,b_h^t)_{h=1}^H\}_{t\in[T]}$, kernel $\phi(\cdot,\cdot,\cdot)$, entropy regularization term $\eta$, discount factor $\gamma$, threshold parameters $(\kappa_h)$, ridge regularization term $\lambda$.}
\Ensure{Confidence set $\wh{\mathcal{R}}$ of feasible rewards.}
\For {$(h,s,a,b)\in[H]\times\mathcal{S}\times\mathcal{A}\times\mathcal{B}$}
\State Compute the empirical QRE $(\wh\mu,\wh\nu)$ by \eqref{mgqrefrequency};
\EndFor
\For{$h=H,H-1,\cdots,1$}
\State Construct confidence set $\wh{\Theta}_h$ by \eqref{mkconfset};
\State Compute feasible Q and V-functions by containing all $\wh{Q}_h$ and $\wh{V}_h$ in \eqref{learnQV}.
\State Compute empirical transition by \eqref{mkridge};
\State Compute the reward $\wh{\mathcal{R}}_h$ by \eqref{mkbelrecover}.
\EndFor
\State\Return $\wh{\mathcal{R}}=\wh{\mathcal{R}}_1\times\wh{\mathcal{R}}_2\times\cdots\times\wh{\mathcal{R}}_H.$
\end{algorithmic}
\end{algorithm}

\subsection{Theoretical Guarantees}\label{sec:3.3}
We derive the theoretical results for the feasible sets obtained by Algorithm \hyperref[a1]{2}. To facilitate our discussion, we first define the base metric to measure the distance between rewards.

\begin{definition}[Uniform metric for rewards]\label{basem}
The metric $D$ between a pair of rewards $r,r^\prime \in \mathcal{R}$ is
\begin{equation}
D(r,r^\prime) = \sup_{h\in[H],(s,a,b)\in\mathcal{S}\times\mathcal{A}\times\mathcal{B}}\left\vert (r_h-r_h^\prime)(s,a,b)\right\vert.\label{unifrewmetric}
\end{equation}
\end{definition}

With the above metric defined, we derive theoretical results for the proposed algorithm. Our goal is to measure the difference between the true feasible set $\mathcal{R}$ and the estimated set $\widehat{\mathcal{R}}$. In particular, we  bound $D(\mathcal{R},\widehat{\mathcal{R}})$, which is the Hausdorff distance induced by the metric in Definition~\ref{basem}. For the first step of our analysis, we determine the thresholds of the confidence sets for Q-function parameters in the next lemma.

\begin{lemma}[Construction error of Q-function parameter set]\label{qest}
    Under Assumption \ref{alin}, denote by $\widehat{\Theta}_h$ the confidence set \eqref{mkconfset} obtained by Algorithm \hyperref[a1]{2} for each $h\in[H]$. Let $\epsilon_1$ and $\epsilon_2$ be two small positive numbers such that $\epsilon_1<\min_{s\in\mathcal{S},a\in[m]}\mu_h^*(a|s)$ and $\epsilon_2<\min_{s\in\mathcal{S},b\in[n]}\nu_h^*(b|s)$.
Suppose that $\text{TV}(\widehat{\mu}_h^*(\cdot|s),\mu_h^*(\cdot|s))\leq \epsilon_1/2$ and $\text{TV}(\widehat{\nu}_h^*(\cdot|s),\nu_h^*(\cdot|s))\leq \epsilon_2/2$ for all $s\in\mathcal{S}$ and all $h\in[H]$.
Let 
\begin{equation*}
\begin{aligned}
\kappa_h&=2R^2\left(\Vert\Phi_1\Vert_{\mathrm{op}}^2\epsilon_1^2+\Vert\Phi_2\Vert_{\mathrm{op}}^2\epsilon_2^2\right)+\frac{2Sm\epsilon_1^2}{\eta^2\Bigl(\underset{s\in\mathcal{S},a\in[m]}{\min}\mu_h^*(a|s)-\epsilon_1\Bigr)^2}+\frac{2Sn\epsilon_2^2}{\eta^2\Bigl(\underset{s\in\mathcal{S},b\in[n]}{\min}\nu_h^*(b|s)-\epsilon_2\Bigr)^2},
\end{aligned}
\end{equation*}
where the normalized feature matrices $\Phi_1\in\mathbb{R}^{d\times s(m-1)n}$ and $\Phi_2\in\mathbb{R}^{d\times s(n-1)m}$ are defined as
\begin{equation*}
\begin{aligned}
&\Phi_1 = \left(\phi(s,a,\cdot)-\phi(s,1,\cdot)\right)_{s\in\mathcal{S},a\in[m]\backslash\{1\}},\quad\Phi_2 = \left(\phi(s,\cdot,b)-\phi(s,\cdot,1)\right)_{s\in\mathcal{S},b\in[n]\backslash\{1\}}.
\end{aligned}
\end{equation*}
Then, we have $\Theta_h\subseteq\wh{\Theta}_h$. Furthermore,  for sufficiently small $\epsilon_1,\epsilon_2>0$, we have
\begin{equation}
D_H(\Theta_h,\wh{\Theta}_h)\lesssim\sqrt{\kappa_h}.\label{mkQHausdorff}
\end{equation}
\end{lemma}
\proof{Proof.}
    See Appendix \ref{pqest} for the complete proof.
\endproof

This lemma provides a strong  guarantee for the reliability of our confidence set in inverse game learning. The inclusion $\Theta_h\subseteq\widehat{\Theta}_h$ ensures that our method fully captures all feasible Q-function parameters under a finite sample setting. Moreover, the error bound \eqref{mkQHausdorff} shows that the estimated confidence set  progressively shrinks to the true feasible set as the threshold $\kappa_h$ decreases. Since the threshold $\kappa_h$ depends explicitly on the QRE estimation errors $\epsilon_1$ and $\epsilon_2$, we  follow the same roadmap as in Section \ref{sec:2}, and derive the sample complexity of QRE estimation.

For the sake of clarity,  we fix the initial state distribution in the Markov game $\rho_1\in\Delta(\mathcal{S})$, and define the marginal state visit distributions associated with policies $\mu,\nu$ at each time step $h\in[H]$ as $d_h^{\mu,\nu}(s)=\mathbb{P}(s_h = s|\rho_1,\mu,\nu)$. Also, we denote the state-action visit distributions as $d_h^{\mu,\nu}(s,a,b)=\mathbb{P}(s_h = s,a_h = a, b_h = b|\rho_1,\mu,\nu)$.

To control the uniform metric defined in (\ref{unifrewmetric}),  we require an estimator for the QRE that performs uniformly well across all states $s\in\mathcal{S}$. When using frequency estimators to approximate the policies $\mu_h^*(\cdot|s)$ and $\nu_h^*(\cdot|s)$, the estimation at each state is conducted independently. As a result, it is essential that the dataset sufficiently covers all states in $\mathcal{S}$ to obtain reliable estimates. To ensure this, we impose the following assumption, which guarantees that every state is visited with a minimum frequency throughout the horizon. 

\begin{assumption}[$C$-well-posedness]\label{wellposed}
There exists a constant $C>0$ such that the marginal state visit distributions satisfy $$d_h^{\mu^*,\nu^*}(s)\geq C$$ for all $s\in\mathcal{S}$ and $h\in[H]$, where $d_h^{\mu^*,\nu^*}(\cdot)$ is the state visit distribution at step $h$ under the QRE policies $(\mu^*,\nu^*).$
\end{assumption}

This assumption guarantees that every state $s\in\mathcal{S}$ is visited with sufficient probability under the equilibrium policy, ensuring the concentration of frequency-based estimators. Based on this, we derive a uniform convergence result for the QRE estimators over the entire state space.

\begin{lemma}[Concentration of QRE]\label{conequ}
     Under Assumptions \ref{alin} and \ref{wellposed}, let $\epsilon$ be a small positive number such that $\epsilon\leq\min_{h\in[H],s\in\mathcal{S},a\in[m],b\in[n]}\{\mu_h^*(a|s),\nu_h^*(b|s)\}/2$. Let the number of sample episodes  be
     \[T=\frac{1}{2C^2}\cdot\log(2HS/\delta)+\frac{m\vee n}{2C\epsilon^2}\cdot\log(4HS/\delta),\]
     where $\delta\in(0,1)$. Then, the concentration event,
     \begin{equation}\label{concevent1}
         \begin{aligned}
             &\mathcal{E}_1=\bigg\{\mathrm{TV}(\wh\mu_h(\cdot|s),\mu_h^*(\cdot|s))\leq\epsilon,\mathrm{TV}(\wh\nu_h(\cdot|s),\nu_h^*(\cdot|s))\leq\epsilon,\forall s\in\mathcal{S},\forall h\in[H]\bigg\},
         \end{aligned}
     \end{equation}
    holds with probability at least $1-\delta$.
\end{lemma}
\proof{Proof.}
    See Appendix \ref{pconequ} for the complete proof.
\endproof

Combining Lemmas \ref{qest} and \ref{conequ}, we establish the sample complexity required to construct the confidence set $\wh{\Theta}_h$.

\begin{corollary}[Sample complexity of constructing feasible Q function set]\label{qsetuncertainty} Under Assumptions~\ref{alin} and \ref{wellposed}, and given any $\delta\in(0,1)$, we set $T\in\mathbb{N}$ such that
\begin{equation*}
    T\geq\frac{1}{C^2}\log\frac{2HS}{\delta},
\end{equation*}
and $\epsilon>0$ such that
\begin{equation*}
\begin{aligned}
    \epsilon&=\sqrt{\frac{m\vee n}{CT}\log\frac{4HS}{\delta}}\leq\frac{1}{4}\min_{s\in\mathcal{S},a\in[m],b\in[n]}\{\mu_h^*(a|s),\nu_h^*(b|s)\}.
\end{aligned}
\end{equation*}
Then the concentration event $\mathcal{E}_1$ in \eqref{concevent1}  holds with probability at least $1-\delta$. Moreover, we set the threshold parameter as $\mathcal{O}(T^{-1})$ that
\begin{equation*}
\begin{aligned}
\kappa_h&=8\epsilon^2\Biggl(R^2(\Vert\Phi_1\Vert_{\mathrm{op}}^2+\Vert\Phi_2\Vert_{\mathrm{op}}^2)+\frac{4Sm}{\eta^2\min_{s\in\mathcal{S},a\in[m]}\mu_h^*(a|s)^2}+\frac{4Sn}{\eta^2\min_{s\in\mathcal{S},b\in[n]}\nu_h^*(b|s)^2}\Biggr).
\end{aligned}
\end{equation*}
Then for each $h\in[H]$, the concentration event,
$$\mathcal{E}_2=\left\{\Theta_h\subseteq\widehat{\Theta}_h,D_H(\Theta_h,\widehat{\Theta}_h)\lesssim\sqrt{\kappa_h},\ \forall h\in[H]\right\},$$
holds with probability at least $1-\delta$.
\end{corollary}

Finally, to derive an error bound for reward function estimation, we incorporate both the Q-function estimation error at each step and the transition kernel estimation error into our analysis. By the Bellman equation \eqref{eq3.7}, the next theorem provides the sample complexity required to control the Hausdorff distance between the true feasible reward set and the constructed confidence set.
\begin{theorem}[Sample complexity of constructing feasible reward set]\label{samp}
Under Assumptions \ref{alin} and~\ref{wellposed}, let $\rho_h=d^*_h$ be the stationary distribution associated with optimal policies $\mu^*$ and $\nu^*$, where $h\in[H]$. We assume that 
\begin{equation*}
\Psi_h=\bbE_{\rho_h}\left[\phi(s_h,a_h,b_h)\phi(s_h,a_h,b_h)^\top\right]\in \mathbb{R}^{d\times d}
\end{equation*}
is nonsingular for all $h\in[H]$. Let $\mathcal{R}$ be the feasible reward set given in Definition \ref{rfeasible}. Given a dataset $\mathcal{D}=\{\mathcal{D}_h\}_{h\in[H]}= \{\{(s_h^t,a_h^t,b_h^t)\}_{t\in[T]}\}_{h\in[H]}$, we set $\lambda=\mathcal{O}(1)$, $\kappa_h=\mathcal{O}(T^{-1})$, and let $\widehat{\mathcal{R}}$ be the output of Algorithm \hyperref[a1]{2}. 
Let $\xi=\min_{h\in[H],s\in\mathcal{S},a\in\mathcal{A},b\in\mathcal{B}}\{\mu_h^*(a|s),\nu_h^*(b|s)\}$. For any $\delta\in(0,1)$, let $T>0$ be sufficiently large such that
\begin{equation*}
\begin{aligned}
&T\geq\max\biggl\{\frac{1}{C^2}\log\frac{2HS}{\delta},\frac{16(m\vee n)}{C\xi^2}\log\frac{4HS}{\delta}, 512\Vert\Psi_h^{-1}\Vert^2_\mathrm{op}\log\frac{2Hd}{\delta},4\lambda\Vert\Psi^{-1}_h\Vert_\mathrm{op}\biggr\}.
\end{aligned}
\end{equation*}
Then, we have, with probability at least $1-3\delta$,
\begin{equation*}
\begin{aligned}
D(\mathcal{R},\widehat{\mathcal{R}})&\lesssim\frac{1}{\sqrt{T}}\Biggl(\sqrt{S(m+n)\log\frac{HS}{\delta}}\left(\sqrt{S(m+n)}+\log T\right)+\left(\sqrt{Sd}+\sqrt{d\log T}\right)\log(mn)\Biggr).
\end{aligned}
\end{equation*}
\end{theorem}
\proof{Proof.}
    See Appendix \ref{psamp} for the complete proof.
\endproof

Theorem \ref{samp} provides a strong guarantee on the accuracy of our reward recovery algorithm in Markov games. Our bound shows that the distance $D(\mathcal{R},\wh{\mathcal{R}})$ diminishes at the rate of $\mathcal{O}(T^{-1/2})$, which matches the optimal statistical rate for empirical risk minimization problems (\citealp{Vaart_1998}). This demonstrates that with sufficient data, the estimated reward functions remain close to the true feasible set, making our method statistically reliable and sample-efficient. The explicit dependence on problem parameters offers insights into how exploration, feature representations, and action space size affect the difficulty of inverse reward learning in Markov games.

We also note that the condition that $\Psi_h$ is nonsingular ensures that the feature representation provides sufficient information for parameter recovery (\citealp{tu2017leastsquarestemporaldifferencelearning}; \citealp{min2022varianceawareoffpolicyevaluationlinear}). The norm $\Vert\Psi_h^{-1}\Vert_{\mathrm{op}}$ appears in the sample complexity bound, indicating that ill-conditioned feature matrices lead to larger estimation errors and require more samples to achieve the same level of accuracy.

\subsection{MLE-Based QRE Estimation}\label{sec:3.4}
To control the error of the frequency estimator $(\wh{\mu},\wh{\nu})$ of QRE, we assume that we visit every state $s\in\mathcal{S}$ with a sufficiently high probability. This strong assumption can be restrictive in practical scenarios, as it requires an exhaustive exploration of the state space $\mathcal{S}$. To alleviate this limitation, we assume that the QRE $(\mu^*,\nu^*)$ exhibits a sparse structure. In particular, we assume that by adjusting the dimension of the parameters, we can capture the key features of the QRE.

\begin{assumption}[Linearly parameterized QRE]\label{policysoftmaxpara}
Let $d_a,d_b\in\mathbb{N}$. Let $\psi_a:\mathcal{S}\times\mathcal{A}\to\bbR^{d_a}$ and $\psi_b:\mathcal{S}\times\mathcal{B}\to\bbR^{d_b}$ be two bounded feature maps such that $\Vert\psi_a(s,a)\Vert,\Vert\psi_b(s,b)\Vert\leq K$ for all $s\in\mathcal{S},a\in\mathcal{A}$ and $b\in\mathcal{B}$. For each $h\in[H]$, there exists a vector $\vartheta_h^*\in\bbR^{d_a}$ such that $\Vert\vartheta_h^*\Vert\leq 1$, and for all $s_h\in\mathcal{S},a_h\in\mathcal{A}$,
\begin{equation*}
\mu_h^*(a_h|s_h)=\frac{\exp\left(\vartheta_h^{*\top}\psi_a(s_h,a_h)\right)}{\sum_{a\in\cal{A}}\exp\left(\vartheta_h^{*\top}\psi_a(s_h,a)\right)};
\end{equation*}
Also, there exists a vector $\zeta_h^*\in\mathbb{R}^{d_b}$ such that $\Vert\zeta_h^*\Vert\leq 1$, and for all $s_h\in\mathcal{S},b_h\in\mathcal{B}$,
\begin{equation*}
\nu_h^*(b_h|s_h)=\frac{\exp\left(\zeta_h^{*\top}\psi_b(s_h,b_h)\right)}{\sum_{b\in\cal{B}}\exp\left(\zeta_h^{*\top}\psi_b(s_h,b)\right)}.
\end{equation*}
\end{assumption}

Essentially, this linear parametrization assumption allows us to integrate the estimation of the QRE $(\mu^*_h(\cdot|s),\nu^*_h(\cdot|s))$ across different states $s\in\cal{S}$ by identifying the parameters with maximum likelihood estimation (MLE). In particular,  given the dataset
\begin{equation*}
    \mathcal{D}_h=\left\{(s_h^t,a_h^t,b_h^t)\right\}_{t=1}^T,\quad h\in[H],
\end{equation*}
for the max player, the negative log-likelihood function is
\begin{equation*}
    \mathcal{L}_h^a(\vartheta_h)=-\sum_{t=1}^T\log\frac{\exp\left(\vartheta_h^{\top}\psi_a(s_h^t,a_h^t)\right)}{\sum_{a\in\cal{A}}\exp\left(\vartheta_h^{\top}\psi_a(s_h^t,a)\right)},
\end{equation*}
and for the min player, it is
\begin{equation*}
    \mathcal{L}_h^b(\zeta_h)=-\sum_{t=1}^T\log\frac{\exp\left(\zeta_h^{\top}\psi_b(s_h^t,b_h^t)\right)}{\sum_{b\in\cal{B}}\exp\left(\zeta_h^{\top}\psi_b(s_h^t,b)\right)}.
\end{equation*}
We estimate the parameters $\vartheta_h^*$ and $\zeta_h^*$ by respectively minimizing their negative log-likelihood that
\begin{equation*}
    \wh{\vartheta}_h=\argmin_{\Vert\vartheta_h\Vert\leq 1}\mathcal{L}_h^a(\vartheta_h),\quad\text{and}\quad
    \wh{\zeta}_h=\argmin_{\Vert\zeta_h\Vert\leq 1}\mathcal{L}_h^b(\zeta_h).
\end{equation*}

As we mentioned in Section \ref{sec:3.3}, it is challenging to control the error of the estimated QRE across all states $s\in\mathcal{S}$. We focus on the average error, and obtain the $L^2$ convergence result for the MLEs in the next lemma. We note that, rather than ensuring uniform accuracy across all states, we focus on the average error under the stationary distribution, which provides a more practical measure of estimation error. 

\begin{lemma}[Convergence of MLE]\label{mlepolicyconvg}
Under Assumption \ref{policysoftmaxpara}, let $\wh{\mu}_h$ and $\wh{\nu}_h$ be the policies generated by $\wh{\vartheta}_h$ and $\wh{\zeta}_h$. Then, there exists a constant $L_K>0$ depending on $m,n$, and $K$ only, such that with probability at least $1-\delta$, it holds that
\begin{equation}
\begin{aligned}
&\bbE\left[H^2\left(\wh{\mu}_h(\cdot|s_h),\mu_h^*(\cdot|s_h)\right)\right]\leq\frac{1}{T}(d_a\log(1+2TL_K)+\log\delta^{-1}+\sqrt{2m}e^K+2),
\end{aligned}\label{hellingerpolicybound}
\end{equation}
where $H(P,Q)=\Vert\sqrt{P}-\sqrt{Q}\Vert_2$ is the Hellinger distance between two probability distributions, and the expectation is taken with respect to the visit distribution $d_h^*$ of state $s_h$ generated by the QRE $(\mu^*,\nu^*)$. Since $\mathrm{TV}(P,Q)\leq\sqrt{2}H(P,Q)$,   (\ref{hellingerpolicybound}) implies
\begin{equation*}
\begin{aligned}
&\frac{1}{2}\bbE\left[\mathrm{TV}^2\left(\wh{\mu}_h(\cdot|s_h),\mu_h^*(\cdot|s_h)\right)\right]\leq\frac{1}{T}(d_a\log(1+2TL_K)+\log\delta^{-1}+\sqrt{2m}e^K+2).
\end{aligned}
\end{equation*}
Similarly, with probability at least $1-\delta$, we have that
\begin{equation*}
\begin{aligned}
&\frac{1}{2}\bbE\left[\mathrm{TV}^2\left(\wh{\nu}_h(\cdot|s_h),\nu_h^*(\cdot|s_h)\right)\right]\leq\frac{1}{T}(d_b\log(1+2TL_K)+\log\delta^{-1}+\sqrt{2n}e^K+2).
\end{aligned}
\end{equation*}
\end{lemma}
\proof{Proof.}
    See Appendix \ref{sec:b.3.2} for the complete proof.
\endproof

This result provides a high probability bound on the estimation error in terms of the Hellinger distance and total variation  distance. Both of the error bounds decay at a rate of $\mathcal{O}(1/T)$, which ensures that the estimated policies remain close to the true policies in terms of both probability mass and likelihood ratios. Since our framework relies on estimating Q-functions from QRE, this result provides a crucial foundation for subsequent reward recovery steps.

In practice, some states $s\in\mathcal{S}$ are visited infrequently, or in the worst case, may not appear in the dataset at all. As a result, the estimation error can be large for these less-visited states. However, such errors are less critical for states with low visit probabilities, as they have a minimal impact on overall decision-making. Therefore, rather than enforcing uniform accuracy across all states, it is sufficient to control the average estimation error weighted by the state visit distribution. To formalize this, we introduce a weaker metric to measure the distance between reward functions.
\begin{definition}[Reward metric ii]
The \textit{state-average metric} $D_1$ between any pair of rewards $r,r^\prime:[H]\times\mathcal{S}\times\mathcal{A}\times\mathcal{B}\to\bbR$ is 
$$D_1(r,r^\prime)=\sup_{h\in[H],a\in\mathcal{A},b\in\mathcal{B}}\bbE_{s\sim\rho_h}\left\vert (r_h-r^\prime_h)(s,a,b)\right\vert.$$
\end{definition}

To align with the estimation of the QRE, it is necessary to adapt the estimation method of Q-functions. Since we estimate frequently visited states more accurately, we  estimate the parameters $(\theta_h)$ by solving the least square problem weighted by visiting probabilities. Before going further, we introduce another type of identifiability result for Q-functions.

\begin{proposition}[Strong Q-identifiability]\label{qidexp}
Under Assumption \ref{alin}, for each $h\in[H]$, let $\theta^*_h$ be a solution to the  linear system that
\begin{equation}\label{explinmg}
    \begin{bmatrix}
    A_h(\nu_h^*)\\
    B_h(\mu_h^*)
\end{bmatrix}\theta_h = \begin{bmatrix}
    c_h(\mu_h^*)\\
    d_h(\nu_h^*)
\end{bmatrix}\quad\text{for all $s\in\mathcal{S}$}.
\end{equation}
Here, following \eqref{mknotations}, we define $A_h$, $B_h$, $c_h$ and $d_h$ respectively as
\begin{equation*}
\begin{aligned}
A_h(\nu_h^*) = \begin{bmatrix}
\sqrt{\rho_h(1)}A_h(1,\nu_h^*)\\ \vdots\\ \sqrt{\rho_h(S)}A_h(S,\nu_h^*)
\end{bmatrix}\in\bbR^{S(m-1)\times d},\quad
B_h(\mu_h^*)=\begin{bmatrix}
\sqrt{\rho_h(1)}B_h(1,\mu_h^*)\\ \vdots\\
\sqrt{\rho_h(S)}B_h(S,\mu_h^*)
\end{bmatrix}\in\bbR^{S(n-1)\times d},
\end{aligned}
\end{equation*}
    and
\begin{equation*}
\begin{aligned}
c_h(\mu_h^*) =\begin{bmatrix}
\sqrt{\rho_h(1)}\,c_h(1,\mu_h^*)\\ \vdots\\
\sqrt{\rho_h(S)}\,c_h(S,\mu_h^*)
\end{bmatrix}\in\bbR^{S(m-1)},\quad
d_h(\nu_h^*) =\begin{bmatrix}
\sqrt{\rho_h(1)}\,d_h(1,\nu_h^*)\\ \vdots \\ \sqrt{\rho_h(S)}\,d_h(S,\nu_h^*)
\end{bmatrix}\in\bbR^{S(n-1)}.
\end{aligned}
\end{equation*}
We let $Q_h^*(s,a,b) = \phi(s,a,b)^\top\theta_h^*$ for all $(s,a,b)\in\mathcal{S}\times\mathcal{A}\times\mathcal{B}$. Then we conclude that the Q-function $Q^*=\{Q_h^*\}_{h\in[H]}$ is identifiable for $\mu^*$ and $\nu^*$. Consider the  rank condition that
\begin{equation}\label{exprankcond}
        \text{rank}\left(\begin{bmatrix}
    A_{h}(\nu_h^*)\\ B_{h}(\mu_h^*)
\end{bmatrix}\right) = d,
\end{equation}
where $d$ is the dimension of the kernel function $\phi$. There exists a uniquely identifiable $\theta_h\in\mathbb{R}^d$ \textbf{if and only if} the above rank condition holds for all $h\in[H]$.

Furthermore, the reward function $r_h$ is uniquely identifiable if and only if the QRE satisfies the rank condition \eqref{exprankcond} for all indices $h,h+1,\cdots,H$.
\end{proposition}
\proof{Proof.}
The result holds by the same arguments for Proposition \ref{p3.3}.
\endproof

\paragraph{Estimation of Q-function.} In accordance with our discussion about the identifiability of $Q$ in Proposition \ref{qidexp}, we tailor the construction of confidence set. Instead of simply concatenating the estimation equations over all $s\in\cal{S}$, we add a weight to each block to adjust the importance of each state that
\begin{equation*}
\begin{aligned}
A_h(\wh\nu_h) = \begin{bmatrix}
\sqrt{\rho_h(1)}A_h(1,\wh\nu_h)\\ \vdots \\ \sqrt{\rho_h(S)}A_h(S,\wh\nu_h)
\end{bmatrix},\quad
B_h(\wh\mu_h)=\begin{bmatrix}
\sqrt{\rho_h(1)}B_h(1,\wh\mu_h)\\ \vdots\\
\sqrt{\rho_h(S)}B_h(S,\wh\mu_h)
\end{bmatrix}.
\end{aligned}
\end{equation*}
Since the true distribution $\rho_h$ of the state $s_h$ is unknown, we replace it with the empirical estimator on the dataset $\mathcal{D}_h$ that
\begin{equation*}
    \wh{\rho}_h(s)=\mathbb{P}_{\mathcal{D}_h}(s_h=s)=\frac{1}{T}\sum_{t=1}^T\mathbf{1}_{\{s_h^t=s\}},\quad s\in\mathcal{S}.
\end{equation*}
The matrices for confidence set construction are then  
\begin{equation}
\begin{aligned}
\wh A_h(\wh\nu_h) = \begin{bmatrix}
\sqrt{\wh\rho_h(1)}A_h(1,\wh\nu_h)\\ \vdots \\ \sqrt{\wh\rho_h(S)}A_h(S,\wh\nu_h)
\end{bmatrix},\quad
\wh B_h(\wh\mu_h)=\begin{bmatrix}
\sqrt{\wh\rho_h(1)}B_h(1,\wh\mu_h)\\ \vdots\\
\sqrt{\wh\rho_h(S)}B_h(S,\wh\mu_h)
\end{bmatrix}.\label{weightedab}
\end{aligned}
\end{equation}
Similarly, we let
\begin{equation}
\begin{aligned}
\wh c_h(\wh\mu_h)=\begin{bmatrix}
\sqrt{\wh\rho_h(1)}c_h(1,\wh\mu_h)\\ \vdots\\
\sqrt{\wh\rho_h(S)}c_h(S,\wh\mu_h)
\end{bmatrix},\quad
\wh d_h(\wh\nu_h) = \begin{bmatrix}
\sqrt{\wh\rho_h(1)}d_h(1,\wh\nu_h)\\ \vdots \\ \sqrt{\wh\rho_h(S)}d_h(S,\wh\nu_h)
\end{bmatrix}.\label{weightedcd}
\end{aligned}
\end{equation}
Given an appropriate threshold $\kappa_h>0$, we construct the confidence set for parameter $\theta_h$ by solving the least square problem that
\begin{equation}
\begin{aligned}
    &\wh{\Theta}_h= \left\{\theta:\left\lVert \begin{bmatrix}
	\wh{A}_h(\wh{\nu}_h)\\
	\wh{B}_h(\wh{\mu}_h)
\end{bmatrix}\theta -\begin{bmatrix}
	\wh{c}_h(\wh{\mu}_h)\\
	\wh{d}_h(\wh{\nu}_h)
\end{bmatrix}\right\rVert^2\leq \kappa_h,\lVert\theta\rVert\leq R\right\},  
\end{aligned}
\label{expconfset}
\end{equation}
Note that the first constraint is equivalent to the weighted least square problem
\begin{equation*}
\sum_{s\in\mathcal{S}}\wh{\rho}_h(s)\left\lVert \begin{bmatrix}
	A_h(s,\wh{\nu}_h)\\
	B_h(s,\wh{\mu}_h)
\end{bmatrix}\theta -\begin{bmatrix}
	c_h(s,\wh{\mu}_h)\\
	d_h(s,\wh{\nu}_h)
\end{bmatrix}\right\rVert^2\leq\kappa_h.
\end{equation*}
We summarize the adapted algorithm for learning reward from actions in Algorithm \hyperref[alg2]{3}.
\begin{algorithm}
\caption{Learning reward from actions with MLE-based QRE estimation}
\begin{algorithmic}[1]\label{alg2}
\Require{Dataset $\mathcal{D} = \{(s_h^t,a_h^t,b_h^t)_{h=1}^H\}_{t\in[T]}$, kernels $\phi(\cdot,\cdot,\cdot)$, $\psi_a(\cdot,\cdot)$, $\psi_b(\cdot,\cdot)$, entropy regularization term $\eta$, discount factor $\gamma$, threshold parameter $(\kappa_h)$, ridge regularization term $\lambda$.}
\Ensure{A confidence set $\wh{\mathcal{R}}$ of the feasible rewards.}
\For {$h\in[H]$}
\State Construct confidence set $\wh\Theta_h$ by \eqref{expconfset}.
\State Compute the feasible Q-functions and V-functions by containing all $\wh{Q}_h$ and $\wh{V}_h$ such that
\begin{equation*}
\begin{aligned}
&\widehat{Q}_h(s,a,b)= \phi(s,a,b)^\top\widehat{\theta}_h\quad\text{where}\ \widehat{\theta}_h\in\widehat{\Theta}_h,\\
&\widehat{V}_{h}(s)= \widehat{\mu}_h(s)^{\top}\widehat{Q}_h(s)\widehat{\nu}_h(s)+\eta^{-1}\mathcal{H}(\widehat{\mu}_h(s))-\eta^{-1}\mathcal{H}(\widehat{\nu}_h(s));
\end{aligned}
\end{equation*}
\State Compute empirical transition by \eqref{mkridge}.
\State Construct confidence set $\wh{\mathcal{R}}$ for feasible rewards by \eqref{mkbelrecover}.
\EndFor
\State\Return $\wh{\mathcal{R}}=\wh{\mathcal{R}}_1\times\wh{\mathcal{R}}_2\times\cdots\times\wh{\mathcal{R}}_H.$
\end{algorithmic}
\end{algorithm}

To evaluate the effectiveness of Algorithm \hyperref[alg2]{3}, we follow the method outlined in Section \ref{sec:3.3}, and establish theoretical guarantees for the feasible sets constructed using this approach. In contrast with the previous confidence set derived from frequency-based QRE estimation, we  consider the refined confidence set \eqref{expconfset} that incorporates MLE-based QRE estimation, leading to improved convergence guarantees.
\begin{lemma}\label{qestmle}
Under Assumptions \ref{alin} and \ref{policysoftmaxpara}, let $\wh{\Theta}_h$ be the confidence set obtained. Set
\begin{equation*}
    \epsilon^2=\mathcal{O}\left(\frac{1}{T}\left((d_a+d_b)\log T+\log\frac{H}{\delta}+\sqrt{m}+\sqrt{n}\right)\right),
\end{equation*}
and
\begin{equation}
\begin{aligned}
&\kappa_h=\mathcal{O}\biggl(\frac{1}{T}\biggl(m^{7/2}+n^{7/2}+(m^3+n^3)\left((d_a+d_b)\log T+\log\frac{H}{\delta}\right)+S(\log mn)^2\biggr)\biggr).
\end{aligned}\label{thresholdchoice1}
\end{equation} 
Then with probability at least $1-3\delta$, we have 
$$\bbE_{s\sim\rho_h}[\mathrm{TV}^2(\wh\nu_h(\cdot|s),\nu_h^*(\cdot|s))]\leq\epsilon^2,\quad\bbE_{s\sim\rho_h}[\mathrm{TV}^2(\wh\mu_h(\cdot|s),\mu_h^*(\cdot|s))]\leq\epsilon^2,$$
and $\Theta_h\subseteq\wh{\Theta}_h$ for all $h\in[H]$. Furthermore, for each $h\in[H]$, the Hausdorff distance between the feasible set $\Theta_h$ and the confidence set $\wh{\Theta}_h$ satisfies 
\begin{equation}
    d_H(\Theta_h,\widehat\Theta_h)\lesssim\sqrt{\kappa_h}.\label{eq:hausb3}
\end{equation}
\end{lemma}
\proof{Proof.}
    See Appendix \ref{sec:c.4} for the complete proof.
\endproof

The result provides statistical efficiency of confidence set construction and a refined convergence guarantee. In contrast to Lemma \ref{qest}, we only require an expected rather than a uniform bound on QRE estimation error over all $s\in\mathcal{S}$, which ensures a more stable and data-efficient estimation of equilibrium strategies. The inclusion $\Theta_h\subseteq\wh{\Theta}_h$ ensures that all feasible parameters are retained within the confidence set, while the Hausdorff distance bound \eqref{eq:hausb3} quantifies the accuracy of our approximation to the true feasible set.

Building on the convergence guarantees for the MLE-based confidence set in Lemma \ref{qestmle}, we now present  the sample complexity of reward set estimation.
\begin{theorem}\label{mlehausconvg}
Under Assumptions \ref{alin} and \ref{policysoftmaxpara}, we let $\rho_h=d^*_h$ be the stationary distribution associated with the optimal policies $\mu^*$ and $\nu^*$, where $h\in[H]$. Suppose that  matrix
\begin{equation*}
\Psi_h=\bbE_{\rho_h}\left[\phi(s_h,a_h,b_h)\phi(s_h,a_h,b_h)^\top\right]\in\mathbb{R}^{d\times d}
\end{equation*}
is nonsingular for all $h\in[H]$. Let $\mathcal{R}$ be the feasible reward set in Definition \ref{rfeasible}. Given a dataset $$\mathcal{D}=\{\mathcal{D}_h\}_{h\in[H]}= \{\{(s_h^t,a_h^t,b_h^t)\}_{t\in[T]}\}_{h\in[H]},$$ we set $\lambda=\mathcal{O}(1)$, and as in (\ref{thresholdchoice1}), set $\kappa_h=\mathcal{O}(T^{-1})$, and denote by $\widehat{\mathcal{R}}$ the output of Algorithm \hyperref[alg2]{3}. For any $\delta\in(0,1)$, we have that, with probability at least $1-4\delta$,
\begin{align*}
&D_1(\mathcal{R},\wh{\mathcal{R}})\lesssim\frac{1}{\sqrt{T}}\biggl(m^{\frac{7}{4}}+n^{\frac{7}{4}}+(m^{\frac{3}{2}}+n^{\frac{3}{2}}+\log T)\sqrt{(d_a+d_b+d)\log T+\log\frac{H}{\delta}}+\sqrt{Sd}\log(mn)\biggr).
\end{align*}
\end{theorem}
\proof{Proof.}
    See Appendix \ref{sec:b.4} for the complete proof.
\endproof

Theorem \ref{mlehausconvg} provides a finite-sample bound on the Hausdorff distance between the identifiable set of feasible rewards $\mathcal{R}$ and the estimated reward set $\wh{\mathcal{R}}$ obtained via our algorithm with MLE-based QRE estimation. This result  quantifies the sample complexity required for reliable and efficient reward inference in entropy-regularized Markov games.
We point out several key insights emerge from this theorem:
\begin{itemize}
\item\textit{Convergence Rate.}
The bound reveals a convergence rate of roughly $\mathcal{O}(T^{-1/2})$, consistent with optimal statistical results. This ensures that as the number of observed samples $T$ grows, our estimated feasible reward set converges toward the true identifiable set.

\item\textit{Dependence on Problem Complexity.}
Larger state or action spaces require more data to achieve the same accuracy level. In particular, the fractional exponents on action spaces $m$ and $n$ in this bound originate from the estimate for Hellinger distance in (\ref{hellingerpolicybound}) and the application of Cauchy-Schwartz inequality. Since the exponent $7/4$ is still less than $2$, the effect is milder than in worst-case scenarios where complexity scales quadratically or cubically.

\item\textit{Role of Feature Representations.}
The terms involving the dimensions $d,d_a$ and $d_b$,  where $d$ is the dimension of feature $\phi$ for reward and transition modeling, and $d_a,d_b$ are the dimensions of features $\psi_a,\psi_b$ for policy modeling. This involvement emphasizes the impact of the complexity of the feature representation on estimation accuracy. A richer representation (with larger dimensions) increases expressive power but also requires more data to ensure robust recovery.

\item\textit{Implications for Practical Applications.}
This result reassures practitioners that our approach, which leverages maximum likelihood estimation for QRE recovery, provides reliable inference, even when the true reward parameter is partially identified. It establishes clear guidelines on data requirements for accurate reward reconstruction.
\end{itemize}

\section{Numerical Experiments}\label{sec:4}
In this section, we implement our reward-learning algorithm and conduct numerical experiments in both entropy-regularized zero-sum matrix games and Markov games. All experiments are conducted in Google Colab (\citealp{googlecolab}). 

\subsection{Entropy-regularized Zero-sum Matrix Games}
We apply the proposed algorithm for two-player entropy-regularized zero-sum matrix games. We discuss both the strong identifiability and the partial identifiability cases.

We evaluate the performance of our algorithm by three key metrics: (1) the error in parameter estimation, which measures the difference between the estimated reward parameter $\widehat{\theta}$ and the true parameter $\theta^*$; (2) the error in the estimated payoff function $\wh{Q}$, which evaluates how accurately the reconstructed payoff function matches the true payoff function; and (3) the error in the estimated QRE, which quantifies the discrepancy between the QRE $(\wh{\mu},\wh{\nu})$ derived from the estimated payoff function and the true QRE $(\mu^*,\nu^*)$. Among these metrics, we are particularly interested in the error in the estimated QRE, which validates whether the algorithm aligns with the players' behaviors.

\paragraph{Setup I.} Let the kernel function be $\phi:\mathcal{A}\times\mathcal{B}\to\bbR^d$ with dimension $d=2$, and set the true parameter as $$\theta^*=(0.8,-0.6)^\top.$$ We set the sizes of action spaces to be $m=4,\,n=6$. The entropy regularization term is $\eta=0.5$. 

To generate the dataset, we sample i.i.d. pairs $\{(a_i,b_i)\}_{i=1}^N$ from the QRE $(\mu^*,\nu^*)$ corresponding to the true payoff function $Q^*(a,b)=\phi(a,b)^\top\theta^*$. We conduct experiments for sample sizes $N$ varying from $10^3$ to $10^6$, and repeat 100 times for each $N$. 
We implement Algorithm \hyperref[algpff]{1} proposed in Section \ref{sec:siden}. 

\paragraph{Setup II.} Let the kernel function be $\phi:\mathcal{A}\times\mathcal{B}\to\bbR^d$ with dimension $d=6$, and set the true parameter as $$\theta^*=(0.8,-0.6,0.75,0.2, 0.5, -0.5)^\top.$$ Also, we set the size of action spaces as $m=6,\,n=6$. The entropy regularization term is $\eta=0.5$. 

To generate the dataset, we sample i.i.d. pairs $\{(a_i,b_i)\}_{i=1}^N$ from the QRE $(\mu^*,\nu^*)$ corresponding to the true payoff function $Q^*(a,b)=\phi(a,b)^\top\theta^*$. We conduct experiments for sample sizes $N$ varying from $10^3$ to $10^6$, and repeat 100 times for each $N$.

We implement Algorithm \hyperref[a1]{2} proposed in Section \ref{sec:piden}. In each experiment, our algorithm outputs a parameter $\wh{\theta}$ in the confidence set $\wh{\Theta}$. We set the bound of feasible parameters as $M=4$, and set the threshold $\kappa=10^3/N$ in \eqref{thetaconfset}, where $N$ is the sample size. 

\begin{figure*}
    \centering
    \caption{The results of numerical simulation on zero-sum matrix games. Both X and Y axes are log-scaled. The X-axis represents the sample size from $10^3$ to $10^6$. The Y-axis represents
    (a,b) the error $\Vert\wh{\theta}-\theta^*\Vert$ of the estimate $\wh{\theta}$; (c,d) The Y-axis represents the error $\Vert\wh{Q}-Q^*\Vert_{\mathrm{F}}$ of the reward function $\wh{Q}$;
    (e,f) The Y-axis represents the error $\mathrm{TV}(\wh{\mu},\mu^*)+\mathrm{TV}(\wh{\nu},\nu^*)$. We repeat 100 experiments for each sample size and plot 95\% confidence interval for the error.\vspace{0.5cm}}
    \begin{subfigure}[b]{0.475\textwidth} 
    \captionsetup{font=small}
        \includegraphics[width=\textwidth]{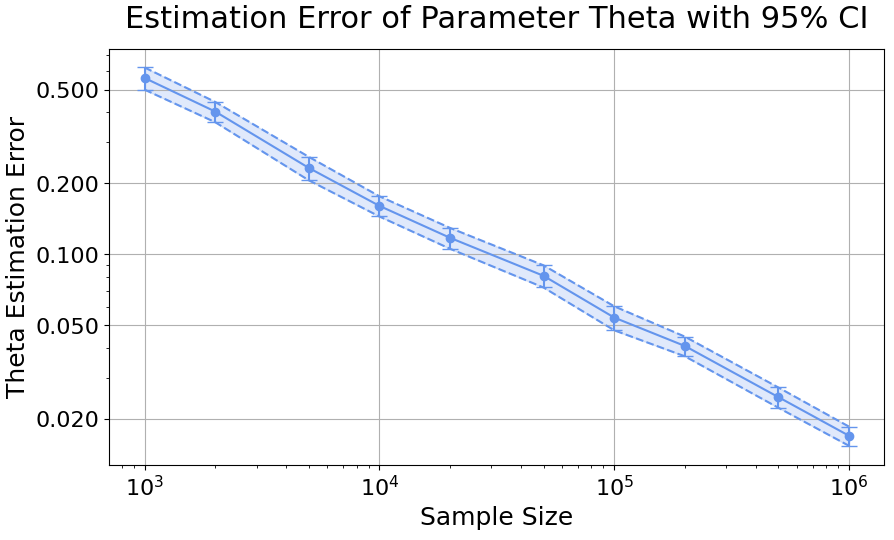}
        \subcaption*{\scriptsize (a) The estimation error of parameter $\theta$ in Setup I}
        \label{simu11}
    \end{subfigure}
    \hfill
    \begin{subfigure}[b]{0.475\textwidth}
    \captionsetup{font=small}
        \includegraphics[width=\textwidth]{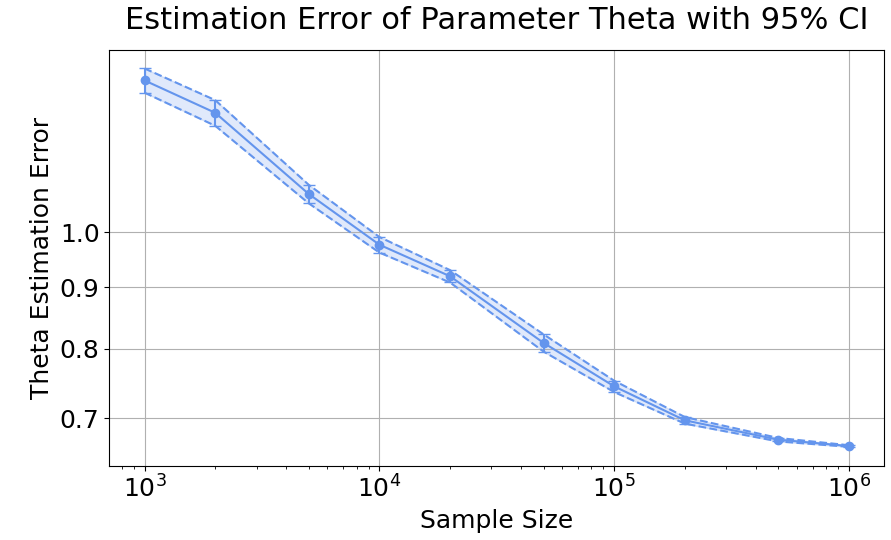}
        \subcaption*{\scriptsize (b) The estimation error of parameter $\theta$ in Setup II}
        \label{simu21}
    \end{subfigure}\vspace{0.2cm}\\
    \begin{subfigure}[b]{0.475\textwidth} 
    \captionsetup{font=small}
        \includegraphics[width=\textwidth]{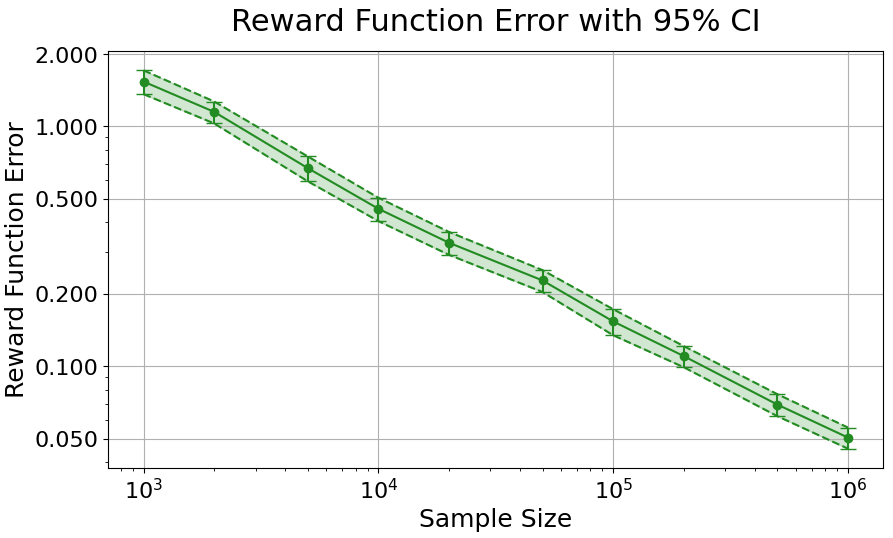}
        \subcaption*{\scriptsize (c) The reconstruction error of payoff matrix $Q$ in Setup I}
        \label{simu12}
    \end{subfigure}
    \hfill
    \begin{subfigure}[b]{0.475\textwidth}
    \captionsetup{font=small}
        \includegraphics[width=\textwidth]{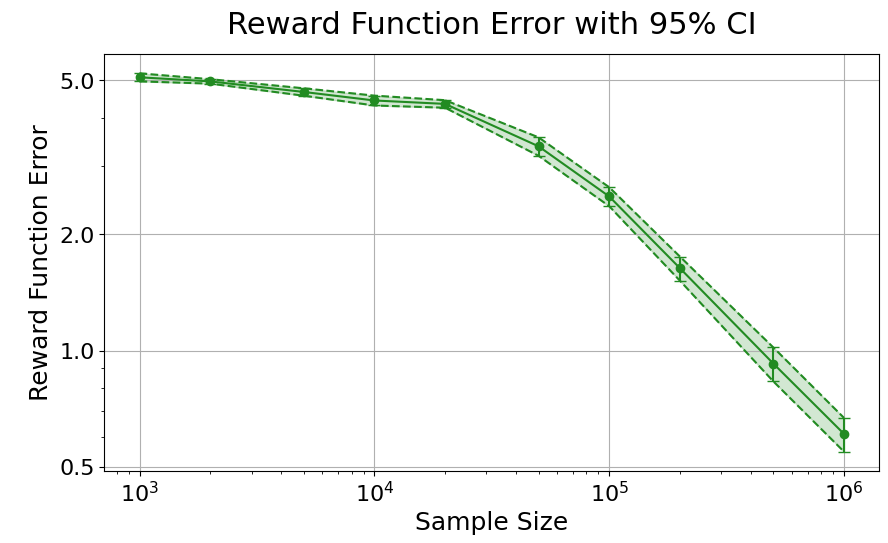}
        \subcaption*{\scriptsize (d) The reconstruction error of payoff matrix $Q$ in Setup II}
        \label{simu22}
    \end{subfigure}\vspace{0.2cm}\\
    \begin{subfigure}[b]{0.475\textwidth} 
    \captionsetup{font=small}
        \includegraphics[width=\textwidth]{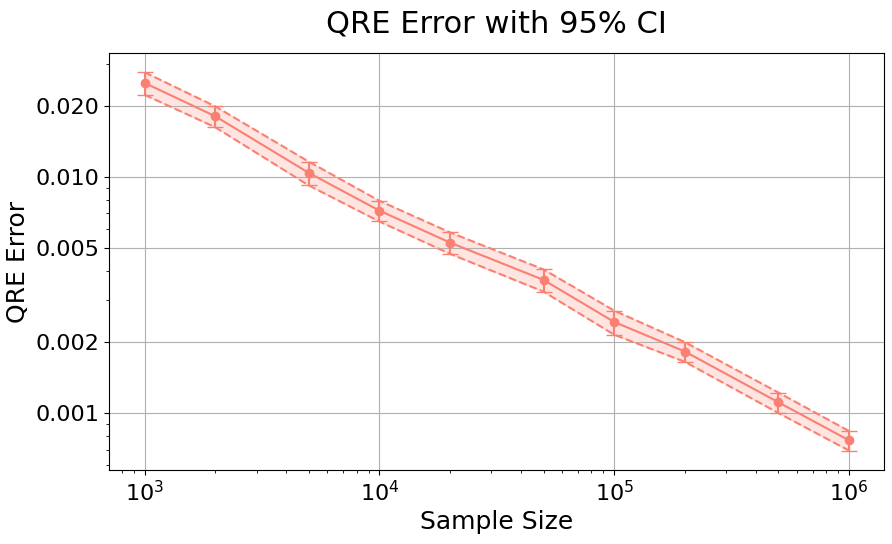}
        \subcaption*{\scriptsize (e) The discrepancy between the QRE $(\wh{\mu},\wh{\nu})$ derived from the estimated payoff $\wh{Q}$ and the true QRE $(\mu^*,\nu^*)$ in Setup I}
        \label{simu13}
    \end{subfigure}
    \hfill
    \begin{subfigure}[b]{0.475\textwidth}
    \captionsetup{font=small}
        \includegraphics[width=\textwidth]{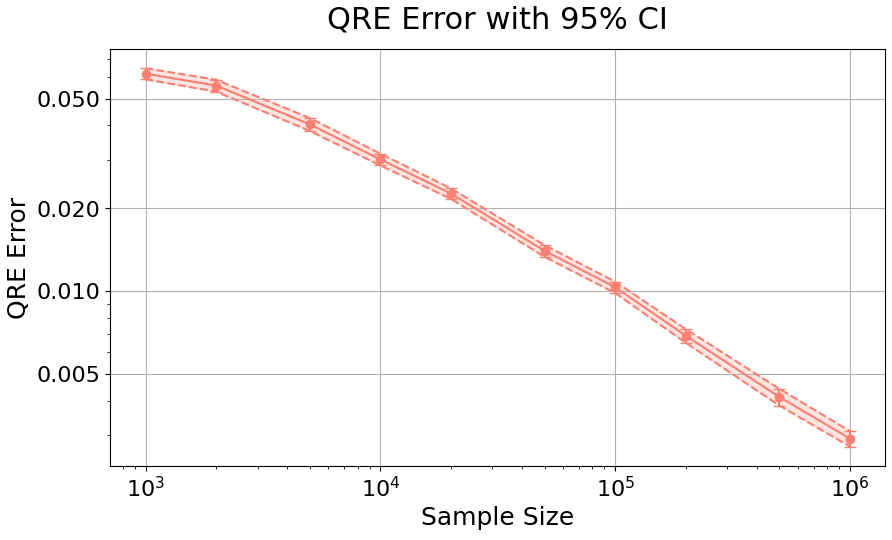}
        \subcaption*{\scriptsize (f) The discrepancy between the QRE $(\wh{\mu},\wh{\nu})$ derived from the estimated payoff $\wh{Q}$ and the true QRE $(\mu^*,\nu^*)$ in Setup II}
        \label{simu23}
    \end{subfigure}
    \label{fig:comparison}
\end{figure*}

\paragraph{Results.} 

In Setup I, the true parameter $\theta^*$ is uniquely identifiable. The results are presented in Figures 1(a), 1(c) and 1(e), where both X and Y axis take the log scale. Figures 1(a) and 1(c) demonstrate that the estimated parameter $\wh\theta$ and the reconstructed payoff matrix $\wh{Q}$ converge to their true values, with the estimation error following an order close to $\mathcal{O}(N^{-1/2})$, which is consistent with our theoretical results. Figure 1(e) shows that the QRE corresponding to our estimated payoff matrix aligns with the true QRE.

In Setup II, the true parameter $\theta^*$ is partially identifiable, meaning that multiple feasible parameters can explain the observed strategies. The results are shown in Figures 1(b), 1(d) and 1(f), where both X and Y axis take the log scale. As expected, Figures 1(b) and 1(d) illustrate that the estimated parameter $\wh\theta$ and the payoff matrix $\wh{Q}$ do not necessarily converge to the true values. Nevertheless, Figure 1(f) shows that the QRE derived from the estimated payoff still converges to the true QRE. This indicates that even when the reward function is not uniquely identifiable, our estimated payoff structure remains a valid explanation of the observed agents' behavior.

\subsection{Entropy-regularized Zero-sum Markov Games}
\paragraph{Setup.} Let the kernel function  be $\phi:\mathcal{A}\times\mathcal{B}\to\bbR^d$ with dimension $d=2$, and set the true parameter $\omega_h$ that specifying reward functions as $$\omega_h^*=(0.8,-0.6)^\top$$
for all steps $h\in[H]$. We set the sizes of action spaces as $m=5$ and $n=5$, the size of state space $S=4$, and the horizon $H=6$. The entropy regularization term is $\eta=0.5$. 

We implement Algorithm \hyperref[alg2]{3} proposed in Section \ref{sec:3.2}. In each experiment, our algorithm outputs a parameter $\wh{\theta}_h$ in the confidence set $\wh{\Theta}_h$. We set the bound of feasible parameters $\theta_h$ as $R=10$, and set the threshold $\kappa_h=10^3/N$ in \eqref{mkconfset}, where $N$ is the sample size. The regularization term in ridge regression is $\lambda=0.01$. 

\paragraph{Metrics.} We evaluate the performance of our algorithm using two metrics: (1) the error in the estimated reward function $(\wh{r}_h)$, which measures how accurately the reconstructed payoff function matches the true reward function; and (2) the error in the estimated QRE, which quantifies the discrepancy between the QRE $(\wh{\mu},\wh{\nu})$ derived from the estimated payoff function and the true QRE $(\mu^*,\nu^*)$. We are particularly interested in the error in the estimated QRE, which validates whether the reconstructed reward functions interpret the observed strategy.

\begin{figure*}\label{simumk}
    \centering
    \caption{The result of simulation in entropy-regularized zero-sum Markov games. (a) The reconstruction error of the reward functions $(\wh{r}_h)_{h=1}^6$. The X-axis represents the time step $h$ from $1$ to $6$, while the Y-axis represents the error $\Vert\wh{r}_h-r^*_h\Vert_{\mathrm{F}}$ of the reward function $\wh{r}$. (b) The discrepancy between the QRE $(\wh{\mu},\wh{\nu})$ corresponding to the estimated reward functions $(\wh{r}_h)_{h=1}^6$ and the true QRE $(\mu^*,\nu^*)$. The X-axis represents the time step $h$ from $1$ to $6$, while the Y-axis represents the errors $\mathrm{TV}(\wh{\mu}_h,\mu^*_h)+\mathrm{TV}(\wh{\nu}_h,\nu^*_h)$. We repeat 100 experiments for each sample size and plot 95\% confidence interval for the error.\vspace{0.5cm}}
    \begin{subfigure}[b]{0.475\textwidth} 
    \captionsetup{font=small}
        \includegraphics[width=\textwidth]{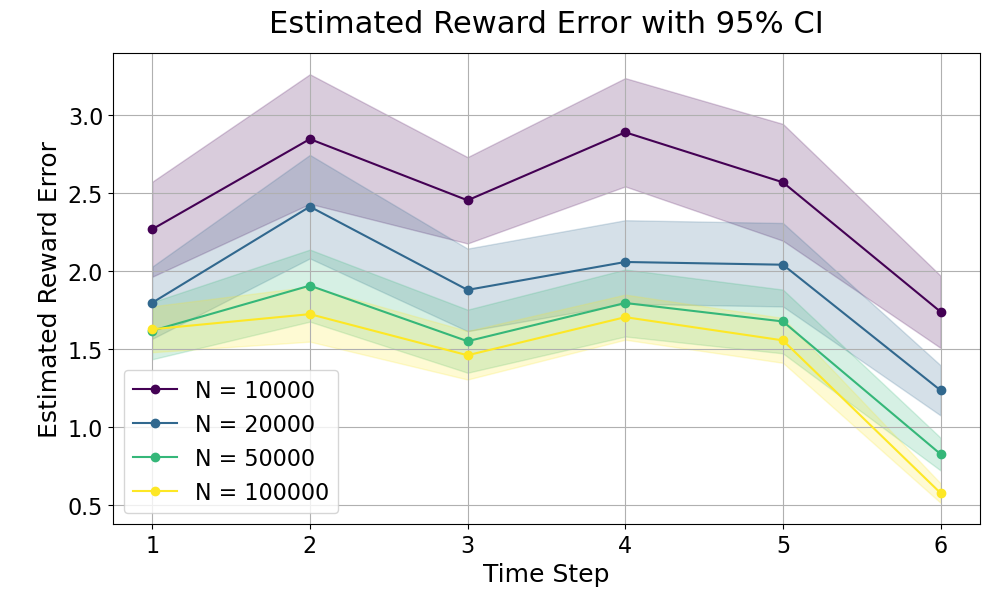}
        \subcaption*{\scriptsize (a) The reconstruction error of reward functions}
        \label{simumk1}
    \end{subfigure}
    \hfill
    \begin{subfigure}[b]{0.475\textwidth}
    \captionsetup{font=small}
        \includegraphics[width=\textwidth]{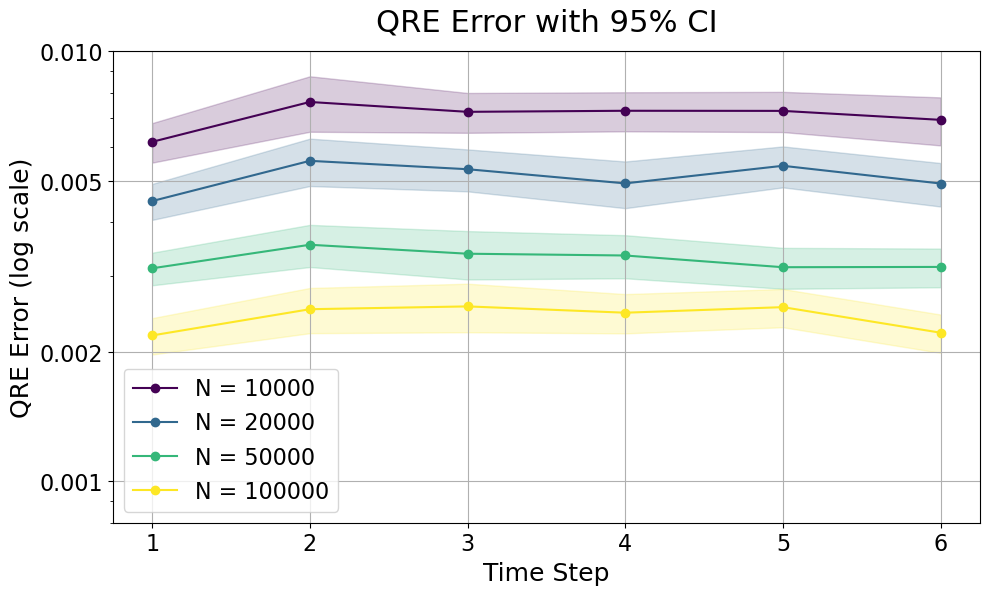}
        \subcaption*{\scriptsize (b) The error of estimated QRE}
        \label{simumk2}
    \end{subfigure}
\end{figure*}

\begin{table*}
\centering
\caption{The mean errors and 95\% confidence intervals for reward and QRE estimation over 100 repetitions in the Markov game setting, across all time steps.\vspace{0.5cm}}
\begin{tabular}{|c|cc|cc|}
\hline
\multirow{2}{*}{\textbf{Sample Size}} & \multicolumn{2}{c|}{\textbf{Reward Error}} & \multicolumn{2}{c|}{\textbf{QRE Error}} \\
\cline{2-5}
 & Mean & 95\% CI & Mean & 95\% CI \\
\hline
10,000  & 2.4611 & $\pm$ 0.1596 & $7.08 \times 10^{-3}$ & $\pm\,4.61 \times 10^{-4}$ \\
20,000  & 1.9031 & $\pm$ 0.1048 & $5.11 \times 10^{-3}$ & $\pm\,3.11 \times 10^{-4}$ \\
50,000  & 1.5609 & $\pm$ 0.0663 & $3.28 \times 10^{-3}$ & $\pm\,1.70 \times 10^{-4}$ \\
100,000 & 1.4398 & $\pm$ 0.0499 & $2.41 \times 10^{-3}$ & $\pm\,1.41 \times 10^{-4}$ \\
\hline
\end{tabular}
\label{tab:markov_experiment}
\end{table*}

\paragraph{Results.} As shown in Figure 2 and Table \ref{tab:markov_experiment}, the overall error of our algorithm's output decreases as the sample size $N$ increases from $10^4$  to $10^5$, demonstrating the improved accuracy of our approach with more data. While the estimation error of reward functions $(\wh{r}_h)_{h=1}^6$ can be relatively large, the corresponding QRE $(\wh{\mu}_h,\wh{\nu}_h)$ remains well-aligned with the true QRE $(\mu^*_h,\nu^*_h)$. Although some fluctuations are observed across time steps, the error remains small, especially for larger sample sizes. These results confirm that our method for reward estimation in Markov games is both statistically consistent and sample-efficient.

\section{Conclusion}\label{sec:5}
To conclude, we explore the challenge of recovering reward functions that explain agents' behavior in competitive games, with a focus on the entropy-regularized zero-sum setting. We propose a framework of inverse game theory concerning the underlying reward mechanisms driving observed behaviors, which applies to both the static setting (Section \ref{sec:2}) and the dynamic setting (Section \ref{sec:3}).

Under a linear assumption, we develop a novel approach for the identifiability of the parameter specifying the current-time payoff. To move forward, we develop an offline algorithm unifying QRE estimation, confidence set construction, transition kernel estimation, and reward recovery, and establish its convergence properties under regular conditions. Additionally, we adapt this algorithm to incorporate a maximum likelihood estimation approach and provide theoretical guarantees for the adapted version. Our algorithms are reliable and effective in both static and dynamic settings, even in the presence of high-dimensional parameter spaces or rank deficiencies. 

Future directions include exploring more complex game settings, such as partially observable games, non-linear payoff functions, and extending the framework to online learning settings. This line of research contributes to the broader effort to make competitive systems more interpretable, offering new insights at the intersection of game theory and reinforcement learning.
\section*{Acknowledgments}
Ethan Fang is partially supported by NSF grants DMS-2346292 and DMS-2434666.

\newpage
\bibliographystyle{ims}
\bibliography{reference}

\newpage 
\appendix
\section{Proof of Entropy-Regularized Matrix Games}

\subsection{Proof of Proposition \ref{p3.2}}
\phantomsection\label{proofprop1}
\proof{Proof.} When $Q=Q_\theta$ is linearly parameterized by the parameter $\theta\in\bbR^d$ according to Assumption \ref{a3.1}, the QRE constraints \eqref{qreconstraints} is equivalent to the linear constraints \eqref{e3.1}. That is, $\theta$ satisfies 
\begin{equation}
    \begin{cases}
        \displaystyle\mu^*(a) = \frac{e^{\eta Q_\theta(a,\cdot)\nu^*}}{\sum_{a\in\mathcal{A}}e^{\eta Q_\theta(a,\cdot)\nu^*}},&\text{for all }a\in\mathcal{A},\vspace{0.2cm}\\
        \displaystyle\nu^*(b) = \frac{e^{-\eta  Q_\theta(\cdot,b)^\top\mu^*}}{\sum_{b\in\mathcal{B}}e^{-\eta  Q_\theta(\cdot,b)^\top\mu^*}},&\text{for all }b\in\mathcal{B}.
    \end{cases}
    \label{qreconstraints1}
\end{equation}
if and only if $\theta$ solves \eqref{e3.1}. By our Assumption \ref{a3.1}, since $(\mu^*,\nu^*)$ are the QRE for the payoff $Q=Q_{\theta^*}$, both the fixed point equation \eqref{qreconstraints1} about $\theta$ and the linear system \eqref{e3.1} has at least one solution $\theta=\theta^*$, which is the true parameter. Therefore it suffices to discuss the uniqueness of solutions to \eqref{e3.1}.

If the rank condition \eqref{eq3.2} holds, by the rank-nullity theorem, the null space of the coefficient matrix $X:=[A(\nu^*)^\top,B(\mu^*)^\top]^\top$ is $\{0\}$. If both $\theta^*,\wt\theta\in\bbR^d$ solve \eqref{e3.1}, we have $X(\wt\theta-\theta^*)=0$, and  $\wt\theta-\theta^*=0$. Conversely, if \eqref{eq3.2} does not hold and $\wt\theta\in\bbR^d$ is a solution to \eqref{e3.1}, we can always take $\xi\neq 0$ from the null space of $X$ and construct another solution $\wt\theta+\xi$ to \eqref{e3.1}. Thus we conclude the proof.
\endproof
\subsection{Proof of Theorem \ref{mxe}}
\phantomsection\label{pmxe}
\proof{Proof.}
To begin with, we decompose $\lVert\widehat{\theta}-\theta^*\rVert^2$ as
    \begin{equation*}
        \begin{aligned}
            \lVert\widehat{\theta}-\theta^*\rVert^2 &\leq \biggl\lVert\left(A(\widehat{\nu})^\top A(\widehat{\nu})+B(\widehat{\mu})^\top B(\widehat{\mu})\right)^{-1}\\
            &\quad\underbrace{\times\begin{bmatrix}
    A(\widehat{\nu})^\top c(\widehat{\mu})+B(\widehat{\mu})^\top d(\widehat{\nu})-A(\nu^*)^\top c(\mu^*)-B(\mu^*)^\top d(\nu^*)
\end{bmatrix}\biggr\rVert^2}_{\displaystyle \mathrm{(I)}}\\
&\quad+\biggl\lVert\left[\left(A(\widehat{\nu})^\top A(\widehat{\nu})+B(\widehat{\mu})^\top B(\widehat{\mu})\right)^{-1}-\left(A(\nu^*)^\top A(\nu^*)+B(\mu^*)^\top B(\mu^*)\right)^{-1}\right]\\
&\qquad\underbrace{\qquad\qquad\qquad\qquad\qquad\qquad\qquad\quad\times\begin{bmatrix}
    A(\nu^*)^\top c(\mu^*)+B(\mu^*)^\top d(\nu^*)
\end{bmatrix}\biggr\rVert^2}_{\displaystyle \mathrm{(II)}}.
        \end{aligned}
    \end{equation*}
    \paragraph{Bounding (I).} 
    \begin{equation}\label{e3.4}
        \begin{aligned}
            \displaystyle \mathrm{(I)}\lesssim &\left\lVert\left(A(\widehat{\nu})^\top A(\widehat{\nu})+B(\widehat{\mu})^\top B(\widehat{\mu})\right)^{-1}\right\rVert^2_{\text{op}}\\
            &\times\left[\lVert A(\widehat{\nu})^\top c(\widehat{\mu})-A(\nu^*)^\top c(\mu^*)\rVert^2+\lVert B(\widehat{\mu})^\top d(\widehat{\nu})-B(\mu^*)^\top d(\nu^*)\rVert^2\right]
        \end{aligned}
    \end{equation}
    We bound the three terms in the RHS of \eqref{e3.4} separately. To begin with, we have
    \[\lVert A(\widehat{\nu})^\top c(\widehat{\mu})-A(\nu^*)^\top c(\mu^*)\rVert^2\lesssim \underbrace{\lVert (A(\widehat{\nu})-A(\nu^*))^\top c(\widehat{\mu})\rVert^2}_{\displaystyle \mathrm{(1)}}+\underbrace{\lVert A(\nu^*)^\top (c(\widehat{\mu})-c(\mu^*))\rVert^2}_{\displaystyle \mathrm{(2)}}.\]
    First, by the definition of $c(\mu) = (\log(\mu_i/\mu_1)/\eta)_{i\in[m]/\{1\}}$, we have
    \begin{equation}\label{e3.5}
        \begin{aligned}
            \displaystyle \mathrm{(2)}&\leq \lVert A(\nu^*)\rVert_{\text{op}}^2\cdot\sum_{i = 2}^m(\log(\widehat{\mu}_i/\widehat{\mu}_1-\log(\mu_i^*/\mu_1^*)))^2/\eta^2\\
            &\leq 2\lVert A(\nu^*)\rVert_{\text{op}}^2 \cdot\left[(m-2)(\log(\widehat{\mu}_1)-\log(\mu_1^*))^2+\sum_{i = 1}^m(\log(\widehat{\mu}_i)-\log(\mu_i^*))^2\right]/\eta^2.
        \end{aligned}
    \end{equation}
    Recall that $A(\nu^*) = (\left(\phi(i,\cdot)-\phi(1,\cdot)\right)\nu^*)_{i\in[m]/\{1\}}$, which can be rewritten as
    \[A(\nu^*)^\top = \Phi_1\cdot(I_{m-1}\otimes\nu^*),\]
    where $\Phi_1 = \left(\phi(i,\cdot)-\phi(1,\cdot)\right)_{i\in[m]\backslash\{1\}}\in\mathbb{R}^{d\times (m-1)n}$, and $\otimes$ denotes the Kronecker product and $I_{m-1}\otimes\nu^*\in\mathbb{R}^{(m-1)n\times (m-1)}$. Therefore, we have
    \begin{equation}\label{e3.6}
        \lVert A(\nu^*)\rVert_{\text{op}}^2\leq\lVert \Phi_1\rVert_{\text{op}}^2\cdot\lVert\nu^*\rVert^2,
    \end{equation}
    where we use the fact that $\lVert I_{m-1}\otimes\nu\rVert_{\text{op}}^2 = \lVert\nu\rVert^2$. Besides, we have that
    \begin{equation}
        \begin{aligned}
            \sum_{i = 1}^m(\log(\widehat{\mu}_i)-\log(\mu_i^*))^2&= \sum_{i = 1}^m\left(\log\left(1+\frac{\widehat{\mu}_i-\mu_i^*}{\mu_i^*}\right)\right)^2\\
            &\leq \sum_{i = 1}^m\max\left\{\left(\frac{\widehat{\mu}_i-\mu_i^*}{\widehat{\mu}_i}\right)^2,\left(\frac{\widehat{\mu}_i-\mu_i^*}{\mu_i^*}\right)^2\right\},
        \end{aligned}\label{logdiffest}
    \end{equation}
    where we use the inequality $x/(1+x)\leq\log(1+x)\leq x$ for all $x>-1$. Moreover, we remark that $|\widehat{\mu}_i-\mu_i^*|\leq\text{TV}(\widehat{\mu},\mu^*)/2\leq \epsilon_1/4$, and we have
    \[\sum_{i = 1}^m(\log(\widehat{\mu}_i)-\log(\mu_i^*))^2\leq\sum_{i = 1}^m\frac{\epsilon_1^2/16}{\min\{\widehat{\mu}_i,\mu_i^*\}^2}\leq\frac{m\cdot\epsilon_1^2}{16(\min_{i\in[m]}\mu_i^*-\epsilon_1)^2}.\]
    Plugging these inequalities into \eqref{e3.5}, we obtain
    \begin{equation}\label{e3.7}
        \displaystyle \mathrm{2)}\lesssim\frac{m\cdot\lVert \Phi_1\rVert_{\text{op}}^2\cdot\lVert\nu\rVert^2\cdot\epsilon_1^2}{\eta^2\cdot(\min_{i\in[m]}\mu_i-\epsilon_1)^2}.
    \end{equation}
    To derive the bound for $\displaystyle\mathrm{(1)}$, we consider an equivalent form of $A(\nu^*)$ that
    \[(A(\widehat{\nu})-A(\nu^*))^\top = \Phi_1\cdot(I_{m-1}\otimes (\widehat{\nu}-\nu^*)).\]
    Plugging in this identity, we obtain
    \begin{equation}\label{e3.8}
        \begin{aligned}
            \displaystyle \mathrm{(1)} &\leq \lVert \Phi_1\rVert_{\text{op}}^2\cdot\lVert I_{m-1}\otimes (\widehat{\nu}-\nu^*)\rVert_{\text{op}}^2\cdot\lVert c(\widehat{\mu})\rVert^2\\
            &\lesssim \lVert \Phi_1\rVert_{\text{op}}^2\cdot\lVert\widehat{\nu}-\nu\rVert^2\cdot (\lVert c(\mu^*)\rVert^2+\lVert c(\widehat{\mu})-c(\mu^*)\rVert^2)\\
            &\lesssim \lVert \Phi_1\rVert_{\text{op}}^2\cdot\epsilon_2^2\cdot\left(\lVert c(\mu^*)\rVert^2+\frac{m\cdot\epsilon_1^2}{\eta^2\cdot(\min_{i\in[m]}\mu_i^*-\epsilon_1)^2}\right).
        \end{aligned}
    \end{equation}
    Combining \eqref{e3.7} and \eqref{e3.8}, we have
    \begin{equation}\label{e3.9}
        \begin{aligned}
            \lVert A(\widehat{\nu})^\top c(\widehat{\mu})-A(\nu^*)^\top c(\mu^*)\rVert^2&\lesssim\lVert \Phi_1\rVert_{\text{op}}^2\cdot\epsilon_2^2\cdot\left(\lVert c(\mu^*)\rVert^2+\frac{m\cdot\epsilon_1^2}{\eta^2\cdot(\min_{i\in[m]}\mu_i^*-\epsilon_1)^2}\right)\\
    &\quad +\frac{m\cdot\lVert \Phi_1\rVert_{\text{op}}^2\cdot\lVert\nu\rVert^2\cdot\epsilon_1^2}{\eta^2\cdot(\min_{i\in[m]}\mu_i^*-\epsilon_1)^2}\\
    &\lesssim \epsilon_2^2\cdot\lVert c(\mu^*)\rVert^2+\frac{m\cdot\epsilon_1^2\cdot(\epsilon_2^2+1)}{\eta^2\cdot(\min_{i\in[m]}\mu_i^*-\epsilon_1)^2}.
        \end{aligned}
    \end{equation}
    By the symmetric of $\lVert A(\widehat{\nu})^\top c(\widehat{\mu})-A(\nu^*)^\top c(\mu^*)\rVert^2$ and $\lVert B(\widehat{\mu})^\top d(\widehat{\nu})-B(\mu^*)^\top d(\nu^*)\rVert^2$, we have
    \begin{equation}\label{e3.10}
        \begin{aligned}
            \lVert B(\widehat{\mu})^\top d(\widehat{\nu})-B(\mu^*)^\top d(\nu^*)\rVert^2&
    &\lesssim \epsilon_1^2\cdot\lVert d(\nu^*)\rVert^2+\frac{n\cdot\epsilon_2^2\cdot(\epsilon_1^2+1)}{\eta^2\cdot(\min_{i\in[n]}\nu_i^*-\epsilon_2)^2}.
        \end{aligned}
    \end{equation}
    Next, we derive the bound for $\left\lVert\left(A(\widehat{\nu})^\top A(\widehat{\nu})+B(\widehat{\mu})^\top B(\widehat{\mu})\right)^{-1}\right\rVert^2_{\text{op}}$. To begin with, we have
    \begin{equation}\label{e3.11}
        \begin{aligned}
            &\left\lVert\left(A(\widehat{\nu})^\top A(\widehat{\nu})+B(\widehat{\mu})^\top B(\widehat{\mu})\right)^{-1}\right\rVert^2_{\text{op}}\\
            &\ \lesssim \left\lVert\left(A(\nu^*)^\top A(\nu^*)+B(\mu^*)^\top B(\mu^*)\right)^{-1}\right\rVert^2_{\text{op}}\\
    &\quad +\left\lVert\left(A(\widehat{\nu})^\top A(\widehat{\nu})+B(\widehat{\mu})^\top B(\widehat{\mu})\right)^{-1}-\left(A(\nu^*)^\top A(\nu^*)+B(\mu^*)^\top B(\mu^*)\right)^{-1}\right\rVert^2_{\text{op}}.
    \end{aligned}
    \end{equation}
    To simplify the notation, we let $U = A(\nu^*)^\top A(\nu^*)+B(\mu^*)^\top B(\mu^*)$ and $V(\widehat{\mu},\widehat{\nu}) = (A(\widehat{\nu})^\top A(\widehat{\nu})+B(\widehat{\mu})^\top B(\widehat{\mu}))-(A(\nu^*)^\top A(\nu^*)+B(\mu^*)^\top B(\mu^*))$. Then,  \eqref{e3.11} can be rewritten as 
    \[\left\lVert\left(A(\widehat{\nu})^\top A(\widehat{\nu})+B(\widehat{\mu})^\top B(\widehat{\mu})\right)^{-1}\right\rVert^2_{\text{op}}\lesssim \lVert U^{-1}\rVert_{\text{op}}^2+\lVert (U+V(\widehat{\mu},\widehat{\nu}))^{-1}-U^{-1}\rVert_{\text{op}}^2.\]
   By the Woodbury formula, we have
    \[(U+V(\widehat{\mu},\widehat{\nu}))^{-1} - U^{-1} = -U^{-1}(I+V(\widehat{\mu},\widehat{\nu})U^{-1})^{-1}V(\widehat{\mu},\widehat{\nu})U^{-1},\]
    which further implies that
    \begin{equation*}
    \begin{aligned}
    &\left\lVert\left(A(\widehat{\nu})^\top A(\widehat{\nu})+B(\widehat{\mu})^\top B(\widehat{\mu})\right)^{-1}\right\rVert^2_{\text{op}}\\
    &\quad\lesssim\lVert U^{-1}\rVert_{\text{op}}^2+\lVert U^{-1}\rVert_{\text{op}}^4\left\lVert V(\widehat{\mu},\widehat{\nu})\right\rVert_{\text{op}}^2\left\lVert(I+V(\widehat{\mu},\widehat{\nu})U^{-1})^{-1}\right\rVert_{\text{op}}^2.
    \end{aligned}
    \end{equation*}
    We bound $\lVert U^{-1}\rVert_{\text{op}}^2,\lVert V(\widehat{\mu},\widehat{\nu})\rVert_{\text{op}}^2,\lVert(I+V(\widehat{\mu},\widehat{\nu})U^{-1})^{-1}\rVert_{\text{op}}^2$ in what follows. To begin with, by Weyl's inequality, we have $\sigma_{\min}(U)\geq \sigma_{\min}(A(\nu^*)^\top A(\nu^*))+\sigma_{\min}(B(\mu^*)^\top B(\mu^*))$ and
    \[\lVert U^{-1}\rVert_{\text{op}}^2 = \frac{1}{\sigma_{\min}(U)^2} \leq \frac{1}{(\sigma_{\min}(A(\nu^*))^2+\sigma_{\min}(B(\mu^*))^2)^2}.\]
    For $\lVert V(\widehat{\mu},\widehat{\nu})\rVert_{\text{op}}^2 = \lVert (A(\widehat{\nu})^\top A(\widehat{\nu})+B(\widehat{\mu})^\top B(\widehat{\mu}))-(A(\nu^*)^\top A(\nu^*)+B(\mu^*)^\top B(\mu^*))\rVert_{\text{op}}^2$, we have
    \begin{equation*}
        \begin{aligned}
            \lVert V(\widehat{\mu},\widehat{\nu})\rVert_{\text{op}}^2&\lesssim \lVert A(\widehat{\nu})^\top A(\widehat{\nu})-A(\nu^*)^\top A(\nu^*)\rVert_{\text{op}}^2+\lVert B(\widehat{\mu})^\top B(\widehat{\mu})-B(\mu^*)^\top B(\mu^*)\rVert_{\text{op}}^2\\
            &\lesssim \lVert (A(\widehat{\nu})-A(\nu^*))^\top A(\widehat{\nu})\rVert_{\text{op}}^2+\lVert A(\nu^*)^\top (A(\widehat{\nu})-A(\nu^*))\rVert_{\text{op}}^2\\
            &\quad +\lVert (B(\widehat{\mu})-B(\mu^*))^\top B(\widehat{\mu})\rVert_{\text{op}}^2+\lVert B(\mu^*)^\top (B(\widehat{\mu})-B(\mu^*))\rVert_{\text{op}}^2.
        \end{aligned}
    \end{equation*}
    By \eqref{e3.8}, we have $\lVert A(\widehat{\nu})-A(\nu^*)\rVert_{\text{op}}^2\leq \epsilon_2^2$ and $\lVert B(\widehat{\mu})-B(\mu^*)\rVert_{\text{op}}^2\leq \epsilon_1^2$, which further implies that
    \begin{equation}\label{e3.12}
        \lVert V(\widehat{\mu},\widehat{\nu})\rVert_{\text{op}}^2\lesssim \epsilon_1^2+\epsilon_2^2.
    \end{equation}
    For the term $\lVert(I+V(\widehat{\mu},\widehat{\nu})U^{-1})^{-1}\rVert_{\text{op}}^2$, we have $\lVert(I+V(\widehat{\mu},\widehat{\nu})U^{-1})^{-1}\rVert_{\text{op}}^2 = 1/\sigma_{\min}(I+V(\widehat{\mu},\widehat{\nu})U^{-1})^2$. By the property of the smallest singular value, we obtain
    \begin{equation}\label{e3.13}
    \begin{aligned}
        \sigma_{\min}(I+V(\widehat{\mu},\widehat{\nu})U^{-1}) &= \inf_{\lVert x\rVert=1}\lVert (I+V(\widehat{\mu},\widehat{\nu})U^{-1})x\rVert\geq 1-\lVert V(\widehat{\mu},\widehat{\nu})U^{-1}\rVert_{\text{op}}\\
        &\geq 1-\lVert V(\widehat{\mu},\widehat{\nu})\rVert_{\text{op}}\lVert U^{-1}\rVert_{\text{op}}.
    \end{aligned}
    \end{equation}
    Combining \eqref{e3.4}, \eqref{e3.9}, \eqref{e3.10}, and \eqref{e3.11}, we have
    \begin{equation}\label{e3.14}
        \displaystyle\mathrm{(I)}\lesssim\epsilon_1^2\cdot\left(\lVert d(\nu^*)\rVert^2+\frac{m\cdot(\epsilon_2^2+1)}{\eta^2\cdot(\min_{i\in[m]}\mu_i-\epsilon_1)^2}\right)+\epsilon_2^2\cdot\left(\lVert c(\mu^*)\rVert^2+\frac{n\cdot(\epsilon_1^2+1)}{\eta^2\cdot(\min_{i\in[n]}\nu_i-\epsilon_2)^2}\right).
    \end{equation}
    \paragraph{Bounding (II).}
    Combining \eqref{e3.11}, \eqref{e3.12}, and \eqref{e3.13}, we obtain
    \begin{equation}\label{e3.15}
    \begin{aligned}
        \mathrm{(II)}&\lesssim \left\lVert\left(A(\widehat{\nu})^\top A(\widehat{\nu})+B(\widehat{\mu})^\top B(\widehat{\mu})\right)^{-1}-\left(A(\nu^*)^\top A(\nu^*)+B(\mu^*)^\top B(\mu^*)\right)^{-1}\right\rVert^2_{\text{op}}\\
        &\quad\times\lVert
    A(\nu^*)^\top c(\mu^*)+B(\mu^*)^\top d(\nu^*)\rVert^2\\
        &\lesssim \lVert U^{-1}\rVert_{\text{op}}^4\cdot\lVert V(\widehat{\mu},\widehat{\nu})\rVert_{\text{op}}^2\cdot\lVert(I+V(\widehat{\mu},\widehat{\nu})U^{-1})^{-1}\rVert_{\text{op}}^2\lesssim\epsilon_1^2+\epsilon_2^2.
    \end{aligned}
    \end{equation}
    Finally, combining \eqref{e3.14} and \eqref{e3.15}, we obtain that
    \[\lVert\widehat{\theta}-\theta^*\rVert^2\lesssim \epsilon_1^2\cdot\left(1+m\cdot(\epsilon_2^2+1)\right)+\epsilon_2^2\cdot\left(1+n\cdot(\epsilon_1^2+1)\right),\]
    which concludes the proof.
\endproof
\subsection{Proof of Theorem \ref{mxs}}\phantomsection\label{pmxs}
\proof{Proof.}
Since we use the frequency estimator to estimate the QRE, we have
\begin{equation}
    \begin{aligned}
    \bbE\left[\mathrm{TV}(\wh{\mu},\mu^*)\right]&=\frac{1}{2}\sum_{a\in\mathcal{A}}\bbE\left[\vert\wh{\mu}(a)-\mu^*(a)\vert\right]\leq\frac{1}{2}\sum_{a\in\mathcal{A}}\sqrt{\bbE\left[(\wh{\mu}(a)-\mu^*(a))^2\right]}\\
    &=\frac{1}{2}\sum_{a\in\mathcal{A}}\sqrt{\frac{1}{N}\mu^*(a)(1-\mu^*(a))}\leq \frac{1}{2\sqrt{N}}\sum_{a\in\mathcal{A}}\sqrt{\mu^*(a)}\leq\frac{1}{2}\sqrt{\frac{\vert\mathcal{A}\vert}{N}}.
\end{aligned}\label{tvb1}
\end{equation}
Let $A_1,\cdots,A_N\sim\mathrm{Multinomial}(\mu^*)$ be the i.i.d. actions taken following strategy $\mu^*$. We then write the total variation as
\begin{equation*}
    \mathrm{TV}(\wh{\mu},\mu) = f(A_1,\cdots,A_N)=\frac{1}{2}\sum_{a\in\mathcal{A}}\left\vert\frac{1}{N}\sum_{i=1}^N\mathbf{1}_{\{A_i=a\}} - \mu^*\right\vert
\end{equation*}
Then the function $f:\mathcal{A}^n\to[0,1]$ satisfy the bounded difference property for all $k\in[N]$ that
\begin{equation*}
    \sup_{a_k,a_k^\prime\in\mathcal{A}}\vert f(a_1,\cdots,a_{k-1},a_k,a_{k+1}\cdots,a_N)-f(a_1,\cdots,a_{k-1},a^\prime_k,a_{k+1}\cdots,a_N)\vert\leq\frac{1}{N}.
\end{equation*}
By McDiarmid's inequality, for any $\epsilon>0$, we have
\begin{equation*}
    \bbP\left(\mathrm{TV}(\wh{\mu},\mu^*)-\bbE\left[\mathrm{TV}(\wh{\mu},\mu^*)\right]\geq\epsilon\right)\leq e^{-2N\epsilon^2}.
\end{equation*}
Combining this inequality with (\ref{tvb1}), we obtain that
\begin{equation}
\begin{aligned}
    \bbP\left(\mathrm{TV}(\wh{\mu},\mu^*)\geq\frac{1}{2}\sqrt{\frac{m}{N}}+\sqrt{\frac{\log(2/\delta)}{2N}}\right)\leq\frac{\delta}{2}.
\end{aligned}\label{mcdiarmidbound}
\end{equation}
Similarly, we  bound the total variation between $\wh{\nu}$ and $\nu$ that
\begin{equation*}
    \bbP\left(\mathrm{TV}(\wh{\nu},\nu^*)\geq\frac{1}{2}\sqrt{\frac{n}{N}}+\sqrt{\frac{\log(2/\delta)}{2N}}\right)\leq\frac{\delta}{2}.
\end{equation*}
Following Theorem \ref{mxe}, $\text{TV}(\widehat{\mu},\mu^*)\leq\epsilon_1/2$ and $\text{TV}(\widehat{\nu},\nu^*)\leq\epsilon_2/2$ imply that
    \[\lVert \widehat{\theta}-\theta^*\rVert^2\lesssim \epsilon_1^2\cdot\left(1+m\cdot(\epsilon_2^2+1)\right)+\epsilon_2^2\cdot\left(1+n\cdot(\epsilon_1^2+1)\right)\simeq m\epsilon_1^2+n\epsilon_2^2.\]
Since $Q(a,b) = \phi(a,b)^\top \theta$ for all $a,b\in\mathcal{A}\times \mathcal{B}$, we have
\begin{equation*}
    \begin{aligned}
        \lVert \widehat{Q}-Q\rVert_F^2 &= \sum_{a,b\in\mathcal{A}\times\mathcal{B}} (\widehat{Q}(a,b)-Q(a,b))^2 =  \sum_{a,b\in\mathcal{A}\times\mathcal{B}} (\phi(a,b)^\top(\widehat{\theta}-\theta))^2\\
        &\leq \Big(\sum_{a,b\in\mathcal{A}\times\mathcal{B}}\lVert\phi(a,b)\rVert^2\Big)\lVert \widehat{\theta}-\theta\rVert^2\lesssim m\epsilon_1^2+n\epsilon_2^2.
    \end{aligned}
\end{equation*}
Therefore, for any $\delta\in(0,1)$, we set
\begin{equation}
    \epsilon_1= \frac{\sqrt{m}+\sqrt{2\log(2/\delta)}}{\sqrt{N}},\quad \epsilon_2=  \frac{\sqrt{n}+\sqrt{2\log(2/\delta)}}{\sqrt{N}},\label{tvvalue}
\end{equation}
and obtain the probability bound that
\[\mathbb{P}\left(\lVert \widehat{Q}-Q\rVert_F^2\lesssim \mathcal{O}\left(\frac{m^2+n^2+(m+n)\log(1/\delta)}{N}\right)\right)\geq 1-\delta,\]
which is desired.
\endproof


\subsection{Parameter Selection}\label{sec:2.4}
As discussed in \S\ref{sec:piden}, the true parameter $\theta^*$ is partially identifiable when the rank condition \eqref{eq3.2} does not hold, and there are infinitely many parameters that lead to the same QRE. To avoid unnecessary large coefficients that might overfit or lead to instability, we can further define a canonical choice of solution $\theta^*$ by selecting the parameter vector with the smallest Euclidean norm that satisfies the QRE constraints. This avoids unnecessary large coefficients, mitigates overfitting, and improves numerical stability. In particular, we define the optimal $\theta^*$ as:
\begin{equation*}
\begin{aligned}
\theta^*&= \mathop{\argmin}_{\theta\in\mathbb{R}^d}\,\lVert \theta\rVert^2,\quad
\text{subject to}\  
    \begin{bmatrix}
    A(\nu^*)\\
    B(\mu^*)
\end{bmatrix}\theta = \begin{bmatrix}
    c(\mu^*)\\
    d(\nu^*)
\end{bmatrix}.
\end{aligned}
\end{equation*}
When the constraint matrix is not of full column rank, this minimum-norm solution is uniquely characterized by the Moore–Penrose pseudoinverse (\citealp{ben2006generalized}) that
\[\theta^* = \begin{bmatrix}
    A(\nu^*)\\
    B(\mu^*)
\end{bmatrix}^\dagger\begin{bmatrix}
    c(\mu^*)\\
    d(\nu^*)
\end{bmatrix},\]
where we denote by $M^\dagger$ the Moore-Penrose pseudoinverse of any matrix $M$. 
Therefore, to estimate the optimal parameter $\theta^*$, we adopt the following plug-in estimator that
\[\wh{\theta} = \begin{bmatrix}
    A(\widehat{\nu})\\
    B(\widehat{\mu})
\end{bmatrix}^\dagger\begin{bmatrix}
    c(\widehat{\mu})\\
    d(\widehat{\nu})
\end{bmatrix}.\]
The next theorem quantifies the estimation error $\lVert\widehat{\theta}-\theta^*\rVert$.
\begin{theorem}[Convergence of the optimal QRE solution]\label{lsop}
Assume that the matrix $$X=\begin{bmatrix}A(\nu^*)\\ B(\mu^*)\end{bmatrix}\in\bbR^{(m+n-2)\times d}$$ is of full row rank, and its smallest singular value is bounded from below that $\sigma_{m+n-2}\left(X\right)\geq\sigma_b$ for some $\sigma_b>0$. Given $N$ samples $\{(a^k,b^k)\}_{k\in[N]}$ following the true QRE $(\mu,\nu)$, we obtain $(\widehat{\mu},\widehat{\nu})$ by the frequency estimator. For any $\delta\in(0,1)$, when $N$ is sufficiently large, we have that, with probability at least $1-\delta$,
\begin{equation*}
    \Vert\wh\theta-\theta^*\Vert\lesssim\frac{m^2+n^2}{\sqrt{N}}.
\end{equation*}
\end{theorem}
\proof{Proof.}
See Appendix \ref{plsop}.
\endproof

We note that the assumptions in Theorem \ref{lsop} are stronger than what we need in Theorem \ref{tmxc}, as we require that $d\geq m+n-2$, and $X$ is of full row rank. This assumption is necessary for establishing the convergence of the Moore-Penrose inverse plug-in estimator that
\begin{equation*}
    \begin{bmatrix}A(\wh\nu)\\ B(\wh\mu)\end{bmatrix}^\dagger\to\begin{bmatrix}A(\nu^*)\\ B(\mu^*)\end{bmatrix}^\dagger.
\end{equation*}

We note that when the rank condition (\ref{e3.2}) is not satisfied, the kernel $\phi$ may encode redundant or irrelevant features, leading to an excessively large hidden feature dimension $d$. In such cases, the design matrix $X\in\bbR^{(m+n-2)\times d}$, constructed from these features, is possibly rank-deficient. Therefore, it is reasonable to require that $X$ has linearly independent rows, ensuring that the QRE provides sufficient information to distinguish between different directions in the parameter space.

In practice, selecting the minimum-norm solution helps avoid overfitting and promotes stability (\citealp{hastie2009elements}). The convergence rate $\mathcal{O}(N^{-1/2})$ matches standard results in statistical estimation, showing the reliability and efficiency of our method in practical settings. 

\subsection{Proof of Theorem \ref{lsop}}\phantomsection\label{plsop}
\proof{Proof.}
We first control the error of the Moore-Penrose inverse that
    \begin{equation}
    \begin{aligned}
        \Vert \wh X^\dagger-X^\dagger\Vert_\mathrm{op}&= \left\Vert\wh X^\top(\wh X\wh X^\top)^{-1}-X^\top(XX^\top)^{-1}\right\Vert_\mathrm{op}\\
        &\leq\left\Vert \wh X^\top(\wh X\wh X^\top)^{-1}-\wh X^\top(XX^\top)^{-1}\right\Vert_\mathrm{op}+\left\Vert(\wh X-X)^\top(XX^\top)^{-1}\right\Vert_\mathrm{op}\\
        &\leq\Vert\wh X\Vert_\mathrm{op}\left\Vert(\wh X\wh X^\top)^{-1}(XX^\top-\wh X\wh X^\top)(XX^\top)^{-1}\right\Vert_\mathrm{op}+\Vert\wh X-X\Vert_\mathrm{op}\left\Vert(XX^\top)^{-1}\right\Vert_\mathrm{op}\\
        &\leq\left(\Vert\wh X\Vert_\mathrm{op}\left\Vert(\wh X\wh X^\top)^{-1}\right\Vert_\mathrm{op}\left\Vert XX^\top-\wh X\wh X^\top\right\Vert_\mathrm{op}+\Vert\wh X-X\Vert_\mathrm{op}\right)\left\Vert(XX^\top)^{-1}\right\Vert_\mathrm{op}.
    \end{aligned}\label{mpibound}
    \end{equation}
    Note that
    \begin{equation}
    \begin{aligned}
    \Vert\wh X\Vert_\mathrm{op}^2=\Vert\wh{X}^\top\wh X\Vert_\mathrm{op}&=\Vert A(\wh{\nu})^\top A(\wh\nu)+B(\wh\mu)^\top B(\wh\mu)\Vert_\mathrm{op}\\
    &\leq\Vert\Phi_1(I_{m-1}\otimes\wh{\nu})\Vert_\mathrm{op}^2+\Vert\Phi_2(I_{n-1}\otimes\wh{\mu})\Vert_\mathrm{op}^2\\
    &\leq\Vert\Phi_1\Vert_\mathrm{op}^2+\Vert\Phi_2\Vert_\mathrm{op}^2,
    \end{aligned}\label{hatopsq}
    \end{equation}
    and
    \begin{equation}
        \Vert XX^\top-\wh X\wh X^\top\Vert_\mathrm{op}\leq\Vert X(X-\wh{X})^\top\Vert_\mathrm{op}+\Vert\wh X(X-\wh{X})^\top\Vert_\mathrm{op}\leq 2\sqrt{\Vert\Phi_1\Vert_\mathrm{op}^2+\Vert\Phi_2\Vert_\mathrm{op}^2}\Vert\wh X-X\Vert_\mathrm{op}.\label{hermiteerr}
    \end{equation}
    Since the smallest singular value of $X$ is bounded from below by $\sigma_b>0$, we have
    \begin{align}
    \left\Vert(XX^\top)^{-1}\right\Vert_\mathrm{op}\leq\frac{1}{\sigma_b^2}.\label{invbopt}
    \end{align}
    By Weyl's inequality, for sufficiently small $\epsilon_1,\epsilon_2>0$ with $\Vert\Phi_1\Vert_\mathrm{op}^2\epsilon_2^2+\Vert\Phi_2\Vert_\mathrm{op}^2\epsilon_1^2\leq\sigma_b^2/4$, we have
    \begin{equation*}
    \sigma_{\min} (\wh{X})\geq\sigma_{\min}(X)-\Vert\wh{X}-X\Vert_\mathrm{op}\geq\sigma_b-\sqrt{\Vert\Phi_1\Vert_\mathrm{op}^2\epsilon_2^2+\Vert\Phi_2\Vert_\mathrm{op}^2\epsilon_1^2}\geq\frac{\sigma_b}{2}.
    \end{equation*}
    Hence, we have
    \begin{equation}
    \left\Vert(\wh X\wh X^\top)^{-1}\right\Vert_\mathrm{op}\leq\frac{4}{\sigma_b^2}.\label{invbhat}   
    \end{equation}
    Combining (\ref{mpibound}), (\ref{hatopsq}), (\ref{hermiteerr}), (\ref{invbopt}) and (\ref{invbhat}), we have
    \begin{equation}
        \Vert \wh X^\dagger-X^\dagger\Vert_\mathrm{op}\leq\left(\frac{8(\Vert\Phi_1\Vert_\mathrm{op}^2+\Vert\Phi_2\Vert_\mathrm{op}^2)}{\sigma_b^4}+\frac{1}{\sigma_b^2}\right)\sqrt{\Vert\Phi_1\Vert_\mathrm{op}^2\epsilon_2^2+\Vert\Phi_2\Vert_\mathrm{op}^2\epsilon_1^2}.\label{mpinvopbound}
    \end{equation}
Meanwhile, as in the proof of Theorem \ref{mxe}, we have
\begin{equation}\label{yadiff}
\begin{aligned}
\lVert\widehat{y}-y\rVert^2 & =\lVert c(\widehat{\nu})-c(\nu^*)\rVert^2+\lVert d(\widehat{\mu})-d(\mu^*)\rVert^2 \\
&\leq\frac{m\epsilon_1^2}{\eta^2(\min_{i\in[m]}\mu_i-\epsilon_1)^2}+\frac{n\epsilon_2^2}{\eta^2(\min_{j\in[n]}\nu_j-\epsilon_2)^2}.
\end{aligned}
\end{equation}
Combining 
(\ref{mpinvopbound}) and (\ref{yadiff}), we have that
\begin{equation*}
\begin{aligned}
    \Vert\wh\theta-\theta^*\Vert&\leq\Vert X^\dagger-\wh X^\dagger\Vert_\mathrm{op}\Vert y\Vert + \Vert \wh X^\dagger\Vert_\mathrm{op}\Vert y-\wh y\Vert\\
    &\leq \frac{(8\Vert\Phi_1\Vert_\mathrm{op}^2+8\Vert\Phi_2\Vert_\mathrm{op}^2+\sigma_b^2)\Vert y\Vert}{\sigma_b^4}\sqrt{\Vert\Phi_1\Vert_\mathrm{op}^2\epsilon_2^2+\Vert\Phi_2\Vert_\mathrm{op}^2\epsilon_1^2}\\
    &\quad +\frac{2}{\sigma_b}\sqrt{\frac{m\epsilon_1^2}{\eta^2(\min_{i\in[m]}\mu_i-\epsilon_1)^2}+\frac{n\epsilon_2^2}{\eta^2(\min_{j\in[n]}\nu_j-\epsilon_2)^2}}.
\end{aligned}
\end{equation*}
By our proof in Appendix \ref{pmxs}, with probability at least $1-\delta$, the bound (\ref{tvvalue}) holds for the total variation error between optimal and empirical policies. Plugging  (\ref{tvvalue}) into the last inequality, we have
\begin{equation*}
    \Vert\wh\theta-\theta^*\Vert\lesssim\frac{m^2+n^2}{\sqrt{N}},
\end{equation*}
which concludes the proof.
\endproof

\section{Proof of Entropy-Regularized Markov Games}

\subsection{Proof of Proposition \ref{p3.3}}\phantomsection\label{pmgc}
\proof{Proof.}
For any $h\in[H]$ and $s\in\mathcal{S}$, recall that the QRE constraint is
\begin{equation*}
\begin{cases}
\mu_h^*(a|s) = \frac{e^{\eta\langle Q_h^*(s,a,\cdot),\nu_h^*(\cdot|s)\rangle_{\mathcal{B}}}}{\sum_{a\in\mathcal{A}}e^{\eta\langle Q_h^*(s,a,\cdot),\nu_h^*(\cdot|s)\rangle_{\mathcal{B}}}}, &\text{for all}\ a\in\mathcal{A},\\
\nu_h^*(b|s) = \frac{e^{-\eta \langle Q_h^*(s,\cdot,b),\mu^*_h(\cdot|s)\rangle_{\mathcal{A}}}}{\sum_{b\in\mathcal{B}}e^{-\eta\langle Q_h^*(s,\cdot,b),\mu^*_h(\cdot|s)\rangle_{\mathcal{A}}}}, &\text{for all}\ b\in\mathcal{B}.
\end{cases}
\end{equation*}
    This non-linear constraint is equivalent to the linear system that
    \begin{equation*}
    \left\{
    \begin{aligned}
        &\left(Q_h(s,a,\cdot)-Q_h(s,1,\cdot)\right)\nu_h^*(\cdot|s) = \log(\mu_h^*(a|s)/\mu_h^*(1|s))/\eta,&\quad&\text{for all }a\in\mathcal{A},\\
        & \left(Q_h(s,\cdot,b)-Q_h(s,\cdot,1)\right)\mu_h^*(\cdot|s) = -\log(\nu_h^*(b|s)/\nu_h^*(1|s))/\eta,&\quad&\text{for all }b\in\mathcal{B}.
    \end{aligned}
    \right.
\end{equation*}
Under Assumption \ref{alin}, these linear equations are equivalent to
\begin{equation*}
    \left\{
    \begin{aligned}
        &\langle\left(\phi(s,a,\cdot)-\phi(s,1,\cdot)\right)\nu_h^*(\cdot|s),\theta_h\rangle = \log(\mu_h^*(a|s)/\mu_h^*(1|s))/\eta,&\quad&\text{for all }a\in\mathcal{A},\\
        & \langle\left(\phi(s,\cdot,b)-\phi(s,\cdot,1)\right)\mu_h^*(\cdot|s),\theta_h\rangle= -\log(\nu_h^*(b|s)/\nu_h^*(1|s))/\eta,&\quad&\text{for all }b\in\mathcal{B}.
    \end{aligned}
    \right.
\end{equation*}
By the definition of $A_h$ and $B_h$ in \eqref{mknotations}, the above linear system can be rewritten as
\begin{equation}
    \begin{bmatrix}
    A_h(s,\nu_h^*)\\
    B_h(s,\mu_h^*)
\end{bmatrix}\theta_h = \begin{bmatrix}
    c_h(s,\mu_h^*)\\
    d_h(s,\nu_h^*)
\end{bmatrix},\label{b1ls}
\end{equation}
where 
\begin{equation*}
\begin{aligned}
c_h(s,\mu_h)&= (\log(\mu_h(a|s)/\mu_h(1|s))/\eta)_{a\in\mathcal{A}/\{1\}}\in\mathbb{R}^{m-1},\\
d_h(s,\nu_h)&= (-\log(\nu_h(b|s)/\nu_h(1|s))/\eta)_{b\in\mathcal{B}/\{1\}}\in\mathbb{R}^{n-1}.
\end{aligned}
\end{equation*}
We can concatenate the linear systems of form \eqref{b1ls} across all $s\in\mathcal{S}$ and obtain
\begin{equation*}
\begin{bmatrix}
    A_h(\nu_h^*)\\
    B_h(\mu_h^*)
\end{bmatrix}\theta_h = \begin{bmatrix}
    c_h(\mu_h^*)\\
    d_h(\nu_h^*)
\end{bmatrix}.
\end{equation*}
Therefore, there exists a unique $\theta_h$ satisfies this linear system for all $s\in\mathcal{S}$ if and only if 
\[\text{rank}\left(\begin{bmatrix}
    A_{h}(\nu_h^*) \\ B_{h}(\mu_h^*)
\end{bmatrix}\right) = d,\] which concludes the proof.
\endproof
\subsection{Proof of Theorem \ref{samp}}\phantomsection\label{psamp}



We first derive some technical results that will be used in the proof.
\paragraph{Ridge Regression Analysis.}
First, given a $\lambda>0$, Algorithm \hyperref[a1]{2} estimates the true vectors $\pi_h(\cdot)$ specifying transition kernels $\bbP_h$ in Assumption \ref{alin} by solving the ridge regression problem that
\begin{equation*}
    \wh\Pi_h=\underset{\Pi_h\in\bbR^{S\times d}}{\mathrm{argmin}}\sum_{t=1}^T\left\Vert\Pi_h\phi(s_h^t,a_h^t,b_h^t)-\delta(s_{h+1}^t)\right\Vert^2+\lambda\Vert\Pi_h\Vert_\mathrm{F}^2,
\end{equation*}
where $\Pi_h=(\pi_h(s))_{s\in\mathcal{S}}^\top\in\bbR^{S\times d}$. Then, we have that
\begin{equation*}
\wh\Pi_h=\sum_{t=1}^T\delta(s_{h+1}^t)\phi(s_h^t,a_h^t,b_h^t)^\top\Lambda_h^{-1},\quad\text{where}\ \ \Lambda_h=\sum_{t=1}^T\phi(s_h^t,a_h^t,b_h^t)\phi(s_h^t,a_h^t,b_h^t)^\top+\lambda I_d.
\end{equation*}
Correspondingly, the estimator for the reward function is 
\begin{equation*}
    \widehat{r}_h(s,a,b) = \widehat{Q}_h(s,a,b)-\gamma\widehat{\mathbb{P}}_h\widehat{V}_{h+1}(s,a,b) = \widehat{Q}_h(s,a,b)-\gamma\phi(s,a,b)^\top\wh\Pi_h^\top\widehat{V}_{h+1}.
\end{equation*}
This is the approximation approach of Algorithm \hyperref[a1]{2}.

\paragraph{Property of estimated V-functions.} By Assumption \ref{alin} and row 3 of Algorithm \hyperref[a1]{2}, for all $h\in[H]$, we have the following bound for estimated V-functions:
\begin{equation}
    -R-\eta^{-1}\log n\leq \wh{V}_{h}(s)\leq R+\eta^{-1}\log m,\quad\forall s\in\mathcal{S}.\label{Vbound}
\end{equation}
We denote by $\wh{\mathcal{V}}_{h}$ the set of V-functions $\wh{V}_{h}$ generated by $\wh\Theta_{h}$. For any $\theta,\theta^\prime\in\wh\Theta_{h}$, we have
\begin{equation*}
\left\vert\wh V_{h}(s;\theta)-\wh V_{h}(s;\theta^\prime)\right\vert=\left\vert\wh\mu_{h}(s)^\top(\wh Q_{h}(s;\theta)-\wh Q_{h}(s;\theta^\prime))\wh\nu_{h}(s)\right\vert\leq\Vert\theta-\theta^\prime\Vert,\quad\forall s\in\mathcal{S}.
\end{equation*}
Consequently, the covering number of $\wh{\mathcal{V}}_{h}$ is upper bounded by the covering number of $\wh\Theta_{h}$. We also note that $\wh\Theta_h$ is contained in the ball $B(0,R)=\{\theta\in\bbR^d:\Vert\theta\Vert\leq R\}$. By Lemma \ref{eucover}, for any $\epsilon>0$, we can find an $\epsilon$-net $\mathcal{N}_\epsilon$ of $(\wh{\mathcal{V}}_{h},\Vert\cdot\Vert_\infty)$ such that
\begin{equation}
    \vert\mathcal{N}_\epsilon\vert\leq\left(1+\frac{2R}{\epsilon}\right)^d.\label{coveringbound}
\end{equation}
Clearly, the same property also applies to the set of feasible V-functions $\mathcal{V}_h$ generated by $\Theta_h$.

\subsubsection{Bounding the Estimation Error of V-functions}\phantomsection\label{sec:b.2.1}
In this subsection, we bound the estimation error of V-functions under the event $\mathcal{E}_2$ given by Corollary \ref{qsetuncertainty}. For any $\wh\theta_h\in\wh\Theta_h$, we can find a feasible $\theta_h\in\Theta_h$ such that $\Vert\wh\theta_h-\theta_h\Vert\leq C_h\sqrt{\kappa_h}$ for some constant $C_h$. Then for the V-functions generated by $\wh\theta_h$ and $\theta_h$, we have
\begin{equation}
\begin{aligned}
    &\vert\wh V_h(s)-V_h(s)\vert\\
    &\leq\Big\vert\wh{\mu}_h(s)^\top\wh Q_h(s)\wh\nu_h(s)+\eta^{-1}\mathcal{H}(\wh\mu_h(s))-\eta^{-1}\mathcal{H}(\wh\nu_h(s))\\
    &\quad-\mu_h^*(s)^\top Q_h(s)\nu_h^*(s)-\eta^{-1}\mathcal{H}(\mu_h^*(s))+\eta^{-1}\mathcal{H}(\nu_h^*(s))\Big\vert\\
    &\leq\underbrace{\left\vert\wh{\mu}_h(s)^\top\wh Q_h(s)\wh\nu_h(s)-\mu_h^*(s)^\top Q_h(s)\nu_h^*(s)\right\vert}_{\displaystyle\text{(i)}}+\underbrace{\eta^{-1}\bigl\vert\mathcal{H}(\wh\mu_h(s))-\mathcal{H}(\mu_h^*(s))\bigr\vert}_{\displaystyle\text{(ii)}}\\
    &\quad+\underbrace{\eta^{-1}\bigl\vert\mathcal{H}(\wh\nu_h(s))-\mathcal{H}(\nu_h^*(s))\bigr\vert}_{\displaystyle\text{(iii)}}.
\end{aligned}\label{verrordecomp}
\end{equation}
Now we successively bound the three terms above. Firstly,
\begin{equation}
\begin{aligned}
\text{(i)}
&\leq\left\vert\wh{\mu}_h(s)^\top(\wh Q_h-Q_h)(s)\wh\nu_h(s)\right\vert+\left\vert(\wh{\mu}_h-\mu_h^*)(s)^\top Q_h(s)\wh\nu_h(s)\right\vert+\left\vert\mu_h^*(s)^\top Q_h(s)(\wh\nu_h-\nu_h^*)(s)\right\vert\\
&\leq\sup_{a\in\mathcal{A},b\in\mathcal{B}}\bigl\vert\wh Q_h-Q_h\bigr\vert(s,a,b)+\Vert\wh{\mu}_h(s)-\mu_h^*(s)\Vert_1\Vert Q_h(s)\wh\nu_h(s)\Vert_\infty\\
&\quad + \Vert\wh{\nu}_h(s)-\nu_h^*(s)\Vert_1\Vert Q_h(s)^\top\wh\mu_h(s)\Vert_\infty\\
&\leq\sup_{a\in\mathcal{A},b\in\mathcal{B}}\bigl\vert\wh Q_h-Q_h\bigr\vert(s,a,b)+2\epsilon\sup_{a\in\mathcal{A},b\in\mathcal{B}}\vert Q_h(s,a,b)\vert +2\epsilon\sup_{a\in\mathcal{A},b\in\mathcal{B}}\vert Q_h(s,a,b)\vert\\
&\leq C_h\sqrt{\kappa_h}+4R\epsilon,
\end{aligned}\label{quadraticbound}
\end{equation}
where the first inequality holds by Hölder's inequality, and the third inequality follows from the following facts that
\begin{equation*}
    \begin{aligned}
    &\bigl\vert\wh Q_h-Q_h\bigr\vert(s,a,b)=\bigl\vert(\wh{\theta}_h-\theta_h)^\top\phi(s,a,b)\bigr\vert\leq\Vert\wh{\theta}_h-\theta_h\Vert\left\Vert\phi(s,a,b)\right\Vert\leq C_h\sqrt{\kappa_h},\\
    &\bigl\vert Q_h(s,a,b)\bigr\vert=\vert\theta_h^\top\phi(s,a,b)\vert\leq\Vert\theta_h\Vert\left\Vert\phi(s,a,b)\right\Vert\leq R,\quad\forall (s,a,b)\in\mathcal{S}\times\mathcal{A}\times\mathcal{B}.
\end{aligned}
\end{equation*}
To bound terms (ii) and (iii), by Lemma \ref{entropybuond}, we have
\begin{equation}
\begin{aligned}
\vert\mathcal{H}(\mu_h^*(s))-\mathcal{H}(\wh\mu(s))\vert&\leq-\epsilon\log\epsilon-(1-\epsilon)\log(1-\epsilon)+\epsilon\log(m-1)\leq\epsilon\left(1+\log\frac{m}{\epsilon}\right);\\
\vert\mathcal{H}(\nu^*_h(s))-\mathcal{H}(\wh\nu(s))\vert&\leq-\epsilon\log\epsilon-(1-\epsilon)\log(1-\epsilon)+\epsilon\log(n-1)\leq\epsilon\left(1+\log\frac{n}{\epsilon}\right).
\end{aligned}\label{vregb}
\end{equation}
Combining (\ref{verrordecomp}), (\ref{quadraticbound}) and (\ref{vregb}), for all $s\in\mathcal{S}$, we have
\begin{equation*}
\begin{aligned}
    \vert\wh V_h(s)-V_h(s)\vert&\leq C_h\sqrt{\kappa_h}+\epsilon\left(4R+2+\log\frac{mn}{\epsilon^2}\right)\\
    &\lesssim\left(\sqrt{S(m+n)}+\log(Tmn)\right)\sqrt{\frac{m\vee n}{T}\log\frac{HS}{\delta}}.
\end{aligned}
\end{equation*}
Furthermore, for all $h\in[H]$ and all $(s,a,b)\in\mathcal{S}\times\mathcal{A}\times\mathcal{B}$, since $\Vert\phi(\cdot,\cdot,\cdot)\Vert\leq 1$, we have
\begin{equation}
\begin{aligned}
\vert\bbP_h\wh V_{h+1}-\bbP_h V_{h+1}\vert(s,a,b)&\leq\Vert\phi(s,a,b)\Vert\left\Vert\Pi_h\right\Vert_\mathrm{op}\Vert\wh V_h-V_h\Vert\\
&\lesssim\frac{S(m+n)+\sqrt{S(m+n)}\log T}{\sqrt{T}}\sqrt{\log\frac{HS}{\delta}}.
\end{aligned}\label{transvbound}
\end{equation}
Note that this bound is valid if the concentration event $$\mathcal{E}_2=\left\{\Theta_h\subseteq\widehat{\Theta}_h\text{ and }D_H(\Theta_h,\widehat{\Theta}_h)\lesssim\sqrt{\kappa_h},\ \forall h\in[H]\right\}$$ holds.

\subsubsection{Bounding the Estimation Error of Transition Kernels}\phantomsection\label{sec:b.2.2}
We bound the difference between $\wh\bbP_h\wh V_{h+1}$ and $\bbP_h\wh V_{h+1}$.
\paragraph{Error decomposition.} We fix $\wh{V}_{h+1}\in\wh{\mathcal{V}}_{h+1}$, and let $\mathscr{H}_h^t$ be the $\sigma$-field generated by variables $(s_1,a_1,b_1,s_2,a_2,b_2,\cdots,s_h,a_h,b_h)$. Define random variables $\eta_h^t=\delta(s_{h+1}^t)-\Pi_h\phi(s_h^t,a_h^t,b_h^t)$. Then $\bbE[\eta_h^t|\mathscr{H}_h^t]=0$. For all $s\in\mathcal{S},a\in\mathcal{A},b\in\mathcal{B}$, 
\begin{equation*}
    \begin{aligned}
&\bigl(\wh\bbP_h\wh V_{h+1}-\bbP_h\wh V_{h+1}\bigr)(s,a,b)=\wh V_{h+1}^\top\bigl(\wh\Pi_h-\Pi_h\bigr)\phi(s,a,b)\\
&\quad=\wh V_{h+1}^\top\left(\sum_{t=1}^T\left(\eta_h^t+\Pi_h\phi(s_h^t,a_h^t,b_h^t)\right)\phi(s_h^t,a_h^t,b_h^t)^\top\Lambda_h^{-1}\phi(s,a,b)-\Pi_h\right)\phi(s,a,b)\\
&\quad=\sum_{t=1}^T\left(\wh V_{h+1}^\top\eta_h^t\right)\phi(s_h^t,a_h^t,b_h^t)^\top\Lambda_h^{-1}\phi(s,a,b)-\lambda\wh V_{h+1}^\top\Pi_h\Lambda_h^{-1}\phi(s,a,b).
\end{aligned}
\end{equation*}
By Cauchy's inequality, we obtain
\begin{equation}
\begin{aligned}
&\bigl\vert\wh\bbP_h\wh V_{h+1}-\bbP_h\wh V_{h+1}\bigr\vert(s,a,b)\\
&\leq\left(\left\Vert\sum_{t=1}^T\left(\wh V_{h+1}^\top\eta_h^t\right)\phi(s_h^t,a_h^t,b_h^t)\right\Vert_{\Lambda_h^{-1}}+\lambda\left\Vert\Pi_h^\top\wh{V}_{h+1}\right\Vert_{\Lambda_h^{-1}}\right)\Vert\phi(s,a,b)\Vert_{\Lambda_h^{-1}},
\end{aligned}\label{transitiondiff}
\end{equation}
where we write $\Vert x\Vert_M=\sqrt{x^\top Mx}$ for any positive definite matrix $M$. Now we bound the three terms in (\ref{transitiondiff}).
\paragraph{Step I: Analysis of the term $\Vert\Pi_h^\top\wh V_{h+1}\Vert_{\Lambda_h^{-1}}$.}
    
Since $$\Lambda_h=\lambda I_d+\sum_{t=1}^T\phi(s_h^t,a_h^t,b_h^t)\phi(s_h^t,a_h^t,b_h^t)^\top,$$ we have $\lambda_{\min}(\Lambda_h)\geq\lambda$. Hence
    \begin{equation*}
        \Vert\Pi_h^\top\wh V_{h+1}\Vert_{\Lambda_h^{-1}}\leq\frac{1}{\sqrt{\lambda}}\Vert\Pi_h^\top\wh V_{h+1}\Vert\leq\frac{1}{\sqrt{\lambda}}\Vert\Pi_h\Vert_\mathrm{op}\Vert\wh V_{h+1}\Vert.
    \end{equation*}
    Note that by Assumption \ref{alin}, we have $\Vert\pi_h(\cdot)\Vert\leq\sqrt{d}$, and thus we have $\Vert\Pi_h\Vert_\mathrm{op}\leq\sqrt{Sd}$. Hence
    \begin{equation}
    \Vert\Pi_h^\top\wh V_{h+1}\Vert_{\Lambda_h^{-1}}\leq\sqrt{\frac{Sd}{\lambda}}(2R+\eta^{-1}\log m+\eta^{-1}\log n).\label{transition1}
    \end{equation}
\paragraph{Step II: Analysis of the self-normalized process.}
By definition of $\eta_h^t$, we have $\Vert\eta_h^t\Vert_\infty\leq 2$. Combining with the bound of $\wh V_{h+1}$ given by (\ref{Vbound}), we have
    \begin{equation*}
        -2(R+\eta^{-1}\log n)\leq \wh V_{h+1}^\top\eta_h^t\leq 2(R+\eta^{-1}\log m).
    \end{equation*}
    By Hoeffding's inequality, $(\wh V_{h+1}^\top\eta_h^t)_{t=1}^T$ are independent $(2R+\eta^{-1}\log m+\eta^{-1}\log n)$-sub-Gaussian random variables with  mean zero. By Lemma \ref{selfnormbound}, we have that, with probability at least $1-\delta$,
    \begin{equation*}
    \Biggl\Vert\sum_{t=1}^T\wh V_{h+1}^\top\eta_h^t\phi(s_h^t,a_h^t,b_h^t)\Biggr\Vert_{\Lambda_h^{-1}}^2\leq 2(2R+\eta^{-1}\log m+\eta^{-1}\log n)^2\log\left(\frac{\det(\Lambda_h)^{1/2}}{\delta\det(\lambda I_d)^{1/2}}\right).
    \end{equation*}
    Now we bound $\det(\Lambda_h)$. To this end, note that
    \begin{equation*}
\det(\Lambda_h)\leq\Vert\Lambda_h\Vert_\mathrm{op}^d\leq\left(\lambda+T\Vert\phi(s_h^t,a_h^t,b_h^t)\Vert^2\right)^d\leq(\lambda+T)^d.
    \end{equation*}
    Consequently, we have
    \begin{equation}
    \left\Vert\sum_{t=1}^T\wh V_{h+1}^\top\eta_h^t\phi(s_h^t,a_h^t,b_h^t)\right\Vert_{\Lambda_h^{-1}}^2\leq 2(2R+\eta^{-1}\log mn)^2\left(\log\frac{1}{\delta}+\frac{d}{2}\log\left(1+\frac{T}{\lambda}\right)\right).\label{selfnormbb1}
    \end{equation}
    Next, we derive a uniform bound for $\wh{\mathcal{V}}_{h+1}$. We choose a $\epsilon$-net of $(\wh{\mathcal{V}}_{h+1},\Vert\cdot\Vert_\infty)$ given in (\ref{coveringbound}). Then for all $\wh V_{h+1}\in\wh{\mathcal{V}}_{h+1}$, we can find $V^*\in\mathcal{N}_\epsilon$ such that $\Vert\wh V_{h+1}-V^*\Vert_\infty\leq\epsilon$. Hence
    \begin{equation*}
        \left\Vert\sum_{t=1}^T(\wh V_{h+1}-V^*)^\top\eta_h^t\phi(s_h^t,a_h^t,b_h^t)\right\Vert_{\Lambda_h^{-1}}\leq \frac{1}{\sqrt{\lambda}}\sum_{t=1}^T\Vert\wh V_{h+1}-V^*\Vert_\infty\Vert\eta_h^t\Vert_1\left\Vert\phi(s_h^t,a_h^t,b_h^t)\right\Vert\leq\frac{2T\epsilon}{\sqrt{\lambda}}.
    \end{equation*}
    We apply a union bound version of (\ref{selfnormbb1}) on $\mathcal{N}_\epsilon$. With probability at least $1-\delta$, the following inequality holds for all $\wh{\mathcal{V}}_{h+1}$ that
    \begin{equation*}
        \begin{aligned}
        &\left\Vert\sum_{t=1}^T\wh V_{h+1}^\top\eta_h^t\phi(s_h^t,a_h^t,b_h^t)\right\Vert_{\Lambda_h^{-1}}\\
        &\quad\leq\left\Vert\sum_{t=1}^T V^{*\top}\eta_h^t\phi(s_h^t,a_h^t,b_h^t)\right\Vert_{\Lambda_h^{-1}}+\left\Vert\sum_{t=1}^T(\wh V_{h+1}-V^*)^\top\eta_h^t\phi(s_h^t,a_h^t,b_h^t)\right\Vert_{\Lambda_h^{-1}}\\
        &\quad\leq\sqrt{2}(2R+\eta^{-1}\log mn)\sqrt{\log\frac{\vert\mathcal{N}_\epsilon\vert}{\delta}+\frac{d}{2}\log\left(1+\frac{T}{\lambda}\right)}+\frac{2T\epsilon}{\sqrt{\lambda}}.
    \end{aligned}
    \end{equation*}
    Letting $\epsilon=1/T$,  we obtain
    \begin{align}
    \left\Vert\sum_{t=1}^T\wh V_{h+1}^\top\eta_h^t\phi(s_h^t,a_h^t,b_h^t)\right\Vert_{\Lambda_h^{-1}}^2\leq(2R+\eta^{-1}\log mn)\sqrt{2\log\frac{(1+2RT)^d}{\delta}+d\log\left(1+\frac{T}{\lambda}\right)}+\frac{2}{\sqrt{\lambda}}.\label{transition2}
    \end{align}
\paragraph{Step III: Analysis of $\Vert\phi(s,a,b)\Vert_{\Lambda_h^{-1}}$.} Let $\rho_h$ be the visit measure of $(s_h,a_h,b_h)$ induced by QRE policies $\mu^*$ and $\nu^*$. For simplicity, we let
\begin{equation*}
    \ol{\Lambda}_h=\frac{1}{T}\Lambda_h=\frac{\lambda}{T}I_d+\frac{1}{T}\sum_{t=1}^T\phi(s_h^t,a_h^t,b_h^t)\phi(s_h^t,a_h^t,b_h^t)^\top,\  \Psi_h=\bbE\left[\phi(s_h,a_h,b_h)\phi(s_h,a_h,b_h)^\top\right].
\end{equation*}
 We first present our main result below.
\begin{lemma}[Adapted from \citealp{min2022varianceawareoffpolicyevaluationlinear} Lemma H.5]\label{transition3}
Let $\{(s_h^t,a_h^t,b_h^t)\}_{t=1}^T$ be i.i.d. samples from the visit distribution $\rho_h$. For any $\delta>0$, if
\begin{equation}
    T\geq\max\left\{512\Vert\Psi_h^{-1}\Vert^2_\mathrm{op}\log\frac{2d}{\delta},4\lambda\Vert\Psi^{-1}_h\Vert_\mathrm{op}\right\},\label{samplecomplexity1}
\end{equation}
then with probability at least $1-\delta$, it holds simultaneously for all $s\in\mathcal{S},a\in\mathcal{A},b\in\mathcal{B}$ that
\begin{equation*}
    \Vert\phi(s,a,b)\Vert_{\Lambda_h^{-1}}\leq\frac{2}{\sqrt{T}}\Vert\phi(s,a,b)\Vert_{\Psi^{-1}_h}.
\end{equation*}
\proof{Proof.}
We first bound the difference between $\ol{\Lambda}_h$ and $\Psi_h$. We write $x_t=\phi(s_h^t,a_h^t,b_h^t)$, and define the matrix-valued function $\Sigma(x_1,\cdots,x_T)=\lambda I_d/T+T^{-1}\sum_{t=1}^Tx_tx_t^\top$. For any $t\in[T]$, if we replace $x_t$ by some $\wt x_t$ with $\Vert\wt x_t\Vert\leq 1$, we have
\begin{equation*}
\begin{aligned}
&\left(\Sigma(x_1,\cdots,x_{t-1},x_t,x_{t+1},\cdots,x_T)-\Sigma(x_1,\cdots,x_{t-1},\wt x_t,x_{t+1},\cdots,x_T)\right)^2\\
&=\frac{1}{T^2}\left(x_tx_t^\top-\wt x_t\wt x_t^\top\right)^2\preceq\frac{1}{T^2}\left(2x_tx_t^\top x_tx_t^\top+2\wt x_t\wt x_t^\top\wt x_t\wt x_t^\top\right)\preceq\frac{4}{T^2}I_d=:A_t^2.
\end{aligned}
\end{equation*}
Let $\sigma^2=\left\Vert\sum_{t=1}^TA_t^2\right\Vert=\frac{4}{T}$. By Lemma \ref{matrixmcdiarmid}, with probability at least $1-\delta$, we have
\begin{equation*}
\left\Vert\ol\Lambda_h-\bbE[\ol\Lambda_h]\right\Vert_\mathrm{op}\leq\frac{4\sqrt{2}}{\sqrt{T}}\sqrt{\log\frac{2d}{\delta}},
\end{equation*}
which implies
\begin{equation}
\left\Vert\ol\Lambda_h-\Psi_h\right\Vert_\mathrm{op}\leq\frac{4\sqrt{2}}{\sqrt{T}}\sqrt{\log\frac{2d}{\delta}}+\frac{\lambda}{T}.\label{diffLambda}
\end{equation}
Next, we bound the term $\Vert v\Vert_{\Lambda_h^{-1}}$ for any $v=\phi(s,a,b)\in\bbR^d$:
\begin{align}
\Vert v\Vert_{\Lambda_h^{-1}}&=\frac{1}{\sqrt{T}}\sqrt{v^\top\Psi_h^{-1}v+v^\top(\ol{\Lambda}_h^{-1}-\Psi_h^{-1})v}\notag\\
&=\frac{1}{\sqrt{T}}\sqrt{v^\top\Psi_h^{-1}v+v^\top\Psi_h^{-1/2}(\Psi_h^{1/2}\ol{\Lambda}_h^{-1}\Psi_h^{1/2}-I_d)\Psi_h^{-1/2}v}\notag\\
&\leq\frac{1}{\sqrt{T}}\sqrt{\Vert v\Vert_{\Psi_h^{-1}}\left(1+\left\Vert\Psi_h^{1/2}\ol{\Lambda}_h^{-1}\Psi_h^{1/2}-I_d\right\Vert_\mathrm{op}\right)\Vert v\Vert_{\Psi_h^{-1}}}\notag\\
&\leq\frac{1}{\sqrt{T}}\left(1+\left\Vert\Psi_h^{1/2}\ol{\Lambda}_h^{-1}\Psi_h^{1/2}-I_d\right\Vert^{1/2}_\mathrm{op}\right)\Vert v\Vert_{\Psi_h^{-1}}.\label{lambdaweightedbound}
\end{align}
To bound the term $\left\Vert\Psi_h^{1/2}\ol{\Lambda}_h^{-1}\Psi_h^{1/2}-I_d\right\Vert_\mathrm{op}$, note that
\begin{equation*}
\left\Vert\Psi_h^{1/2}\ol{\Lambda}_h^{-1}\Psi_h^{1/2}-I_d\right\Vert_\mathrm{op}\leq\left\Vert\left(\Psi_h^{-1/2}\ol{\Lambda}_h\Psi_h^{-1/2}\right)^{-1}\right\Vert_\mathrm{op}\left\Vert I_d-\Psi_h^{-1/2}\ol{\Lambda}_h\Psi_h^{-1/2}\right\Vert_{\mathrm{op}}
\end{equation*}
Combining (\ref{samplecomplexity1}) and the bound (\ref{diffLambda}), we have
\begin{equation*}
    \left\Vert I_d-\Psi_h^{-1/2}\ol{\Lambda}_h\Psi_h^{-1/2}\right\Vert_{\mathrm{op}}=\Vert\Psi_h^{-1}\Vert_\mathrm{op}\Vert\ol{\Lambda}_h-\Psi_h^{-1}\Vert_\mathrm{op}\leq\frac{1}{2}.
\end{equation*}
Moreover, by Weyl's inequality, we have
\begin{equation*}
\lambda_{\min}\left(\Psi_h^{-1/2}\ol{\Lambda}_h\Psi_h^{-1/2}\right)\geq 1-\lambda_{\max}\left(I_d-\Psi_h^{-1/2}\ol{\Lambda}_h\Psi_h^{-1/2}\right)\geq\frac{1}{2}.
\end{equation*}
Consequently, we have
\begin{equation*}
\left\Vert\Psi_h^{1/2}\ol{\Lambda}_h^{-1}\Psi_h^{1/2}-I_d\right\Vert_\mathrm{op}\leq\frac{1}{\lambda_{\min}\left(\Psi_h^{-1/2}\ol{\Lambda}_h\Psi_h^{-1/2}\right)}\left\Vert I_d-\Psi_h^{-1/2}\ol{\Lambda}_h\Psi_h^{-1/2}\right\Vert_{\mathrm{op}}\leq 2\cdot\frac{1}{2}=1.
\end{equation*}
Plugging the above into (\ref{lambdaweightedbound}) completes the proof.
\endproof
\end{lemma}
\paragraph{Final bound.} We  obtain the final bound by combining (\ref{transitiondiff}), (\ref{transition1}), (\ref{transition2}), and Lemma \ref{transition3}. In particular, for any $\delta>0$, if
\begin{equation}
    T\geq\max\left\{512\Vert\Psi_h^{-1}\Vert^2_\mathrm{op}\log\frac{2Hd}{\delta},4\lambda\Vert\Psi^{-1}_h\Vert_\mathrm{op}\right\},\label{sc2}
\end{equation}
then, with probability at least $1-2\delta$, it holds simultaneously for all $h\in[H]$, all $\wh{V}_{h+1}\in\wh{\mathcal{V}}_{h+1}$, and all $s\in\mathcal{S},a\in\mathcal{A},b\in\mathcal{B}$ that
\begin{equation}
\begin{aligned}
&\bigl\vert\wh\bbP_h\wh V_{h+1}-\bbP_h\wh V_{h+1}\bigr\vert(s,a,b)\\
&\leq\left(\frac{2R+\eta^{-1}\log mn}{\sqrt{T}}\left(\sqrt{\frac{Sd}{\lambda}}+\sqrt{2\log\frac{H(1+2RT)^d}{\delta}+d\log\frac{\lambda+T}{\lambda }}\right)+\frac{2}{\sqrt{\lambda T}}\right)\Vert\phi(s,a,b)\Vert_{\Psi_h^{-1}}\\
&\leq\frac{2R+\eta^{-1}\log mn}{\sqrt{T}}\left(\sqrt{\frac{4Sd}{\lambda}}+\sqrt{2\log\frac{H}{\delta}+2d\log(1+2RT)+d\log\frac{\lambda+T}{\lambda}}\right)\Vert\Psi_h^{-1}\Vert_\mathrm{op}^{1/2}\\
&\lesssim\frac{\sqrt{Sd}+\sqrt{d\log T}+\sqrt{\log(H/\delta)}}{\sqrt{T}}\log(mn).
\end{aligned}\label{finaltransitionbound}
\end{equation}
This is our final bound for the error between $\wh{\bbP}_h\wh{V}_{h+1}$ and $\wh{\bbP}_h V_{h+1}$, which converges to zero as the sample size $T\to\infty$. 

\subsubsection{Bounding the Estimation Error of Rewards}
We provide the proof of Theorem \ref{samp}, and bound $D_H(\mathcal{R},\wh{\mathcal{R}})$.
\proof{Proof.}[Proof of Theorem \ref{samp}]
For every $h\in[H]$ and $\wh r\in\wh{\mathcal{R}}$, the estimated reward function $\wh r_h$ can be written as
\[\widehat{r}_h(s,a,b) = \widehat{Q}_h(s,a,b)-\gamma\widehat{\mathbb{P}}_h\widehat{V}_{h+1}(s,a,b)\quad\text{for all }(s,a,b)\in\mathcal{S}\times\mathcal{A}\times\mathcal{B},\]
which is generated by some $\wh\theta_h\in\wh\Theta_h$ and $\wh V_{h+1}\in\wh{\mathcal{V}}_{h+1}$. When the concentration event $\mathcal{E}_2$ given in Corollary \ref{conequ} holds, we have a feasible $\theta_h\in\Theta_h$ such that $\Vert\wh\theta_h-\theta_h\Vert\leq C_h\sqrt{\kappa_h}$ for some constant $C_h>0$. Moreover, by the conclusion in Appendix \ref{sec:b.2.1}, we also have a feasible $V_{h+1}\in\mathcal{V}_{h+1}$ such that (\ref{transvbound}) holds for all possible choices of $h$ and $(s,a,b)$. Now we consider the  feasible reward that
\begin{equation*}
    r_h(s,a,b)=\phi(s,a,b)^\top\theta_h-\gamma\bbP_h V_{h+1}(s,a,b),\quad r=(r_1,r_2,\cdots,r_H)\in\mathcal{R}.
\end{equation*}
First, we have the  decomposition that
\begin{equation*}
\begin{aligned}
    &\vert\wh r_h-r_h\vert(s,a,b)=\left\vert\phi(s,a,b)^\top(\wh\theta_h-\theta_h)-\gamma(\wh\bbP_h\wh V_{h+1}-\bbP_h V_{h+1})(s,a,b)\right\vert\\
    &\quad\leq\underbrace{\left\Vert\phi(s,a,b)\right\Vert \Vert\wh\theta_h-\theta_h\Vert}_{\displaystyle\leq C_h\sqrt{\kappa_h}} + \underbrace{\gamma\bigl\vert\wh\bbP_h\wh V_{h+1}-\bbP_h\wh V_{h+1}\bigr\vert(s,a,b)}_{\displaystyle\text{bounded by (\ref{finaltransitionbound})} }+\underbrace{\gamma\bigl\vert\bbP_h\wh V_{h+1}-\bbP_h V_{h+1}\bigr\vert(s,a,b)}_{\displaystyle\text{bounded by (\ref{transvbound})}}.
\end{aligned}
\end{equation*}
Note that in the above, both the first and third bounds hold under the concentration event $\mathcal{E}_2$, and the second holds with probability $1-2\delta$ when $T$ satisfies (\ref{sc2}). Furthermore, all these bounds are uniform, and do not depend on our choice of $h$, $\wh\theta_h,\wh V_{h+1}$, and $(s,a,b)$. Thus, with probability at least $1-3\delta$, we have
\begin{equation}
    \sup_{\wh r\in\wh{\mathcal{R}}} d(\wh{r},\mathcal{R})\lesssim\sqrt{\frac{S(m+n)}{T}\log\frac{HS}{\delta}}\left(\sqrt{S(m+n)}+\log T\right)+\frac{\sqrt{Sd}+\sqrt{d\log T}}{\sqrt{T}}\log(mn).\label{hausdist1}
\end{equation}

On the other hand, for any $r\in\mathcal{R}$ generated by $\theta_h\in\Theta_h\subseteq\wh\Theta_h$ and $\theta_{h+1}\in\Theta_{h+1}\subseteq\wh\Theta_{h+1}$, we pick the following estimated reward $\wh{r}=(\wh r_1,\wh r_2,\cdots,\wh r_H)\in\wh{\mathcal{R}}$ that
\begin{equation*}
\begin{aligned}
&\wt V_{h+1}(s)=\wh\mu_{h+1}(s)^\top Q_{h+1}(s)\wh\nu_{h+1}(s)+\eta^{-1}\mathcal{H}(\wh\nu_{h+1}(s))-\eta^{-1}\mathcal{H}(\wh\nu_{h+1}(s)),\\
&\wh r_h(s,a,b)=\phi(s,a,b)^\top\theta_h - \wh\bbP_h\wt{V}_{h+1}(s,a,b),\quad h\in[H],s\in\mathcal{S},a\in\mathcal{A},b\in\mathcal{B}.
\end{aligned}
\end{equation*}
Since $\Theta_{h+1}\subseteq\wh\Theta_{h+1}$, the estimated V-function $\wt V_{h+1}\in\wh{\mathcal{V}}_{h+1}$, and the uniform bound (\ref{finaltransitionbound}) still holds if we replacing $\wh V_{h+1}$ by $\wt V_{h+1}$. Furthermore, similar to our estimation procedure (\ref{verrordecomp})-(\ref{transvbound}), the estimation error of $V$-function satisfies that
\begin{equation*}
\vert\wt V_{h+1}(s)-V_{h+1}(s)\vert\leq\left(4R+2+\log\frac{mn}{\epsilon^2}\right)\epsilon\lesssim\log(Tmn)\sqrt{\frac{m\vee n}{T}\log\frac{HS}{\delta}},
\end{equation*}
and
\begin{equation}
\vert\bbP_h\wt{V}_{h+1}-\bbP_hV_{h+1}\vert(s,a,b)\lesssim\frac{\sqrt{S(m+n)}\log(Tmn)}{\sqrt{T}}\sqrt{\log\frac{HS}{\delta}},\quad\forall(s,a,b)\in\mathcal{S}\times\mathcal{A}\times\mathcal{B}.
\label{transvbound2}
\end{equation}
Then, the following inequality also holds for all choices of $h$, $r_h$, and $(s,a,b)$:
\begin{equation*}
    \begin{aligned}
\vert\wh{r}_h-r_h\vert(s,a,b)&=\vert\wh\bbP_h\wt{V}_{h+1}-\bbP_hV_{h+1}\vert(s,a,b)\\
&\leq\underbrace{\vert\wh\bbP_h\wt{V}_{h+1}-\bbP_h\wt V_{h+1}\vert(s,a,b)}_{\displaystyle\text{bounded by (\ref{finaltransitionbound})}}+\underbrace{\vert\bbP_h\wt{V}_{h+1}-\bbP_hV_{h+1}\vert(s,a,b)}_{\displaystyle\text{bounded by (\ref{transvbound2})}}\\
&\lesssim\sqrt{\frac{S(m+n)}{T}\log\frac{HS}{\delta}}\log (Tmn)+\left(\sqrt{\frac{Sd}{T}}+\sqrt{\frac{d\log T}{T}}\right)\log(mn).
\end{aligned}
\end{equation*}
Therefore, we have
\begin{equation}
\sup_{r\in\mathcal{R}}d(r,\wh{\mathcal{R}})\lesssim\sqrt{\frac{S(m+n)}{T}\log\frac{HS}{\delta}}\log (Tmn)+\left(\sqrt{\frac{Sd}{T}}+\sqrt{\frac{d\log T}{T}}\right)\log(mn).\label{hausdist2}
\end{equation}
Combining (\ref{hausdist1}) and (\ref{hausdist2}), we obtain a bound of the Hausdorff distance between $\mathcal{R}$ and $\wh{\mathcal{R}}$ that
\begin{equation*}
    \begin{aligned}
    D_H(\mathcal{R},\wh{\mathcal{R}})&=\max\left\{\sup_{\wh r\in\wh{\mathcal{R}}} d(\wh{r},\mathcal{R}),\ \sup_{r\in\mathcal{R}}d(r,\wh{\mathcal{R}})\right\}\\
    &\lesssim\sqrt{\frac{S(m+n)}{T}\log\frac{HS}{\delta}}\left(\sqrt{S(m+n)}+\log T\right)+\frac{\sqrt{Sd}+\sqrt{d\log T}}{\sqrt{T}}\log(mn),
\end{aligned}
\end{equation*}
which completes the proof.
\endproof
\subsection{Convergence of MLE in policy estimation}
\subsubsection{Linear Parameterized Strategies}
We prove the case for the max player, whose action space is $\cal{A}$. We assume that we have a feature map $\psi:\mathcal{S}\times\mathcal{A}\to\bbR^{d_a}$ such that $\Vert\psi(s,a)\Vert\leq K$ for all $s\in\mathcal{S},a\in\mathcal{A}$. The strategy $\mu$ of the max player is then parameterized by a group of vectors $\vartheta_h\in\bbR^{d_a}$:
\begin{equation*}
\mu_h(a_h|s_h)=\mu_{\vartheta_h}(a_h,s_h)=\frac{\exp\left(\vartheta_h^\top\psi(s_h,a_h)\right)}{\sum_{a\in\cal{A}}\exp\left(\vartheta_h^\top\psi(s_h,a)\right)},\quad s_h\in\mathcal{S},a_n\in\mathcal{A},h=1,2,\cdots,H.
\end{equation*}

We assume that the true parameters $\vartheta^*_h\in\Gamma$, where $\Gamma=B(0,1)$ is the unit closed ball in $\bbR^{d_a}$ centered at $0$. We define a pseudometric $\rho$ on $\Gamma$:
\begin{equation}
\begin{aligned}
\rho(\vartheta,\wt{\vartheta})&=\max_{s\in\cal{S}}H\left(\mu_{\vartheta}(\cdot|s),\mu_{\wt{\vartheta}}(\cdot|s)\right)=\max_{s\in\cal{S}}\sqrt{H^2\left(\mu_{\vartheta}(\cdot|s),\mu_{\wt{\vartheta}}(\cdot|s)\right)}\\
&=\max_{s\in\cal{S}}\sqrt{\frac{1}{2}\sum_{a\in\cal{A}}\left(\sqrt{\mu_{\vartheta}(a|s)}-\sqrt{\mu_{\wt{\vartheta}}(a|s)}\right)^2},
\end{aligned}
\end{equation}
where $H^2(\cdot,\cdot)$ is the squared Hellinger distance between two probability distributions.
\paragraph{Lower bound for actions.}
Since the parameters $\vartheta_h\in\Gamma$, and the feature map $\psi:\mathcal{S}\times\mathcal{B}\to\bbR^{d_a}$ ranges in $B(0,K)$, we have
\begin{equation*}
\vert\vartheta_h^\top\psi(s_h,a_h)\vert\leq\Vert\vartheta_h\Vert\left\Vert\psi(s_h,a_h)\right\Vert\leq K,\quad\forall s_h\in\mathcal{S},a_h\in\mathcal{A}.
\end{equation*}
Consequently,
\begin{equation*}
\mu_h(a_h|s_h)=\frac{\exp\left(\vartheta_h^\top\psi(s_h,a_h)\right)}{\sum_{a\in\cal{A}}\exp\left(\vartheta_h^\top\psi(s_h,a)\right)}\geq\frac{e^{-K}}{\sum_{a\in\cal{A}}e^{K}}=\frac{e^{-2K}}{m}.
\end{equation*}
Therefore, for every $h\in[H]$, $a_h\in\mathcal{A}$, $s_h\in\mathcal{S}$ and $\vartheta_h\in\Gamma$, the following bound holds:
\begin{equation}
    \frac{e^{-2K}}{m}\leq\mu_h(a_h|s_h)\leq 1.\label{strategybound}
\end{equation}
This bound is very useful when we analyze the variation of parameters $\vartheta_h$.
\begin{lemma}
There exists a constant $L_K>0$ such that for any $\vartheta,\wt{\vartheta}\in\Gamma$,
\begin{equation}
    \rho(\vartheta,\wt{\vartheta})\leq L_K\Vert \vartheta-\wt{\vartheta}\Vert.\label{hellingerlipschitz}
\end{equation}
\end{lemma}
\proof{Proof.}
Define the softmax function $p=(p_1,\cdots,p_m):\bbR^{m}\to[0,1]^m$ by
\begin{equation*}
    p(x_1,\cdots,x_m)=\left(\frac{e^{x_1}}{\sum_{i=1}^me^{x_i}},\cdots,\frac{e^{x_m}}{\sum_{i=1}^me^{x_i}}\right).
\end{equation*}
The Jacobian matrix is then 
\begin{equation*}
    J_p=\begin{pmatrix}
        p_1(1-p_1) & -p_1p_2 & \cdots & -p_1p_m\\
        -p_2p_1 & p_2(1-p_2) & \cdots & -p_2p_m\\
        \vdots & \vdots & \ddots & \vdots\\
        -p_mp_1 & -p_mp_2 & \cdots & p_m(1-p_m)
    \end{pmatrix}.
\end{equation*}
Note that $p_1+p_2+\cdots+p_m=1$. By triangle inequality,
\begin{equation*}
    \Vert J_p\Vert_{\mathrm{op}}=\Vert\mathrm{diag}(p)-pp^\top\Vert_{\mathrm{op}}\leq\Vert\mathrm{diag}(p)\Vert_{\mathrm{op}}+\Vert pp^\top\Vert_{\mathrm{op}}\leq 2.
\end{equation*}
Hence $p$ is a Lipschitz function. Now for every $s\in\cal{S}$, define
\begin{equation*}
x^s=\left(\vartheta^\top\psi(s,1),\cdots,\vartheta^\top\psi(s,m)\right)^\top,\quad \wt{x}^s=\left(\wt{\vartheta}^\top\psi(s,1),\cdots,\wt{\vartheta}^\top\psi(s,m)\right)^\top.
\end{equation*}
Then, we have
\begin{equation}
    \Vert x^s-\wt{x}^s\Vert\leq\sqrt{\Vert\vartheta-\wt{\vartheta}\Vert^2\sum_{a\in\mathcal{A}}\Vert\psi(x,a)\Vert^2}\leq \sqrt{m}K\Vert\vartheta-\wt{\vartheta}\Vert.\label{xsb}
\end{equation}
Furthermore, we  bound the Hellinger distance between $\mu_\vartheta$ and $\mu_{\wt{\vartheta}}$ by
\begin{equation}
\begin{aligned}
\sqrt{H^2\left(\mu_{\vartheta}(\cdot|s),\mu_{\wt{\vartheta}}(\cdot|s)\right)}&\leq\sqrt{\frac{1}{2}\sum_{a\in\cal{A}}\frac{me^{2K}}{2}\left\vert\mu_{\vartheta}(a|s)-\mu_{\wt{\vartheta}}(a|s)\right\vert^2}=\sqrt{\frac{m}{4}}e^K\Vert p(x^s)-p(\wt{x}^s)\Vert\\
&\leq \sqrt{m}e^K\Vert x^s-\wt{x}\Vert\overset{\text{(\ref{xsb})}}{\leq}mKe^K\Vert\vartheta-\wt{\vartheta}\Vert,
\end{aligned}\label{hellingerbound}
\end{equation}
where in the first inequality, we use the fact that
\begin{equation*}
    \begin{aligned}
\left\vert\sqrt{\mu_{\vartheta}(a|s)}-\sqrt{\mu_{\wt{\vartheta}}(a|s)}\right\vert&\leq\frac{1}{2\sqrt{\min\{\mu_{\vartheta}(a|s),\mu_{\wt{\vartheta}}(a|s)\}}}\left\vert\mu_{\vartheta}(a|s)-\mu_{\wt{\vartheta}}(a|s)\right\vert\\
&\overset{\text{(\ref{strategybound})}}{\leq}\sqrt{\frac{me^{2K}}{2}}\left\vert\mu_{\vartheta}(a|s)-\mu_{\wt{\vartheta}}(a|s)\right\vert.
\end{aligned}
\end{equation*}
The result (\ref{hellingerlipschitz}) follows by taking the uniform bound (\ref{hellingerbound}) for all $s\in\mathcal{S}$.
\endproof

\subsubsection{Proof of Lemma \ref{mlepolicyconvg}}\phantomsection\label{sec:b.3.2}
\proof{Proof.}
This conclusion is inspired by \citealp{chen2023actionsspeakwantprovably}. To recover the strategy taken by the max player from an offline dataset $\{(s_1^t,a_1^t),(s_2^t,a_2^t),\cdots,(s_H^t,a_H^t)\}_{t=1}^T,$
we employ the maximum likelihood estimator on parameters $\vartheta_h$. Then the estimator $\wh{\vartheta}_h$ is obtained by solving the minimization problem of the following negative log-likelihood (for notation simplicity we drop the superscript $a$ in $\mathcal{L}_h^a$, which indicates the player):
\begin{equation*}
    \mathcal{L}_h(\vartheta_h)=-\sum_{t=1}^T\log\mu_{\vartheta_h}(a_h^t|s_h^t).
\end{equation*}
With $\epsilon>0$ fixed, we take a minimal $\epsilon$-net $\wt{\Gamma}_\epsilon$ of $\Gamma$ under the pseudometric $\rho$, whose cardinality is $\vert\wt{\Gamma}_\epsilon\vert=\mathcal{N}(\epsilon,\Gamma,\rho)$. We then fix $\wt{\vartheta}_h\in\wh{\Gamma}_h$ and apply Lemma \ref{foster}, replacing $\delta$ with $\delta/\mathcal{N}(\epsilon,\Gamma,\rho)$, to obtain that
\begin{align*}
\frac{1}{2}\left(\mathcal{L}_h(\vartheta_h^*)-\mathcal{L}_h(\wt{\vartheta}_h)\right)&\leq\sum_{t=1}^T\log\bbE\left[\sqrt{\frac{\mu_{\wt{\vartheta}_h}(a_h^t|s_h^t)}{\mu_h^*(a_h^t|s_h^t)}}\right]+\log\frac{\mathcal{N}(\epsilon,\Gamma,\rho)}{\delta}\\
&\leq\sum_{t=1}^T\left(\bbE\left[\sqrt{\frac{\mu_{\wt{\vartheta}_h}(a_h^t|s_h^t)}{\mu_h^*(a_h^t|s_h^t)}}\right]-1\right)+\log\frac{\mathcal{N}(\epsilon,\Gamma,\rho)}{\delta}\\
&=T\bbE_{s_h\sim\rho_h}\left[\bbE_{a_h\sim\mu_h^*(\cdot|s_h)}\left[\sqrt{\frac{\mu_{\wt{\vartheta}_h}(a_h|s_h)}{\mu_h^*(a_h|s_h)}}\right]-1\right]+\log\frac{\mathcal{N}(\epsilon,\Gamma,\rho)}{\delta}\\
&=-T\bbE\left[H^2\left(\mu_{\wt{\vartheta}_h}(\cdot|s_h),\mu_h^*(\cdot|s_h)\right)\right]+\log\frac{\mathcal{N}(\epsilon,\Gamma,\rho)}{\delta},
\end{align*}
which holds with probability at least $1-\delta/\mathcal{N}(\epsilon,\Gamma,\rho)$. Taking a union bound, we conclude that, with probability at least $1-\delta$, the above inequality holds for all $\wt{\vartheta}_h\in\wt{\Gamma}_\epsilon$.
By (\ref{strategybound}), for all $\vartheta_h\in\Gamma$ and all $\wt{\vartheta}_h\in\wt{\Gamma}_\epsilon$, we have
\begin{align*}
\frac{1}{2}\left\vert\mathcal{L}_h(\vartheta_h)-\mathcal{L}_h(\wt{\vartheta}_h)\right\vert&=\sum_{t=1}^T\left\vert\log\sqrt{\mu_{\wt{\vartheta}_h}(a_h^t|s_h^t)}-\log\sqrt{\mu_{\vartheta_h}(a_h^t|s_h^t)}\right\vert\\
&\leq\sum_{t=1}^T\sqrt{m}e^K\left\vert\sqrt{\mu_{\wt{\vartheta}_h}(a_h^t|s_h^t)}-\sqrt{\mu_{\vartheta_h}(a_h^t|s_h^t)}\right\vert\\
    &\leq\sqrt{2m}Te^K\rho(\vartheta_h,\wt{\vartheta}_h).
\end{align*}
Therefore, with probability at least $1-\delta$, it holds for all $\vartheta_h\in\Gamma$ and all $\wt{\vartheta}_h\in\wt{\Gamma}_\epsilon$ that
\begin{align}
\bbE\left[H^2\left(\mu_{\wt{\vartheta}_h}(\cdot|s_h),\mu_h^*(\cdot|s_h)\right)\right]\leq\frac{\mathcal{L}_h(\vartheta_h)-\mathcal{L}_h(\vartheta_h^*)}{2T}+\frac{1}{T}\log\frac{\mathcal{N}(\epsilon,\Gamma,\rho)}{\delta}+\sqrt{2m}e^K\rho(\vartheta_h,\wt{\vartheta}_h).\label{hellingernetbound}
\end{align}
On the other hand, for all $s_h\in\mathcal{S}$, we have
\begin{equation}
\begin{aligned}
&\left\vert H^2\left(\mu_{\wt{\vartheta_h}}(\cdot|s_h),\mu_h^*(\cdot|s_h)\right)-H^2\left(\mu_{\vartheta_h}(\cdot|s_h),\mu_h^*(\cdot|s_h)\right)\right\vert\\
&\quad=\left(H\left(\mu_{\wt{\vartheta_h}}(\cdot|s_h),\mu_h^*(\cdot|s_h)\right)+H\left(\mu_{\vartheta_h}(\cdot|s_h),\mu_h^*(\cdot|s_h)\right)\right)\\
&\quad\quad\times\left\vert H\left(\mu_{\wt{\vartheta_h}}(\cdot|s_h),\mu_h^*(\cdot|s_h)\right)-H\left(\mu_{\vartheta_h}(\cdot|s_h),\mu_h^*(\cdot|s_h)\right)\right\vert\\
&\quad\leq 2H\left(\mu_{\wt{\vartheta_h}}(\cdot|s_h),\mu_{\vartheta_h}(\cdot|s_h)\right)\leq 2\rho(\wt{\vartheta}_h,\vartheta_h).
\end{aligned}\label{hellingerapprox}
\end{equation}
Combining (\ref{hellingernetbound}) and (\ref{hellingerapprox}), and take $\wt{\vartheta}_h\in\wt{\Gamma}_\epsilon$ for each $\vartheta_h\in\Gamma$ such that $\rho(\vartheta_h,\wt{\vartheta}_h)\leq\epsilon$, we have
\begin{align}
\bbE\left[H^2\left(\mu_{\vartheta_h}(\cdot|s_h),\mu_h^*(\cdot|s_h)\right)\right]\leq\frac{\mathcal{L}_h(\vartheta_h)-\mathcal{L}_h(\vartheta_h^*)}{2T}+\frac{1}{T}\log\frac{\mathcal{N}(\epsilon,\Gamma,\rho)}{\delta}+(\sqrt{2m}e^K+2)\epsilon,\label{hellingerepsbound}
\end{align}
which holds with probability at least $1-\delta$ for all $\vartheta_h\in\Gamma$.

\paragraph{Bounding the Covering Number.} Now we  bound the covering number $\mathcal{N}(\epsilon,\Gamma,\rho)$. By Lemma~\ref{hellingerlipschitz}, we have the following inclusion of balls under different metrics that
\begin{equation*}
    B_{\Vert\cdot\Vert}\left(\vartheta,\frac{\epsilon}{L_K}\right)\subseteq B_\rho(\vartheta,\epsilon).
\end{equation*}
Combining this result with Lemma \ref{coveringbound}, we have
\begin{equation}
    \mathcal{N}\left(\epsilon,\Gamma,\rho\right)\leq \mathcal{N}\left(\frac{\epsilon}{L_K},\Gamma,\Vert\cdot\Vert\right)\leq\left(1+\frac{2L_K}{\epsilon}\right)^{d_a}.\label{gammacoverbound}
\end{equation}
\paragraph{Final bound.} We combine (\ref{hellingerepsbound}) and (\ref{gammacoverbound}), and take $\epsilon=1/T$. Then with probability at least $1-\delta$, it holds for all $\vartheta_h\in\Gamma$ that
\begin{equation*}
\bbE\bigl[H^2\bigl(\mu_{\vartheta_h}(s_h),\mu_h^*(s_h)\bigr)\bigr]\leq\frac{\mathcal{L}_h(\vartheta_h)-\mathcal{L}_h(\vartheta_h^*)}{2T}+\frac{d_a\log(1+2TL_K)+\log\delta^{-1}+\sqrt{2m}e^K+2}{T}.
\end{equation*}
We take the maximum likelihood estimator:
\begin{equation*}
\wh{\vartheta}_h=\underset{\vartheta_h\in\Gamma}{\mathrm{argmin}}\,\mathcal{L}_j(\vartheta_h).
\end{equation*}
Since $\mathcal{L}_h(\wh{\vartheta}_h)\leq\mathcal{L}_h(\vartheta_h^*)$, we have
\begin{equation*}
\bbE\left[H^2\left(\wh{\mu}_h(s_h),\mu_h^*(s_h)\right)\right]\leq\frac{d_a\log(1+2TL_K)+\log\delta^{-1}+\sqrt{2m}e^K+2}{T},
\end{equation*}
where $\wh{\mu}_h$ is the strategy associated with the estimator $\wh{\vartheta}_h$. Note that
\begin{equation*}
\mathrm{TV}(P,Q)\leq\sqrt{2}H(P,Q)
\end{equation*}
holds for all distributions $P$ and $Q$, we have
\begin{equation}
\bbE\left[\mathrm{TV}^2\left(\wh{\mu}_h(s_h),\mu_h^*(s_h)\right)\right]\leq\frac{2d_a\log(1+2TL_K)+2\log\delta^{-1}+2\sqrt{2m}e^K+4}{T},\label{tvsqfinalbound}
\end{equation}
which concludes the proof.
\endproof
\subsection{Proof of Theorem \ref{mlehausconvg}}\phantomsection\label{sec:b.4}
\proof{Proof.}
The proof follows three steps:
\paragraph{Bounding the estimation error of V-functions.} Similar to Section~\ref{sec:b.2.1}, we bound the error between $\wh{V}_h$ and $V_h$ in expectation that
\begin{equation}\label{expverrordecomp}
\begin{aligned}
    &\bbE_{s\sim\rho_h}\left[\vert\wh V_h(s)-V_h(s)\vert\right]\leq\underbrace{\bbE_{s\sim\rho_h}\left[\left\vert\wh{\mu}_h(s)^\top\wh Q_h(s)\wh\nu_h(s)-\mu_h^*(s)^\top Q_h(s)\nu_h^*(s)\right\vert\right]}_{\displaystyle\text{(i)}}\\
    &\quad+\underbrace{\bbE_{s\sim\rho_h}\left[\frac{1}{\eta}\bigl\vert\mathcal{H}(\wh\mu_h(s))-\mathcal{H}(\mu_h^*(s))\bigr\vert\right]}_{\displaystyle\text{(ii)}}+\underbrace{\bbE_{s\sim\rho_h}\left[\frac{1}{\eta}\bigl\vert\mathcal{H}(\wh\nu_h(s))-\mathcal{H}(\nu_h^*(s))\bigr\vert\right]}_{\displaystyle\text{(iii)}}.
\end{aligned}
\end{equation}
We bound the three terms above. Note that,
\begin{equation*}
\begin{aligned}
&\left\vert\wh{\mu}_h(s)^\top\wh Q_h(s)\wh\nu_h(s)-\mu_h^*(s)^\top Q_h(s)\nu_h^*(s)\right\vert\\
&\leq\left\vert\wh{\mu}_h(s)^\top(\wh Q_h-Q_h)(s)\wh\nu_h(s)\right\vert+\left\vert(\wh{\mu}_h-\mu_h^*)(s)^\top Q_h(s)\wh\nu_h(s)\right\vert+\left\vert\mu_h(s)^\top Q_h(s)(\wh\nu_h-\nu_h^*)(s)\right\vert\\
&\leq\sup_{a\in\mathcal{A},b\in\mathcal{B}}\bigl\vert\wh Q_h-Q_h\bigr\vert(s,a,b)+\Vert\wh{\mu}_h(s)-\mu_h^*(s)\Vert_1\Vert Q_h(s)\wh\nu_h(s)\Vert_\infty \\
&\quad+ \Vert\wh{\nu}_h(s)-\nu_h^*(s)\Vert_1\Vert Q_h(s)^\top\wh\mu_h(s)\Vert_\infty\\
&\leq\sup_{a,b}\left\vert\wh Q_h-Q_h\right\vert(s,a,b)+2\left(\mathrm{TV}(\mu_h^*(s_h),\widehat\mu_h(s_h))+\mathrm{TV}(\nu_h^*(s_h),\widehat\nu_h(s_h))\right)\sup_{a,b}\vert Q_h(s,a,b)\vert\\
&\leq C_h\sqrt{\kappa_h}+2R\cdot\mathrm{TV}(\mu_h^*(s_h),\widehat\mu_h(s_h))+2R\cdot\mathrm{TV}(\nu_h^*(s_h),\widehat\nu_h(s_h)),
\end{aligned}
\end{equation*}
where the second and third inequalities follow from Hölder's inequality, and the fourth from boundedness of parameter $\theta_h$. By taking expectation on both sides of the last display, we have
\begin{equation}
    \text{(i)}\leq C_h\sqrt{\kappa_h}+4R\epsilon.\label{expquadraticbound}
\end{equation}
Again, we use Lemma \ref{entropybuond} to bound term (ii). Letting $\tau=\mathrm{TV}(\mu_h^*(s_h),\widehat\mu_h(s_h))$, we have
\begin{equation*}
\vert\mathcal{H}(\mu_h^*(s))-\mathcal{H}(\wh\mu_h(s))\vert\leq-\tau\log\tau-(1-\tau)\log(1-\tau)+\tau\log(m-1)\leq\tau\left(1+\log\frac{m}{\tau}\right).
\end{equation*}
Similarly, by taking the expectation on both sides and using Jensen's inequality,
\begin{equation*}
    \text{(ii)}\leq\epsilon\left(1+\log\frac{m}{\epsilon}\right).
\end{equation*}
A similar bound holds for term (iii) by replacing $m$ by $n$. Combining (\ref{expverrordecomp}), (\ref{expquadraticbound}) and the entropy bounds, we get
\begin{equation*}
    \bbE_{s\sim\rho_h}\left[\vert\wh V_h(s)-V_h(s)\vert\right]\leq C_h\sqrt{\kappa_h}+\epsilon\left(4R+2+\log\frac{mn}{\epsilon^2}\right).
\end{equation*}
Furthermore, for all $h\in[H]$ and all $(s,a,b)\in\mathcal{S}\times\mathcal{A}\times\mathcal{B}$, since $\Vert\phi(\cdot,\cdot,\cdot)\Vert\leq 1$, we have
\begin{align}
&\bbE_{s\sim\rho_h}\left[\vert\bbP_h\wh V_{h+1}-\bbP_h V_{h+1}\vert(s,a,b)\right]\leq\Vert\phi(s,a,b)\Vert\left\Vert\Pi_h\right\Vert_\mathrm{op}\bbE_{s\sim\rho_h}\left[\Vert\wh V_h-V_h\Vert\right]\notag\\
&\quad\leq\left\Vert\Pi_h\right\Vert_\mathrm{op}\left(C_h\sqrt{\kappa_h}+\epsilon\left(4R+2+\log\frac{mn}{\epsilon^2}\right)\right)\label{exptransvbound}\\
&\quad\lesssim\frac{1}{\sqrt{T}}\left(m^{7/4}+n^{7/4}+(m^{3/2}+n^{3/2}+\log T)\sqrt{(d_a+d_b)\log T+\log(H/\delta)}+\sqrt{S}\log(mn)\right).\notag
\end{align}

\paragraph{Bounding the transition error.} 
In Section~\ref{sec:b.2.2}, we bound the error between $\wh\bbP_h\wh{V}_{h+1}$ and $\bbP_h\wh V_{h+1}$ over all $V_{h+1}\in\widehat{\mathcal{V}}_{h+1}$. This bound does not depend on the total variation distance between the true and the estimated policies. Therefore,  bound (\ref{finaltransitionbound}) still holds in this case for all $s\in\mathcal{S},a\in\mathcal{A}$, and $b\in\mathcal{B}$. Therefore, under some certain concentration event $\mathcal{E}_t$ with probability at least $1-\delta$,
\begin{equation}
\bbE_{s\sim\rho_h}\left[\bigl\vert\wh\bbP_h\wh V_{h+1}-\bbP_h\wh V_{h+1}\bigr\vert(s,a,b)\right]
\lesssim\frac{\sqrt{Sd}+\sqrt{d\log T}+\sqrt{\log(H/\delta)}}{\sqrt{T}}\log(mn).
\label{exptransitionbound}
\end{equation}

\paragraph{Bounding the Hausdorff distance between sets.} According to Lemma \ref{qestmle}, for any $\wh\theta_h\in\wh\Theta_h$, we can find a feasible $\theta_h\in\Theta_h$ such that $\Vert\wh\theta_h-\theta_h\Vert\leq C_h\sqrt{\kappa_h}$ for some constant $C_h$. Moreover, for any $\wh V_{h+1}\in\wh{\mathcal{V}}_{h+1}$, we can also find a feasible V-function $V_{h+1}\in\mathcal{V}_{h+1}$ such that (\ref{exptransvbound}) holds for all possible choice of $(a,b)$. Consider the reward functions $\wh{r}_h$ and $r_h$ generated by $(\wh\theta_h,\wh{V}_{h+1})$ and $(\theta_h,V_{h+1})$, respectively. We have
\begin{equation*}
    \begin{aligned}
    &\bbE_{s\sim\rho_h}\left[\vert\wh r_h-r_h\vert(s,a,b)\right]=\bbE_{s\sim\rho_h}\left[\left\vert\phi(s,a,b)^\top(\wh\theta_h-\theta_h)-\gamma(\wh\bbP_h\wh V_{h+1}-\bbP_h V_{h+1})(s,a,b)\right\vert\right]\notag\\
    &\leq\bbE_{s\sim\rho_h}\bigl[\underbrace{\left\Vert\phi(s,a,b)\right\Vert \Vert\wh\theta_h-\theta_h\Vert}_{\displaystyle\leq R\sqrt{\kappa_h}} + \gamma\underbrace{\bigl\vert\wh\bbP_h\wh V_{h+1}-\bbP_h\wh V_{h+1}\bigr\vert(s,a,b)}_{\displaystyle\text{bounded\ by\ (\ref{exptransvbound})}}+\underbrace{\gamma\bigl\vert\bbP_h\wh V_{h+1}-\bbP_h V_{h+1}\bigr\vert(s,a,b)}_{\displaystyle\text{bounded\ by\ (\ref{exptransitionbound})}}\bigr]\notag\\
    &\lesssim\frac{1}{\sqrt{T}}\left(m^{7/4}+n^{7/4}+(m^{3/2}+n^{3/2}+\log T)\sqrt{(d_a+d_b+d)\log T+\log(H/\delta)}+\sqrt{Sd}\log(mn)\right).
\end{aligned}
\end{equation*}
Since this bound is valid for any choice of $(\wh{\theta}_h)$, we have
\begin{equation}\label{exphausdist1}
\begin{aligned}
    \sup_{\wh r\in\wh{\mathcal{R}}}D_1(\wh{r},\mathcal{R})
    &\lesssim\frac{1}{\sqrt{T}}\biggl(m^{7/4}+n^{7/4}\\
    &\quad +(m^{3/2}+n^{3/2}+\log T)\sqrt{(d_a+d_b+d)\log T+\log(H/\delta)}+\sqrt{Sd}\log(mn)\biggr).
\end{aligned}
\end{equation}

On the other hand, for any $r\in\mathcal{R}$ generated by $\theta_h\in\Theta_h\subseteq\wh\Theta_h$ and $\theta_{h+1}\in\Theta_{h+1}\subseteq\wh\Theta_{h+1}$, we pick the following estimated reward $\wh{r}=(\wh r_1,\wh r_2,\cdots,\wh r_H)\in\wh{\mathcal{R}}$ that
\begin{equation*}
\begin{aligned}
&\wt V_{h+1}(s)=\wh\mu_{h+1}(s)^\top Q_{h+1}(s)\wh\nu_{h+1}(s)+\eta^{-1}\mathcal{H}(\wh\nu_{h+1}(s))-\eta^{-1}\mathcal{H}(\wh\nu_{h+1}(s)),\\
&\wh r_h(s,a,b)=\phi(s,a,b)^\top\theta_h - \wh\bbP_h\wt{V}_{h+1}(s,a,b),\quad h\in[H],s\in\mathcal{S},a\in\mathcal{A},b\in\mathcal{B}.
\end{aligned}
\end{equation*}
Since $\Theta_{h+1}\subseteq\wh\Theta_{h+1}$, the estimated V-function $\wt V_{h+1}\in\wh{\mathcal{V}}_{h+1}$, and the bound (\ref{exptransitionbound}) still holds if we replace $\wh V_{h+1}$ by $\wt V_{h+1}$. Furthermore, similar to our estimation procedure (\ref{expverrordecomp})-(\ref{exptransvbound}), the estimation error of $V$-function satisfies the following bound that
\begin{equation}
\begin{aligned}
    &\bbE_{s\sim\rho_h}\left[\vert\bbP_h\wt{V}_{h+1}-\bbP_hV_{h+1}\vert(s,a,b)\right]\leq \left\Vert\Pi_h\right\Vert_\mathrm{op}\left(C_h\sqrt{\kappa_h}+\epsilon\left(4R+2+\log\frac{mn}{\epsilon^2}\right)\right),\\
    &\lesssim\frac{1}{\sqrt{T}}\left(m^{7/4}+n^{7/4}+(m^{3/2}+n^{3/2}+\log T)\sqrt{(d_a+d_b)\log T+\log(H/\delta)}+\sqrt{S}\log(mn)\right).
\end{aligned}\label{exptransvbound2}
\end{equation}
Consequently, the following inequality simultaneously holds for all choice of $h$, $r_h$ and $(a,b)$:
\begin{equation*}
    \begin{aligned}
&\bbE_{s\sim\rho_h}\left[\vert\wh{r}_h-r_h\vert(s,a,b)\right]=\bbE_{s\sim\rho_h}\left[\vert\wh\bbP_h\wt{V}_{h+1}-\bbP_hV_{h+1}\vert(s,a,b)\right]\\
&\leq\underbrace{\bbE_{s\sim\rho_h}\left[\vert\wh\bbP_h\wt{V}_{h+1}-\bbP_h\wt V_{h+1}\vert(s,a,b)\right]}_{\displaystyle\text{bounded by (\ref{exptransitionbound})}}+\underbrace{\bbE_{s\sim\rho_h}\left[\vert\bbP_h\wt{V}_{h+1}-\bbP_hV_{h+1}\vert(s,a,b)\right]}_{\displaystyle\text{bounded by (\ref{exptransvbound2})}}\\
&\lesssim\frac{1}{\sqrt{T}}\left(m^{7/4}+n^{7/4}+(m^{3/2}+n^{3/2}+\log T)\sqrt{(d_a+d_b+d)\log T+\log(H/\delta)}+\sqrt{Sd}\log(mn)\right).
\end{aligned}
\end{equation*}
Therefore we have
\begin{equation*}
\begin{aligned}
\sup_{r\in\mathcal{R}}D_1(r,\wh{\mathcal{R}})
&\lesssim\frac{1}{\sqrt{T}}\biggl(m^{7/4}+n^{7/4}\notag\\
&\quad+(m^{3/2}+n^{3/2}+\log T)\sqrt{(d_a+d_b+d)\log T+\log(H/\delta)}+\sqrt{Sd}\log(mn)\biggr).\label{exphausdist2}
\end{aligned}
\end{equation*}
Combining (\ref{exphausdist1}) and (\ref{exphausdist2}), we obtain the Hausdorff distance between $\mathcal{R}$ and $\wh{\mathcal{R}}$:
\begin{equation*}
\begin{aligned}
    &D_1(\mathcal{R},\wh{\mathcal{R}})=\max\left\{\sup_{\wh r\in\wh{\mathcal{R}}} D_1(\wh{r},\mathcal{R}),\ \sup_{r\in\mathcal{R}}D_1(r,\wh{\mathcal{R}})\right\}\\
    &\lesssim\frac{1}{\sqrt{T}}\left(m^{7/4}+n^{7/4}+(m^{3/2}+n^{3/2}+\log T)\sqrt{(d_a+d_b+d)\log T+\log(H/\delta)}+\sqrt{Sd}\log(mn)\right),
\end{aligned}
\end{equation*}
which concludes the proof.
\endproof

\section{Proof of Auxiliary Lemmas}
\subsection{Proof of Lemma \ref{mxc}}\phantomsection\label{pmxc}
\proof{Proof.}
We first prove that $\Theta\subseteq\widehat{\Theta}$. We adopt the following notations for ease of presentation that 
    \[X=\begin{bmatrix}
    A(\nu^*)\\
    B(\mu^*)
\end{bmatrix}, y= \begin{bmatrix}
    c(\mu^*)\\
    d(\nu^*)
\end{bmatrix}\quad\text{and}\quad\widehat{X}=\begin{bmatrix}
    A(\widehat{\nu})\\
    B(\widehat{\mu})
\end{bmatrix},\widehat{y}= \begin{bmatrix}
    c(\widehat{\mu})\\
    d(\widehat{\nu})
\end{bmatrix}.\]
    For any $\theta\in\Theta$, we have $X\theta = y$ by the definition of $\Theta$. Plugging this in, we have
    \begin{equation}
    \lVert \widehat{X}\theta-\widehat{y}\rVert^2 = \lVert (\widehat{X}-X)\theta-(\widehat{y}-y)\rVert^2\leq 2(\lVert \widehat{X}-X\rVert^2_{\text{op}}\lVert\theta\rVert^2+\lVert\widehat{y}-y\rVert^2).\label{xhatthetayhatdiff}
    \end{equation}
    Next, we bound the operator norm $\lVert \widehat{X}-X\rVert_{\text{op}}^2$ that
    \begin{equation*}
        \begin{aligned}
            \lVert \widehat{X}-X\rVert^2_{\text{op}} &= \lVert (\widehat{X}-X)^\top(\widehat{X}-X)\rVert_{\text{op}}\\
            &=\lVert (A(\widehat{\nu})-A(\nu^*))^\top(A(\widehat{\nu})-A(\nu^*))+(B(\widehat{\mu})-B(\mu^*))^\top(B(\widehat{\mu})-B(\mu^*))\rVert_{\text{op}}\\
            &\leq \lVert (A(\widehat{\nu})-A(\nu^*))^\top(A(\widehat{\nu})-A(\nu^*))\rVert_{\text{op}}+\lVert(B(\widehat{\mu})-B(\mu^*))^\top(B(\widehat{\mu})-B(\mu^*))\rVert_{\text{op}}.
        \end{aligned}
    \end{equation*}
    Using \eqref{e3.6}, we have
    \begin{equation}
    \lVert \widehat{X}-X\rVert^2_{\text{op}}\leq\lVert A(\widehat{\nu})-A(\nu^*)\rVert_{\text{op}}^2+\lVert B(\widehat{\mu})-B(\mu^*)\rVert_{\text{op}}^2\leq \lVert \Phi_1\rVert_{\text{op}}^2\cdot\epsilon_2^2+\lVert \Phi_2\rVert_{\text{op}}^2\cdot\epsilon_1^2,\label{Xopnormdiff}
    \end{equation}
    and
    \begin{equation}
    \begin{aligned}
        \lVert\widehat{y}-y\rVert^2&=\lVert c(\widehat{\nu})-c(\nu^*)\rVert^2+\lVert d(\widehat{\mu})-d(\mu^*)\rVert^2\\
        &\leq\frac{m\epsilon_1^2}{\eta^2(\min_{i\in[m]}\mu_i-\epsilon_1)^2}+\frac{n\epsilon_2^2}{\eta^2(\min_{j\in[n]}\nu_j-\epsilon_2)^2}.
    \end{aligned}\label{ydiff}
    \end{equation}
    Plugging  (\ref{Xopnormdiff}) and (\ref{ydiff}) into (\ref{xhatthetayhatdiff}), we obtain that
    \begin{equation}
        \lVert\widehat{X}\theta-\widehat{y}\rVert^2\leq\kappa,
    \end{equation}
    where
    \begin{equation*}
        \kappa=2\left(M\lVert \Phi_1\rVert_{\text{op}}^2+\frac{n}{\eta^2(\min_{j\in[n]}\nu_j-\epsilon_2)^2}\right)\epsilon_2^2+2\left(M\lVert \Phi_2\rVert_{\text{op}}^2+\frac{m}{\eta^2(\min_{i\in[m]}\mu_i-\epsilon_1)^2}\right)\epsilon_1^2.
    \end{equation*}
    Consequently, we have $\Theta\subseteq\widehat{\Theta}$, and $\inf_{\wh\theta\in \widehat{\Theta}}\lVert\wh\theta-\theta\rVert^2 = 0$ for any $\theta\in\Theta$. Therefore, the Hausdorff distance $d_H(\Theta,\widehat{\Theta})$ is
    \begin{equation*}
    d_H(\Theta,\widehat{\Theta}) = \sup_{\wh\theta\in\widehat{\Theta}}\inf_{\theta\in \Theta}\lVert\wh\theta-\theta\rVert.
    \end{equation*}
    Intuitively, $\inf_{\theta\in\Theta}\lVert\wh\theta-\theta\rVert^2$ is the distance between $\wh\theta$ and its projection onto the candidate set $\Theta$. For any $\wh\theta\in\widehat{\Theta}$, the projection of $\wh{\theta}$ onto the affine subspace $S_{X,y}=\{\theta\in\bbR^d:X\theta = y\}$ is 
    \begin{equation*}
    \wt\theta = X^\dagger y+(I-X^\dagger X)\wh\theta.
    \end{equation*}
    Since $\Theta=\{\theta:X\theta = y,\Vert\theta\Vert^2\leq M\}$ is the ball of radius $\sqrt{M-\Vert X^\dagger y\Vert^2}$ in $S_{X,y}$ centered at $X^\dagger y$, we decompose the distance from $\wh\theta\in\widehat{\Theta}$ to its projection $\theta^*$ onto $\Theta$ that
    \begin{align}
        \Vert\wh\theta-\theta^*\Vert^2&=\Vert\wh\theta-\wt\theta\Vert^2 + \Vert\wt\theta-\theta^*\Vert^2\\
        &=\underbrace{\Vert X^\dagger(X\wh\theta - y)\Vert^2}_{\displaystyle\text{(i)}} + \underbrace{\left[\Vert (I-X^\dagger X)\wh\theta\Vert-\sqrt{M-\Vert X^\dagger y\Vert^2}\right]_+^2}_{\displaystyle\text{(ii)}},\label{confhausest}
    \end{align}
    where $[x]_+=\max\{x,0\}$. By the triangle inequality,
    \begin{equation}
    \begin{aligned}
        \Vert X^\dagger(X\wh\theta - y)\Vert&\leq \lVert X^\dagger\rVert_{\text{op}}\Vert X\wh{\theta}-y\Vert\\
        &\leq\lVert X^\dagger\rVert_{\text{op}}\left(\Vert \wh{X}\wh{\theta}-\wh{y}\Vert+\Vert X-\wh{X}\Vert_\mathrm{op}\Vert\wh{\theta}\Vert + \Vert\wh{y}-y\Vert\right)\\
        &\leq\Vert X^\dagger\Vert_\mathrm{op}\Vert \wh{X}\wh{\theta}-\wh{y}\Vert+\Vert X^\dagger\Vert_\mathrm{op}\sqrt{2\Vert X-\wh{X}\Vert_\mathrm{op}^2\Vert\wh{\theta}\Vert^2 + 2\Vert\wh{y}-y\Vert^2}.
    \end{aligned}\label{boundtermi}
    \end{equation}
    Since $\wh{\theta}\in\wh{\Theta}$, we have $\Vert\wh{X}\wh{\theta}-\wh{y}\Vert\leq\sqrt{\kappa}$ and $\Vert\wh{\theta}\Vert^2\leq M$. Plugging  the estimators (\ref{Xopnormdiff}) and (\ref{ydiff}) into (\ref{boundtermi}), we obtain
    \begin{equation}
    \Vert X^\dagger(X\wh\theta-y)\Vert\leq\Vert X^\dagger\Vert_\mathrm{op}\sqrt{\kappa}+\Vert X^\dagger\Vert_\mathrm{op}\sqrt{2M\Vert X-\wh{X}\Vert_\mathrm{op}^2 + 2\Vert\wh{y}-y\Vert^2}\leq 2\sqrt{\kappa}\cdot\Vert X^\dagger\Vert_{\mathrm{op}}.\label{xthetahatydiff}
    \end{equation}
    Hence
    \begin{equation}
    \text{(i)}=\Vert X^\dagger(X\wh\theta-y)\Vert^2\leq 4\kappa\cdot\Vert X^\dagger\Vert_{\mathrm{op}}.\label{termiest}
    \end{equation}
    Applying the triangle inequality to (\ref{xthetahatydiff}), we also have
    \begin{equation*}
    \Vert X^\dagger X\wh\theta\Vert\geq \Vert X^\dagger y\Vert-2\sqrt{\kappa}\cdot\Vert X^\dagger\Vert_{\mathrm{op}}
    \end{equation*}
    Using the orthogonal decomposition $\wh\theta=X^\dagger X\wh\theta+(I-X^\dagger X)\wh\theta$, we have
    \begin{equation*}
    \Vert (I-X^\dagger X)\wh\theta\Vert^2=\Vert\wh{\theta}\Vert^2-\Vert X^\dagger X\wh\theta\Vert^2\leq M-\left(\Vert X^\dagger y\Vert-2\sqrt{\kappa}\cdot\Vert X^\dagger\Vert_{\mathrm{op}}\right)^2.
    \end{equation*}
    Therefore,
    \begin{align}
        \text{(ii)}&\leq\left[\sqrt{M-\left(\Vert X^\dagger y\Vert-2\sqrt{\kappa}\cdot\Vert X^\dagger\Vert_{\mathrm{op}}\right)^2}-\sqrt{M-\Vert X^\dagger y\Vert^2}\right]^2\notag\\
        &\leq\left[\sqrt{M-\Vert X^\dagger y\Vert^2+4\sqrt{\kappa}\cdot\Vert X^\dagger y\Vert\Vert X^\dagger\Vert_{\mathrm{op}}}-\sqrt{M-\Vert X^\dagger y\Vert^2}\right]^2\notag\\
        &\leq\frac{8\kappa\cdot\Vert X^\dagger y\Vert^2\Vert X^\dagger\Vert_{\mathrm{op}}^2}{\sqrt{M-\Vert X^\dagger y\Vert}},\label{termiiest}
    \end{align}
    where we use the inequality $\sqrt{a+b}-\sqrt{a}\leq\frac{b}{2\sqrt{a}}$ for any $a,b>0$ in the last inequality. Combining (\ref{confhausest}), (\ref{termiest}) and (\ref{termiiest}), we have
    \begin{align*}
    \Vert\wh{\theta}-\theta^*\Vert^2\leq 4\kappa\cdot\left(\Vert X^\dagger\Vert_{\mathrm{op}}+\frac{2\Vert X^\dagger y\Vert^2\Vert X^\dagger\Vert_{\mathrm{op}}^2}{\sqrt{M-\Vert X^\dagger y\Vert}}\right).
    \end{align*}
    Therefore, we obtain an upper bound of the Hausdorff distance that
    \begin{equation*}
        d_H(\Theta,\widehat{\Theta}) =\sup_{\wh\theta\in\widehat{\Theta}}\inf_{\theta\in\Theta}\lVert\wh\theta-\theta\rVert\leq\mathcal{O}(\sqrt{\kappa}),
    \end{equation*}
    where the notation $\mathcal{O}(\cdot)$ absorbs a constant depending only on $\Phi_1,\Phi_2,X,y$ and $M$. Thus we conclude the whole proof. 
\endproof

\subsection{Proof of Lemma \ref{qest}}\phantomsection\label{pqest}
\proof{Proof.}
Akin to the proof of Lemma \ref{mxc}, we adopt the following notations that
    \[X_h=\begin{bmatrix}
    A_h(\nu_h^*)\\
    B_h(\mu_h^*)
\end{bmatrix}, y_h= \begin{bmatrix}
    c_h(\mu_h^*)\\
    d_h(\nu_h^*)
\end{bmatrix}\quad\text{and}\quad\widehat{X}_h=\begin{bmatrix}
    A_h(\widehat{\nu}_h)\\
    B_h(\widehat{\mu}_h)
\end{bmatrix},\widehat{y}_h= \begin{bmatrix}
    c_h(\widehat{\mu}_h)\\
    d_h(\widehat{\nu}_h)
\end{bmatrix}.\]
Then we have
\begin{align*}
    \Vert\wh{X}_h-X_h\Vert_{\mathrm{op}}^2 &= \Vert(\wh{X}_h-X_h)^\top(\wh{X}_h-X_h)\Vert_{\mathrm{op}}\\
    &=\biggl\Vert\sum_{s\in\mathcal{S}}(A_h(s,\wh{\nu}_h)-A_h(s,\nu_h^*))^\top(A_h(s,\wh{\nu}_h)-A_h(s,\nu_h^*))\\
    &\qquad+\sum_{s\in\mathcal{S}}(B_h(s,\wh{\mu}_h)-B_h(s,\mu_h^*))^\top(B_h(s,\wh{\mu}_h^*)-B_h(s,\mu_h^*))\biggr\Vert_{\mathrm{op}}\\
    &\leq\sum_{s\in\mathcal{S}}\left\Vert(A_h(s,\wh{\nu}_h)-A_h(s,\nu_h^*))^\top(A_h(s,\wh{\nu}_h)-A_h(s,\nu_h^*))\right\Vert_{\mathrm{op}}\\
    &\qquad+\sum_{s\in\mathcal{S}}\left\Vert(B_h(s,\wh{\mu}_h)-B_h(s,\mu_h^*))^\top(B_h(s,\wh{\mu}_h)-B_h(s,\mu_h^*))\right\Vert_{\mathrm{op}}\\
    &\leq \sum_{s\in\mathcal{S}}\left\Vert A_h(s,\wh{\nu}_h)-A_h(s,\nu_h^*)\right\Vert_{\mathrm{op}}^2+\sum_{s\in\mathcal{S}}\left\Vert B_h(s,\wh{\mu}_h)-B_h(s,\mu_h^*)\right\Vert_{\mathrm{op}}^2\\
    &\leq \sum_{s\in\mathcal{S}}\Vert\Phi_{1,s}\Vert_{\mathrm{op}}^2\epsilon_1^2 + \sum_{s\in\mathcal{S}}\Vert\Phi_{2,s}\Vert_{\mathrm{op}}^2\epsilon_2^2\\
    &=\Vert\Phi_1\Vert_{\mathrm{op}}^2\epsilon_1^2 + \Vert\Phi_2\Vert_{\mathrm{op}}^2\epsilon_2^2,
\end{align*}
where $\Phi_{1,s}\in\mathbb{R}^{d\times(m-1)n}$ and $\Phi_{2,s}\in\mathbb{R}^{d\times(n-1)m}$ are defined as
\begin{equation*}
\Phi_{1,s} = \left(\phi(s,a,\cdot)-\phi(s,1,\cdot)\right)_{a\in[m]\backslash\{1\}},\quad
\Phi_{2,s} = \left(\phi(s,\cdot,b)-\phi(s,\cdot,1)\right)_{b\in[n]\backslash\{1\}},
\end{equation*}
and we let
\begin{equation*}
\Phi_1=\begin{bmatrix}
    \Phi_{1,1} &\Phi_{1,2} &\cdots & \Phi_{1,S}
\end{bmatrix},\quad \Phi_2=\begin{bmatrix}
    \Phi_{2,1} &\Phi_{2,2} &\cdots & \Phi_{2,S}\end{bmatrix}.
\end{equation*}
Furthermore,
\begin{align*}
    \Vert\wh{y}_h-y_h\Vert^2 &= \sum_{s\in\mathcal{S}}\Vert c_h(s,\wh{\nu}_h)-c_h(s,\nu_h^*)\Vert^2 + \sum_{s\in\mathcal{S}}\Vert d_h(s,\wh{\mu}_h)-d_h(s,\mu_h^*)\Vert^2\\
    &\leq\frac{Sm\epsilon_1^2}{\eta^2\left(\min_{s\in\mathcal{S},a\in[m]}\mu_h^*(a|s)-\epsilon_1\right)^2}+\frac{Sn\epsilon_2^2}{\eta^2\left(\min_{s\in\mathcal{S},b\in[n]}\nu_h^*(b|s)-\epsilon_2\right)^2}.
\end{align*}
Consequently, for any $\theta_h\in\Theta_h$, we have
\begin{equation*}
    \Vert\wh{X}_h\theta_h-\widehat{y}_h\Vert^2=\Vert(\wh{X}_h-X_h)\theta_h-(\widehat{y}_h-y_h)\Vert^2\leq 2\left(\Vert \wh{X}_h-X_h\Vert_{\mathrm{op}}^2\Vert\theta_h\Vert^2+\Vert\wh{y}_h-y_h\Vert^2\right)\leq\kappa_h,
\end{equation*}
where
\begin{equation*}
    \begin{aligned}
\kappa_h&=2\left(R^2\Vert\Phi_1\Vert_{\mathrm{op}}^2+\frac{Sm}{\eta^2\left(\min_{s\in\mathcal{S},a\in[m]}\mu_h^*(a|s)-\epsilon_1\right)^2}\right)\epsilon_1^2\\
&\quad+2\left(R^2\Vert\Phi_2\Vert_{\mathrm{op}}^2+\frac{Sn}{\eta^2\left(\min_{s\in\mathcal{S},b\in[n]}\nu_h^*(b|s)-\epsilon_2\right)^2}\right)\epsilon_2^2.
\end{aligned}
\end{equation*}
Therefore $\Theta_h\subseteq\wh{\Theta}_h$, and the Hausdorff distance is
\begin{equation*}
d_H(\Theta_h,\wh{\Theta}_h)=\sup_{\wh{\theta}_h\in\wh{\Theta}_h}\inf_{\theta_h\in\Theta_h}\Vert\wh{\theta}_h-\theta_h\Vert.
\end{equation*}
Analogous to the proof of Lemma \ref{mxc}, for each $h\in[H]$, there exists a constant $C_h$ depending on $\Phi_1,\Phi_2,X_h,y_h$ and $d$ such that
\begin{equation*}
    d_H(\Theta_h,\wh{\Theta}_h)\leq C_h\sqrt{\kappa_h}.
\end{equation*}
Thus we conclude the proof.
\endproof

\subsection{Proof of Lemma \ref{conequ}}\phantomsection\label{pconequ}
\proof{Proof.}
Similar to (\ref{mcdiarmidbound}), given any $h\in[H]$ and $s\in\mathcal{S}$, we have the following McDiarmid's bound for all $\epsilon>0$ that
\begin{equation*}
    \bbP\left(\mathrm{TV}(\wh\mu_h(\cdot|s),\mu_h^*(\cdot|s))\leq\frac{1}{2}\sqrt{\frac{m}{N_h(s)}}+\epsilon\,\Big|\,N_h(s)\right)\geq 1-e^{-2N_h(s)\epsilon^2},
\end{equation*}
and
\begin{equation*}
    \bbP\left(\mathrm{TV}(\wh\nu_h(\cdot|s),\nu_h^*(\cdot|s))\leq\frac{1}{2}\sqrt{\frac{n}{N_h(s)}}+\epsilon\,\Big|\,N_h(s)\right)\geq 1-e^{-2N_h(s)\epsilon^2}.
\end{equation*}
Let $\lambda>0$, and $\mathcal{H}_\lambda$ be the event that $N_h(s)\geq\lambda$ for all $h\in[H]$ and $s\in\mathcal{S}$. Then under event $\mathcal{H}_\lambda$, one can derive the union bound that
\begin{equation}
    \begin{aligned}
\bbP&\left(\mathrm{TV}(\wh\mu_h(s),\mu_h^*(s))\leq\sqrt{\frac{m}{4\lambda}}+\epsilon,\ \mathrm{TV}(\wh\nu_h(s),\nu_h^*(s))\leq\sqrt{\frac{n}{4\lambda}}+\epsilon,\ \forall s\in\mathcal{S},\forall h\in[H]\,\Big|\,\mathcal{H}_\lambda\right)\\
&\geq 1-2HSe^{-2\lambda\epsilon^2}.\label{tvbound1}
\end{aligned}
\end{equation}
We let $\lambda(\epsilon)=\frac{1}{2\epsilon^2}\log\frac{4HS}{\delta}\geq\frac{1}{2\epsilon^2}$. Since $m,n\geq 2$, (\ref{tvbound1}) becomes
\begin{equation}
\begin{aligned}
    &\bbP\left(\mathrm{TV}(\wh\mu_h(s),\mu_h^*(s))\leq\sqrt{2m}\epsilon,\ \mathrm{TV}(\wh\nu_h(s),\nu_h^*(s))\leq\sqrt{2n}\epsilon,\ \forall s\in\mathcal{S},\forall h\in[H]\,\Big|\,\mathcal{H}_{\lambda(\epsilon)}\right)\\
    &\quad\geq 1-\frac{\delta}{2}.
\end{aligned}\label{tvtailmarkov}
\end{equation}
For sufficiently large sample size $T>0$, the Hoeffding's inequality implies
\begin{equation*}
\begin{aligned}
    \bbP\left(N_h(s)\leq\lambda\right)&=\bbP\left(\frac{1}{T}N_h(s)-d_h^{\mu^*,\nu^*}(s)\leq\frac{\lambda}{T}-d_h^{\mu^*,\nu^*}(s)\right)\\
    &\leq\bbP\left(\frac{1}{T}\left(N_h(s)-\bbE[N_h(s)]\right)\leq\frac{\lambda}{T}-C\right)\leq\exp\left(-\frac{2}{T}\left(CT-\lambda\right)^2\right).
\end{aligned}
\end{equation*}
Meanwhile, one can derive the union bound that
\begin{equation*}
    \bbP(\mathcal{H}_\lambda) \geq 1-HS\exp\left(-\frac{2}{T}\left(CT-\lambda\right)^2\right)\geq 1-HS\exp\left(-2C^2T+C\lambda\right).
\end{equation*}
We replace $\epsilon$ by $\frac{\epsilon}{\sqrt{2(m\vee n)}}$ in (\ref{tvtailmarkov}), and fix $\lambda=\frac{m\vee n}{\epsilon^2}\log\frac{4HS}{\delta}$. We set
\begin{equation}
  \begin{aligned}
T=\frac{1}{2C^2}\log\frac{2HS}{\delta}+\frac{\lambda}{2C}=\frac{1}{2C^2}\log\frac{2HS}{\delta}+\frac{m\vee n}{2C\epsilon^2}\log\frac{4HS}{\delta}.
\end{aligned}  \label{tvsamplesize}
\end{equation}
Consequently, we have $\bbP(\mathcal{H}_\lambda)\geq 1-\delta/2$, and
\begin{equation*}
\bbP\left(\mathcal{E}_1\right)
=\bbP\left(\mathcal{E}_1\,|\,\mathcal{H}_\lambda\right)\bbP(\mathcal{H}_\lambda)\geq 1-\delta.
\end{equation*}
Therefore (\ref{tvsamplesize}) concludes the proof.
\endproof

\subsection{Proof of Lemma \ref{qestmle}}\phantomsection\label{sec:c.4}
\paragraph{Notations.} We adopt the following notations for simplicity, 
\begin{equation*}
\begin{aligned}
    A_h(\wh\nu_h)&=\begin{bmatrix}
    \sqrt{\rho_h(1)}A_h(1,\wh{\nu}_h)\\
    \vdots\\
\sqrt{\rho_h(S)}A_h(S,\wh{\nu}_h)
\end{bmatrix},\quad
B_h(\wh\nu_h)=\begin{bmatrix}
    \sqrt{\rho_h(1)}B_h(1,\wh{\mu}_h)\\
    \vdots\\
\sqrt{\rho_h(S)}B_h(S,\wh{\mu}_h)
\end{bmatrix},\\
c_h(\wh\mu_h)&=\begin{bmatrix}
    \sqrt{\rho_h(1)}\,c_h(1,\wh{\nu}_h)\\
    \vdots\\
\sqrt{\rho_h(S)}\,c_h(S,\wh{\nu}_h)
\end{bmatrix},\quad\ 
d_h(\wh\nu_h)=\begin{bmatrix}
    \sqrt{\rho_h(1)}\,d_h(1,\wh{\mu}_h)\\
    \vdots\\
\sqrt{\rho_h(S)}\,d_h(S,\wh{\mu}_h)
\end{bmatrix},
\end{aligned}
\end{equation*}
and
\begin{equation*}
    \begin{aligned}
X_h&=\begin{bmatrix}
    A_h(\nu_h^*)\\
    B_h(\mu_h^*)
\end{bmatrix},\quad 
\Tilde{X}_h=\begin{bmatrix}
    A_h(\widehat{\nu}_h)\\
    B_h(\widehat{\mu}_h)
\end{bmatrix},\quad
\widehat{X}_h=\begin{bmatrix}
    \widehat{A}_h(\widehat{\nu}_h)\\
    \widehat{B}_h(\widehat{\mu}_h)
\end{bmatrix},\\ 
y_h&= \begin{bmatrix}
    c_h(\mu_h^*)\\
    d_h(\nu_h^*)
\end{bmatrix},\quad\ \ 
\Tilde{y}_h= \begin{bmatrix}
     c_h(\widehat{\mu}_h)\\
     d_h(\widehat{\nu}_h)
\end{bmatrix},\quad\ \ 
\widehat{y}_h= \begin{bmatrix}
    \widehat c_h(\widehat{\mu}_h)\\
    \widehat d_h(\widehat{\nu}_h)
\end{bmatrix},
\end{aligned}
\end{equation*}
where $\rho_h$ is the visit measure of the agents at step $h$ using the QRE policies $\mu^*$ and $\nu^*$, and $\mathcal{D}_h$ is the empirical distribution at step $h$ associated with the dataset. The confidence set
is
\begin{equation*}
\widehat{\Theta}_h = \left\{\theta:\left\lVert \wh{X}_h\theta -\wh{y}_h\right\rVert^2\leq \kappa_h,\lVert\theta\rVert\leq R\right\}.
\end{equation*}
We focus on finding an appropriate threshold $\kappa_h>0$.
\proof{Proof.}
The formal proof has three main steps.
\paragraph{Bound the X-matrix error.} We bound the error of the matrix $\wh{X}_h$ in the least square problem. First note that
\begin{equation*}
    \begin{aligned}
    &\Vert\wt{X}_h-X_h\Vert_{\mathrm{op}}^2=\left\Vert(\wt{X}_h-X_h)^\top(\wt{X}_h-X_h)\right\Vert_{\mathrm{op}}\\
    &\leq\left\Vert(A_h(\wh{\nu}_h)-A_h(\nu_h))^\top(A_h(\wh{\nu}_h)-A_h(\nu_h))\right\Vert_{\mathrm{op}}\\
    &\quad+\left\Vert(B_h(\wh{\mu}_h)-B_h(\mu_h))^\top(B_h(\wh{\mu}_h)-B_h(\mu_h))\right\Vert_{\mathrm{op}}\\
    &=\sum_{s\in\mathcal{S}}\rho_h(s_h)\left\Vert A_h(s_h,\wh{\nu}_h)-A_h(s_h,\nu_h)\right\Vert_{\mathrm{op}}^2+\sum_{s\in\mathcal{S}}\rho_h(s_h)\left\Vert B_h(s_h,\wh{\mu}_h)-B_h(s_h,\mu_h)\right\Vert_{\mathrm{op}}^2\\
    &=\bbE_{s_h\sim\rho_h}\left[\left\Vert A_h(s_h,\wh{\nu}_h)-A_h(s_h,\nu_h)\right\Vert_{\mathrm{op}}^2\right]+\bbE_{s_h\sim\rho_h}\left[\left\Vert B_h(s_h,\wh{\mu}_h)-B_h(s_h,\mu_h)\right\Vert_{\mathrm{op}}^2\right].
\end{aligned}
\end{equation*}
Note that
\begin{equation*}
    \begin{aligned}
\bbE_{s_h\sim\rho_h}\left[\left\Vert A_h(s_h,\wh{\nu}_h)-A_h(s_h,\nu_h)\right\Vert_{\mathrm{op}}^2\right]
&=\bbE_{s_h\sim\rho_h}\left[\left\Vert\Phi_{1,s_h}\left(I_{m-1}\otimes\widehat\nu_h(s_h)-I_{m-1}\otimes\nu_h(s_h)\right)\right\Vert_{\mathrm{op}}^2\right]\\
&\leq\max_{s\in\mathcal{S}}\Vert\Phi_{1,s}\Vert_{\mathrm{op}}^2\,\bbE_{s_h\sim\rho_h}\left[\left\Vert \nu_h(\cdot|s_h)-\widehat\nu_h(\cdot|s_h)\right\Vert^2\right]\\
&\leq 4\max_{s\in\mathcal{S}}\Vert\Phi_{1,s}\Vert_{\mathrm{op}}^2\,\bbE_{s_h\sim\rho_h}\left[\mathrm{TV}^2(\nu_h(\cdot|s_h),\widehat\nu_h(\cdot|s_h))\right],
\end{aligned}
\end{equation*}
and similarly,
\begin{equation*}
\bbE_{s_h\sim\rho_h}\left[\left\Vert B_h(s_h,\wh{\mu}_h)-B_h(s_h,\mu_h)\right\Vert_{\mathrm{op}}\right]\leq 4\max_{s\in\mathcal{S}}\Vert\Phi_{2,s}\Vert_{\mathrm{op}}^2\,\bbE_{s_h\sim\rho_h}\left[\mathrm{TV}^2(\mu_h(\cdot|s_h),\widehat\mu_h(\cdot|s_h))\right].
\end{equation*}
Since $\bbE_{s_h\sim\rho_h}\left[\mathrm{TV}^2 (\nu_h(\cdot|s_h),\widehat\nu_h(\cdot|s_h))\right]\leq\epsilon
^2$ and $\bbE_{s_h\sim\rho_h}\left[\mathrm{TV}^2(\mu_h(\cdot|s_h),\widehat\mu_h(\cdot|s_h))\right]\leq\epsilon^2
$, we have
\begin{equation}
\Vert X_h-\wt{X}_h\Vert_{\mathrm{op}}\leq 2\epsilon\sqrt{\max_{s\in\mathcal{S}}\Vert\Phi_{1,s}\Vert_{\mathrm{op}}^2+\max_{s\in\mathcal{S}}\Vert\Phi_{2,s}\Vert_{\mathrm{op}}^2}.\label{XXtildeerror}
\end{equation}
Now we analyze the error from approximating the visit measure by the frequency estimator. Note that
\begin{equation*}
    \begin{aligned}
&\Vert\wh{X}_h-\wt{X}_h\Vert_{\mathrm{op}}^2=\left\Vert(\wh{X}_h-\wt X_h)^\top(\wh{X}_h-\wt X_h)\right\Vert_{\mathrm{op}}\\
    &\leq\left\Vert(\wh A_h(\wh{\nu}_h)-A_h(\wh\nu_h))^\top(\wh A_h(\wh{\nu}_h)-A_h(\wh\nu_h))\right\Vert_{\mathrm{op}}\\
    &\quad+\left\Vert(\wh B_h(\wh{\mu}_h)-B_h(\wh\mu_h))^\top(\wh B_h(\wh{\mu}_h)-B_h(\wh\mu_h))\right\Vert_{\mathrm{op}}\\
    &=\sum_{s_h\in\mathcal{S}}\left\vert\sqrt{\wh{\rho}_h(s_h)}-\sqrt{\rho_h(s_h)}\right\vert^2\left\Vert A_h(s_h,\wh{\nu}_h)\right\Vert_{\mathrm{op}}^2+\sum_{s_h\in\mathcal{S}}\left\vert\sqrt{\wh{\rho}_h(s_h)}-\sqrt{\rho_h(s_h)}\right\vert^2\left\Vert B_h(s_h,\wh{\mu}_h)\right\Vert_{\mathrm{op}}^2\\
    &\leq\sum_{s_h\in\mathcal{S}}\left\vert\wh{\rho}_h(s_h)-\rho_h(s_h)\right\vert\left\Vert A_h(s_h,\wh{\nu}_h)\right\Vert_{\mathrm{op}}^2+\sum_{s_h\in\mathcal{S}}\left\vert\wh{\rho}_h(s_h)-\rho_h(s_h)\right\vert\left\Vert B_h(s_h,\wh{\mu}_h)\right\Vert_{\mathrm{op}}^2,
\end{aligned}
\end{equation*}
where $\wh\rho_h(s)$ is the frequency of the visit to a state $s\in\mathcal{S}$ at the step $h$. Then
\begin{equation*}
\begin{aligned}
\sum_{s_h\in\mathcal{S}}\left\vert\wh{\rho}_h(s_h)-\rho_h(s_h)\right\vert\left\Vert A_h(s_h,\wh{\nu}_h)\right\Vert_{\mathrm{op}}^2
&=\sum_{s\in\cal{S}}\left\vert\wh\rho_h(s)-\rho_h(s)\right\vert\Vert\Phi_{1,s}\Vert_{\mathrm{op}}^2\Vert\nu_h(\cdot|s)\Vert^2\\
&\leq 2\max_{s\in\cal{S}}\Vert \Phi_{1,s}\Vert_{\mathrm{op}}^2\mathrm{TV}(\wh\rho_h,\rho_h).
\end{aligned}
\end{equation*}
Applying the same procedure to $\sum_{s_h\in\mathcal{S}}\left\vert\wh{\rho}_h(s_h)-\rho_h(s_h)\right\vert\left\Vert B_h(s_h,\wh{\mu}_h)\right\Vert_{\mathrm{op}}^2$, we have
\begin{equation*}
\Vert\wh{X}_h-\wt{X}_h\Vert_{\mathrm{op}}\leq 2\sqrt{\max_{s\in\cal{S}}\Vert\Phi_{1,s}\Vert_{\mathrm{op}}^2+\max_{s\in\cal{S}}\Vert\Phi_{2,s}\Vert_{\mathrm{op}}^2}\mathrm{TV}(\wh\rho_h,\rho_h).
\end{equation*}
Using the triangle inequality,
\begin{equation}
\begin{aligned}
    \Vert\wh{X}_h-X_h\Vert_{\mathrm{op}}&\leq\Vert\wh{X}_h-\wt{X}_h\Vert_{\mathrm{op}}+\Vert\wt{X}_h-X_h\Vert_{\mathrm{op}}\\
    &\leq 2\sqrt{\max_{s\in\cal{S}}\Vert\Phi_{1,s}\Vert_{\mathrm{op}}^2+\max_{s\in\cal{S}}\Vert\Phi_{2,s}\Vert_{\mathrm{op}}^2}\left(2\epsilon+\mathrm{TV}(\wh\rho_h,\rho_h)\right).
\end{aligned}\label{Xexperror}
\end{equation}
\paragraph{Bound the y-vector error.} We bound the error of vector $\wh{y}_h$ in the least square problem. By definition and Jensen's inequality,
\begin{equation*}
\begin{aligned}
\Vert\wt{y}_h-y_h\Vert^2&=\sum_{s_h\in\mathcal{S}}\rho_h(s_h)\left\Vert c_h(s_h,\wh\mu_h)-c_h(s_h,\mu_h)\right\Vert^2+\sum_{s_h\in\mathcal{S}}\rho_h(s_h)\left\Vert d_h(s_h,\wh\nu_h)-d_h(s_h,\nu_h)\right\Vert^2\\
&=\bbE_{s_h\sim\rho_h}\left[\left\Vert c_h(s_h,\wh\mu_h)-c_h(s_h,\mu_h)\right\Vert^2\right]+\bbE_{s_h\sim\rho_h}\left[\left\Vert d_h(s_h,\wh\nu_h)-d_h(s_h,\nu_h)\right\Vert^2\right].
\end{aligned}
\end{equation*}
By definition of $c_h(s_h,\mu_h)$ and $c_h(s_h,\wh\mu_h)$,
\begin{equation*}
\begin{aligned}
&\bbE_{s_h\sim\rho_h}\left[\left\Vert c_h(s_h,\wh\mu_h)-c_h(s_h,\mu_h)\right\Vert^2\right]=\frac{1}{\eta^2}\bbE_{s_h\sim\rho_h}\left[\sum_{a=2}^m\left(\log\frac{\widehat\mu_h(a|s_h)}{\widehat\mu_h(1|s_h)}-\log\frac{\mu_h(a|s_h)}{\mu_h(1|s_h)}\right)^2\right]\\
&\quad\leq\frac{2}{\eta^2}\bbE_{s_h\sim\rho_h}\left[(m-1)\left(\log\frac{\widehat\mu_h(1|s_h)}{\mu_h(1|s_h)}\right)^2+\sum_{a=2}^m\left(\log\frac{\widehat\mu_h(a|s_h)}{\mu_h(a|s_h)}\right)^2\right].
\end{aligned}
\end{equation*}
By estimator (\ref{logdiffest}) and bound (\ref{strategybound}), for all $a\in\mathcal{A}$ and $s_h\in\mathcal{S}$, we have
\begin{equation*}
    \left(\log\frac{\widehat\mu_h(a|s_h)}{\mu_h(a|s_h)}\right)^2\leq\left(\frac{\widehat{\mu}_h(a|s_h)-\mu_h(a|s_h)}{e^{-2K}/m}\right)^2\leq m^2e^{4K}\mathrm{TV}^2(\wh\mu_h(\cdot|s_h),\mu_h(\cdot|s_h)).
\end{equation*}
Hence
\begin{equation*}
\bbE_{s_h\sim\rho_h}\left[\left\Vert c_h(s_h,\wh\mu_h)-c_h(s_h,\mu_h)\right\Vert^2\right]
\leq\frac{4m^3e^{4K}}{\eta^2}\bbE_{s_h\sim\rho_h}\left[\mathrm{TV}^2(\wh\mu_h(\cdot|s_h),\mu_h(\cdot|s_h))\right].
\end{equation*}
Applying the same procedure to the other term in the decomposition of $\Vert\wt y_h-y_h\Vert_{\mathrm{op}}^2$, and using the fact that $\bbE_{s_h\sim\rho_h}\left[\mathrm{TV}^2 (\nu_h(\cdot|s_h),\widehat\nu_h(\cdot|s_h))\right]\leq\epsilon^2
$ and $\bbE_{s_h\sim\rho_h}\left[\mathrm{TV}^2 (\mu_h(\cdot|s_h),\widehat\mu_h(\cdot|s_h))\right]\leq\epsilon^2
$, we have
\begin{equation}
    \Vert\wt y_h-y_h\Vert^2\leq\frac{4e^{4K}(m^3+n^3)}{\eta^2}\epsilon
    ^2.\label{ytildeyerror}
\end{equation}
Next, we bound the empirical error $\Vert\wh{y}_h-\wt{y}_h\Vert$ that
\begin{equation*}
\begin{aligned}
\Vert\wh{y}_h-\wt{y}_h\Vert^2&=\sum_{s\in\mathcal{S}}\left\vert\sqrt{\wh{\rho}_h(s)}-\sqrt{\rho_h(s)}\right\vert^2\left\Vert c_h(s,\wh\mu_h)\right\Vert^2+\sum_{s\in\mathcal{S}}\left\vert\sqrt{\wh{\rho}_h(s)}-\sqrt{\rho_h(s)}\right\vert^2\left\Vert d_h(s,\wh\nu_h)\right\Vert^2\\
&\leq\sum_{s\in\mathcal{S}}\left\vert\wh{\rho}_h(s)-\rho_h(s)\right\vert\left\Vert c_h(s,\wh\mu_h)\right\Vert+\sum_{s\in\mathcal{S}}\left\vert\wh{\rho}_h(s)-\rho_h(s)\right\vert\left\Vert d_h(s,\wh\nu_h)\right\Vert^2.
\end{aligned}
\end{equation*}
By the definition of $c_h(s,\wh{\mu_h})$, we have
\begin{equation*}
\sum_{s\in\mathcal{S}}\left\vert\wh\rho_h(s)-\rho_h(s)\right\vert\Vert c_h(s,\wh\mu_h)\Vert
\leq\sum_{s\in\mathcal{S}}\left\vert\wh\rho_h(s)-\rho_h(s)\right\vert\frac{2K\log m}{\eta}\leq\frac{4K\log m}{\eta}\,\mathrm{TV}(\wh\rho_h,\rho_h).
\end{equation*}
Similarly, it holds that
\begin{equation*}
\sum_{s_h\in\mathcal{S}}\left\vert\wh{\rho}_h(s_h)-\rho_h(s_h)\right\vert\left\Vert d_h(s_h,\wh\nu_h)\right\Vert^2\leq\frac{4K\log n}{\eta}\,\mathrm{TV}(\wh\rho_h,\rho_h).
\end{equation*}
Hence, we have
\begin{equation}
\Vert\wh{y}_h-\wt{y}_h\Vert\leq\frac{4K\sqrt{(\log m)^2+(\log n)^2}}{\eta}\mathrm{TV}(\wh\rho_h,\rho_h).\label{hatytildeyerror}
\end{equation}
Combining (\ref{ytildeyerror}) and (\ref{hatytildeyerror}) by the triangle inequality, we have
\begin{equation}
\Vert\widehat y_h-y_h\Vert\leq\frac{2e^{2K}\sqrt{m^3+n^3}}{\eta}\epsilon+\frac{4K\sqrt{(\log m)^2+(\log n)^2}}{\eta}\mathrm{TV}(\wh\rho_h,\rho_h).\label{yexperror}
\end{equation}
\paragraph{Determine the threshold.} By (\ref{Xexperror}) and (\ref{yexperror}), we have
\begin{equation}
\begin{aligned}
\Vert\wh{X}_h\theta_h-\widehat{y}_h\Vert^2&=\Vert(\wh{X}_h-X_h)\theta_h-(\widehat{y}_h-y_h)\Vert^2\\
    &\leq 2\left(\Vert \wh{X}_h-X_h\Vert_{\mathrm{op}}^2\Vert\theta_h\Vert^2+\Vert\wh{y}_h-y_h\Vert^2\right)\\
    &\lesssim R^2\left(\max_{s\in\cal{S}}\Vert\Phi_{1,s}\Vert_{\mathrm{op}}^2+\max_{s\in\cal{S}}\Vert\Phi_{2,s}\Vert_{\mathrm{op}}^2+\frac{e^{4K}(m^3+n^3)}{\eta^2}\right)\epsilon^2\\
    &\quad +\left(\max_{s\in\cal{S}}\Vert\Phi_{1,s}\Vert_{\mathrm{op}}^2+\max_{s\in\cal{S}}\Vert\Phi_{2,s}\Vert_{\mathrm{op}}^2+\frac{4K(\log m n)^2}{\eta^2}\right)\mathrm{TV}^2(\wh{\rho}_h,\rho_h).
\end{aligned}\label{thresholdchoice}
\end{equation}
We consider the following three concentration events:
\begin{itemize}
\item[(i)] \textit{The estimated policy of the max player has small error.} By  bound (\ref{tvsqfinalbound}), with probability at least $1-\delta$, the following event holds that
\begin{equation*}
    \mathcal{E}_a=\left\{\bbE_{s_h\sim\rho_h}\left[\mathrm{TV}^2(\wh{\mu}_h(\cdot|s_h),\mu_h^*(\cdot|s_h))\right]\lesssim\frac{d_a\log T+\log(H/\delta)+\sqrt{m}}{T},\quad\forall h\in[H]\right\}.
\end{equation*}
\item[(ii)] \textit{The estimated policy of the min player has small error.} Similar to the bound (\ref{tvsqfinalbound}), with probability at least $1-\delta$, the following event holds:
\begin{equation*}
    \mathcal{E}_b=\left\{\bbE_{s_h\sim\rho_h}\left[\mathrm{TV}^2(\wh{\nu}_h(\cdot|s_h),\nu_h^*(\cdot|s_h))\right]\lesssim\frac{d_b\log T+\log(H/\delta)+\sqrt{n}}{T},\quad\forall h\in[H]\right\}.
\end{equation*}
\item[(iii)] \textit{The frequency estimator for the visit measure has small error.} Similar to the tail bound (\ref{mcdiarmidbound}), with probability at least $1-\delta$, it holds that
\begin{equation*}
    \mathcal{E}_s=\left\{\mathrm{TV}(\wh{\rho}_h,\rho_h)\leq\frac{1}{2}\sqrt{\frac{S}{T}}+\sqrt{\frac{\log(H/\delta)}{2T}},\quad\forall h\in[H]\right\}.
\end{equation*}
\end{itemize}
The event $\mathcal{E}_a\cap \mathcal{E}_b\cap \mathcal{E}_s$ holds with probability at least $1-3\delta$. Corresponding to $\mathcal{E}_a$ and $\mathcal{E}_b$, we take
\begin{equation*}
    \epsilon^2=\mathcal{O}\left(\frac{1}{T}\left((d_a+d_b)\log T+\log(H/\delta)+\sqrt{m}+\sqrt{n}\right)\right). 
\end{equation*}
By (\ref{thresholdchoice}), we set
\begin{equation*}
    \kappa_h=\mathcal{O}\left(\frac{1}{T}\left(m^{7/2}+n^{7/2}+(m^3+n^3)((d_a+d_b)\log T+\log(H/\delta))+S(\log mn)^2\right)\right).
\end{equation*}
Then we have $\Vert\widehat X_h\theta_h-\wh y_h\Vert^2\leq\kappa_h$, and the confidence set with respect to this threshold contains the true parameter $\theta_h$ with probability at least $1-3\delta$. 

Since both $\Vert\wh{X}_h-X_h\Vert_{\mathrm{op}}$ and $\Vert\wh{y}_h-y_h\Vert$ are bounded by $\mathcal{O}(T^{-1})$ under event $\mathcal{E}_a\cap \mathcal{E}_b\cap \mathcal{E}_s$, by applying the same procedure as in Lemma \ref{mxc}, we conclude that, for each $h\in[H]$, there exists a constant $C_h$ independent of $T$ such that
\begin{equation*}
    d_{\mathrm{H}}(\Theta_h,\wh{\Theta}_h)\leq C_h\sqrt{\kappa_h},
\end{equation*}
which concludes the proof.
\endproof

\section{Auxiliary Lemmas}
Here we present some lemmas we make use of in the preceding proofs.
\begin{lemma}[\citealp{zhang07entropy}]\label{entropybuond}
Let $\mu$ and $\mu^*$ be two discrete probability distributions on $\{1,2,\cdots,n\}$ such that $\mathrm{TV}(\mu,\mu^*)\leq\epsilon\leq1/2$. Then
\begin{equation*}
    \vert\mathcal{H}(\mu^*)-\mathcal{H}(\mu^*)\vert\leq-\epsilon\log\epsilon-(1-\epsilon)\log(1-\epsilon)+\epsilon\log(n-1).
\end{equation*}
\end{lemma}

\begin{lemma}[\citealp{Vershynin_2018}, Proposition 5.2.2]\label{eucover}
Let $R>0$, and let $B(0,R)=\{x\in\bbR^d:\Vert x\Vert\leq R\}$ be the $R$-ball centered at $0$ in the Euclidean space $(\bbR^d,\Vert\cdot\Vert)$. Then for any $\epsilon>0$, the covering number $N(\epsilon,B(0,R),\Vert\cdot\Vert)$ admits the following bound:
\begin{equation*}
    N(\epsilon,B(0,R),\Vert\cdot\Vert)\leq\left(1+\frac{2R}{\epsilon}\right)^d.
\end{equation*}
\end{lemma}

\begin{lemma}[\citealp{Tropp_2011}, Corollary 7.5]\label{matrixmcdiarmid} Let $Z_1,\cdots,Z_n$ be a family of independent random variables, and let $H$ be a function that maps $n$ variables to a self-adjoint matrix of
dimension $d$. Consider a sequence $A_1,\cdots,A_n$ of fixed self-adjoint matrices that satisfy
\begin{equation*}
\left(H(z_1,\cdots,z_{k-1},z_k,z_{k+1},\cdots,z_n)-H(z_1,\cdots,z_{k-1},z_k^\prime,z_{k+1},\cdots,z_n)\right)^2\preceq A_k^2,
\end{equation*}
where $z_i$ and $z_i^\prime$ range over all possible values of $Z_i$ for each index $i$. Let
\begin{equation*}
    \sigma^2=\left\Vert\sum_{k=1}^nA_k^2\right\Vert.
\end{equation*}
Then for all $t>0$,
\begin{equation*}
\bbP\left(\lambda_{\max}\left(H(Z_1,\cdots,Z_n)-\bbE\left[H(Z_1,\cdots,Z_n)\right]\right)\geq t\right)\leq 1-d\cdot e^{-t^2/8\sigma^2}.
\end{equation*}
\end{lemma}

\begin{lemma}[\citealp{Abbasi2011}, Lemma 9]\label{selfnormbound}
Let $(\mathscr{F}_t)_{t=1}^\infty$ be a filtration, and let $(\eta_t)_{t=1}^\infty$ be an adapted and conditional $\sigma$-sub-Gaussian process, i.e. $\eta_t$ is $\mathscr{F}_t$-measurable, and
\begin{equation*}
\bbE[\eta_t|\mathscr{F}_{t-1}]=0,\qquad\bbE\left[e^{\lambda\eta_t}\big|\mathscr{F}_{t-1}\right]\leq e^{\lambda^2\sigma^2/2},\quad \forall\lambda\in\bbR.
\end{equation*}
Let $(X_t)_{t=1}^\infty$ be a predictable $\bbR^d$-valued process with respect to $(\mathscr{F}_t)_{t=1}^\infty$, i.e. $X_t$ is $\mathscr{F}_{t-1}$ measurable. Assume that $V_0\in\bbR^{d\times d}$ is a positive definite matrix, and for any $t\geq 0$, let $V_t=V_0+\sum_{s=1}^tX_sX_s^\top$. Let $\tau$ be a stopping time with respect to $(\mathscr{F}_t)_{t=1}^\infty$. Then, for any $\delta>0$, the following inequality holds with probability at least $1-\delta$:
\begin{equation*}
    \left\Vert\sum_{t=1}^\tau X_t\eta_t\right\Vert^2_{V_\tau^{-1}}\leq 2\sigma^2\log\left(\frac{\det(V_\tau)^{1/2}\det(V_0)^{-1/2}}{\delta}\right).
\end{equation*}
\end{lemma}

\begin{lemma}[\citealp{foster2023statisticalcomplexityinteractivedecision}, Lemma A.4]\label{foster}
Let $(\mathscr{F}_t)_{t=1}^T$ be a filtration, and let $(X_t)_{t=1}^T$ be an adapted process. Then for any $\delta>0$, it holds with probability at least $1-\delta$ that for all $t\leq T$,
\begin{equation*}
    \sum_{i=1}^t X_i\leq\sum_{i=1}^t\log\bbE\left[e^{X_i}|\mathscr{F}_{i-1}\right]+\log\frac{1}{\delta}.
\end{equation*}
\end{lemma}
\end{document}